%% file: main.tex
\documentclass[runningheads]{llncs}

\usepackage{eccv}

\usepackage{eccvabbrv}
\usepackage{wrapfig}

\usepackage{graphicx}
\usepackage{booktabs}

\usepackage[accsupp]{axessibility}  

\input{setup/packages}
\input{setup/commands}

\usepackage[pagebackref]{hyperref}

\usepackage{bibunits}
\defaultbibliographystyle{splncs04}
\defaultbibliography{main}

\makeatletter
\def\bu@hyper@bibcite#1#2{\global\@namedef{b@#1}{\hyper@@link[cite]{}{cite.#1}{#2}}}
\AtBeginDocument{\let\bibcite\bu@hyper@bibcite}
\makeatother

\makeatletter
\let\bu@org@input\@input
\def\@input#1{\begingroup\makeatletter\bu@org@input{#1}\endgroup}
\AtBeginDocument{%
  \let\bu@org@backout\Hy@backout
  \def\Hy@backout#1{\begingroup\let\@auxout\@mainaux\bu@org@backout{#1}\endgroup}%
}
\makeatother

\renewcommand*{\backref}[1]{}

\AtBeginDocument{%
}

\usepackage{orcidlink}

\renewcommand{\thefootnote}{\fnsymbol{footnote}}

\begin{document}

    \title{Comprehensive language--image pre-training for 3D medical image understanding}

    \titlerunning{COLIPRI}

    \author{
        Tassilo Wald\inst{1,2}$^{,\dagger}$\orcidlink{0009-0007-5222-2683}
        \and Ibrahim Ethem Hamamci\inst{1}$^{,\dagger}$\orcidlink{0000-0003-2932-3105}
        \and Yuan Gao\inst{1}$^{,\dagger}$\orcidlink{0009-0001-3817-5266}
        \and Sam Bond-Taylor\inst{1}\orcidlink{0000-0003-1538-7909}
        \and Harshita Sharma \inst{1}\orcidlink{0000-0003-4683-2606}
        \and Maximilian Ilse \inst{1}\orcidlink{0009-0005-8712-7307}
        \and Cynthia Lo \inst{1}\orcidlink{0000-0003-2875-0498}
        \and Olesya Melnichenko \inst{1}\orcidlink{0009-0005-3309-7413}
        \and Anton Schwaighofer \inst{1}\orcidlink{0000-0003-1557-0527}
        \and Noel C. F. Codella \inst{1}\orcidlink{0000-0001-6735-9067}
        \and Maria Teodora Wetscherek \inst{1,3}\orcidlink{0000-0003-2924-7587}
        \and Klaus H. Maier-Hein \inst{2,4}\orcidlink{0000-0002-6626-2463}
        \and Panagiotis Korfiatis \inst{5}\orcidlink{0000-0003-2516-1751}
        \and Valentina Salvatelli \inst{1}\orcidlink{0000-0002-3232-4101}
        \and Javier Alvarez-Valle \inst{1}\orcidlink{0000-0003-0906-4177}
        \and Fernando Pérez-García \inst{1}\orcidlink{0000-0001-9090-3024}
    }

    \authorrunning{T.~Wald et al.}

    \institute{
        Microsoft
        \and German Cancer Research Center (DKFZ)
        \and Department of Radiology, University of Cambridge and Cambridge University Hospitals NHS Foundation Trust
        \and Pattern Analysis and Learning Group, Heidelberg University Hospital
        \and Department of Radiology, Mayo Clinic
    }

    \maketitle
    {\renewcommand{\thefootnote}{\textdagger}%
    \footnotetext{Work done during an internship at Microsoft Research.}}

    \begin{bibunit}

    \input{setup/acronyms}

    \begin{abstract}
        In the 3D medical image domain, \visionlanguage pre-training is used to create \acp{VLE} that can support radiologists by retrieving patients with similar abnormalities, predicting likelihoods of abnormality, or, with downstream adaptation, generating radiological reports.
        While the methodology holds promise, three challenges limit the capabilities of current 3D \acp{VLE}:
        data scarcity due to privacy concerns,
        high computational costs resulting from the volumetric nature of the images, and
        a domain shift between the long reports used for training and the short prompts used during inference for, \eg, zero-shot classification.
        As a consequence, natural-image \ac{VLE} recipes do not directly transfer to 3D medical imaging.

        In this paper, we overcome these challenges by injecting additional supervision via a report generation objective and combining \visionlanguage with vision-only pre-training, allowing us to leverage both image-only and paired \imagetext 3D datasets.
        Further, we propose a novel loss that addresses the domain shift between long reports and short textual prompts.
        Through these additional objectives, paired with best practices of the 3D medical imaging domain, we develop the \textbf{Co}mprehensive \textbf{L}anguage--\textbf{I}mage \textbf{Pr}e-tra\textbf{i}ning (COLIPRI) encoder family.
        Our \acs{COLIPRI} encoders achieve \acl{SOTA} performance in report generation, semantic segmentation, classification probing, and zero-shot classification.

        The model weights and inference code are freely available at \modelurl.

    \end{abstract}

    \input{figures/pca_small}

    \section{Introduction}
    \label{sec:introduction}

    \Ac{CLIP}~\cite{radford2021learning} has established itself as one of the strongest paradigms to learn general-purpose image and text representations.
    Aside from being a solid starting point for adaptation to downstream tasks of interest \cite{goyal2023finetune,luddecke2022image}, having language-aligned vision embeddings allows leveraging natural language for open-set classification \cite{radford2021learning} and open-set segmentation \cite{zhou_extract_2022}.
    In 3D medical imaging, this training paradigm is especially relevant as every image acquired in a clinical setting is accompanied by a report, making this kind of paired data abundant in hospitals.
    The reports associated with the images offer (weak) supervision and enable zero-shot tasks for clinical support, such as image-to-report retrieval or open-set abnormality classification.

    Despite these promises, the field of \visionlanguage pre-training with medical images is not as mature as its general-domain counterpart.
    While \ac{CLIP}~\cite{radford2021learning}, Perception Encoder~\cite{bolya_perception_2025}, or \sigliptwo~\cite{tschannen_siglip_2025} are well established in the general domain, 3D medical \acp{VLE} have only recently started to garner attention~\cite{hamamci_developing_2024,blankemeier_merlin_2024}.
    We believe that this can be attributed to two key issues:
    \begin{inlinelist}
        \item the lack of large, publicly available, paired datasets in the medical domain and
        \item domain- and modality-specific methodological and engineering hurdles.
    \end{inlinelist}

    The largest publicly available 3D datasets with \imagereport pairs reach a combined size of about 78k \imagereport pairs (\cref{apx:tab:datasets}), which is far from the 400 million image--text pairs which \ac{CLIP} \cite{radford2021learning} was trained on, and even further from the 12 billion alt-texts used to train \sigliptwo~\cite{tschannen_siglip_2025}.
    However, the relatively large number of available unpaired images has the potential to substantially increase the amount of overall usable data, when leveraged through \ac{SSL} (\cref{apx:tab:datasets}).

    The number of voxels in 3D medical images is orders of magnitude larger than the number of pixels in 2D images from the general or medical domains.
    For example, a typical chest CT volume may be composed of \sizeaniso{512}{512}{320} voxels, making it difficult to use entire images in native resolution during training due to excessive VRAM requirements.
    It is therefore common to train with image crops~\cite{shui_large-scale_2025,perez-garcia_torchio_2021} or downsampled images~\cite{hamamci_developing_2024,blankemeier_merlin_2024}.
    Cropping may complicate \ac{CLIP} training as clinical reports refer to the entire volume, not just the \ac{FOV} in the subvolume, and downsampling dismisses high-resolution information which might be crucial for detecting small abnormalities.
    Lastly, the reports of 3D images are substantially longer than the short prompts typically used during inference for, \eg, zero-shot classification, which results in a large domain shift that needs to be addressed.

    In this work, we overcome the aforementioned challenges, advancing the state of the art in 3D medical \visionlanguage models.
    Our contributions can be summarised as follows:
    \begin{enumerate}
        \item To reduce the domain shift between long reports used for training and short textual prompts used for zero-shot classification, we introduce the novel \acf{OSL}, which improves results substantially by closing the gap between domains.
        \item To address the data scarcity, we increase supervision from medical reports by adding a text generation objective, and introduce a vision-only objective so that large image-only datasets can be leveraged for complementary \ac{SSL} pre-training. 
        \item We comprehensively evaluate the resulting models through zero-shot classification, classification probes, report generation, retrieval, and semantic segmentation, demonstrating that our \colipricrm models represent the new state of the art.

    \item We share model weights, code for inference, and multiple usage examples at \href{https://huggingface.co/microsoft/colipri}{https://huggingface.co/microsoft/colipri}.  
    \end{enumerate}

    \input{figures/diagram}

    \newpage

    \section{Related work}
    \label{sec:related_work}
    Contrastive \visionlanguage models such as \ac{CLIP}~\cite{radford2021learning} have demonstrated that large-scale paired image--text data enables transferable multimodal representations.
    Later variants improved the contrastive objective~\cite{zhai_sigmoid_2023}, or proposed a language-generation objective~\cite{tschannen2023image,wan2024locca}, with others including additional self-supervised objectives~\cite{naeem2024silc,maninis_tips_2025,tschannen_siglip_2025}.
    In parallel to this, \acf{SSL} methods~\cite{zhou2021ibot,simeoni2025dinov3} were developed to learn general purpose features from unpaired images, achieving strong general-purpose visual features in natural imaging.

    Compared to natural imaging, 3D medical imaging incurs much higher computational costs due to its higher dimensionality and faces stricter data access constraints, limiting the overall amount of available data.
    These factors have hindered progress of 3D image \ac{SSL}, with a recent benchmark~\cite{wald2025openmind} showing that features generated by most 3D \ac{SSL} methods~\cite{zhou2021models,tang2022self,wu2024voco} improve either dense or global downstream tasks, but generally not both.
    For dense downstream tasks, \acp{MAE}~\cite{munk2024amaes,wald2025revisiting} remain the strongest pre-training method, yielding features that adapt well for 3D semantic segmentation.

    Medical \acp{VLE} have largely been trained on (2D) chest X-rays with paired reports~\cite{huang2021gloria,boecking2022making,Bannur_2023_CVPR,wang2022medclip}.
    Early work on 3D medical images focused on adapting contrastive language--image pre-training to 3D settings~\cite{hamamci_developing_2024,blankemeier_merlin_2024}, incorporate anatomy awareness~\cite{shui_large-scale_2025}, or auxiliary supervision through chest X-rays~\cite{cao_bootstrapping_2024}.
    Despite increased interest in the field of 3D \ac{SSL}~\cite{pai_vision_2025,Xu2025-vv} and \visionlanguage pre-training, a research gap remains between natural imaging and 3D medical imaging.
    Thus far, no work has combined self-supervised and \visionlanguage objectives, nor has any research in the 3D imaging domain introduced a text generation objective.

    \section{Method}
    \label{sec:method}
    We present \acf{COLIPRI}, a multimodal pre-training framework that learns joint representations between 3D medical images and associated radiology reports.
    \method combines contrastive \visionlanguage alignment, increases supervision of limited data through \acf{RRG}, extends available data to unpaired images through \ac{MIM}, and closes the long-report-to-short-zero-shot-prompt distribution shift through our novel \acf{OSL} to produce 3D \visionlanguage representations applicable to both global reasoning and dense prediction tasks (\cref{fig:colipri_diagram}).
    We develop our models in stages, each adding a learning objective to assess the effects of each building block.

    \subsection{Report preprocessing}
    \label{sec:method:report_preprocessing}
    Radiological reports often contain details about all organs imaged, stating whether the findings are normal or abnormal.
    Should abnormalities be present, they are explicitly named; however, when absent, the abnormalities are often not listed, so as not to yield an excessive list of abnormalities a patient does not suffer from.
    The exceptions are typically abnormalities that might have prompted the imaging study in the first place.
    This style of reporting results in reports of substantial token sequence lengths, with the average \findings section of \ctrate~\cite{hamamci_developing_2024} being 243 tokens long when using the \bert tokeniser~\cite{boecking2022making} (\cref{fig:long_report_domain_shift}).  
    This is in stark contrast to Zhang \etal~\cite{zhang2022contrastive} or Radford \etal~\cite{radford2021learning}, which either sample a single sentence from the paired caption or whose datasets contain single-sentence captions.

    To enable the objectives introduced in \cref{sec:method:clip_adaptation,sec:methods:rrg}, we use a \ac{LLM} to restructure each report into eight clinical subsections, denoted as $C$, assigning every sentence to one subsection.
    We also leverage the \ac{LLM} to create concise versions of all sentences $s$ and to label each as a \textit{positive finding} (pathology present) $s^+$ or \textit{negative finding} (normal) $s^-$.
    These preprocessed reports enable
    \begin{inlinelist}
        \item language regularisation,
        \item the \ac{OSL}, and
        \item our radiological report generation formulation.
    \end{inlinelist}
    Details, prompts and examples are provided in \cref{apx:report_preprocessing}.

    \subsection{Adapting CLIP to 3D radiology}
    \label{sec:method:clip_adaptation}
    \method leverages the Primus-M transformer~\cite{wald2025primus} architecture as the image encoder, and \bert~\cite{boecking2022making} (pre-trained on chest X-rays) as the text encoder.
    The dense tokens of the vision and text encoders are aggregated through a multi-head attention-pooling mechanism to yield global representations.
    Due to different embedding dimensions of image and text encoders, the image embeddings are projected to the lower-dimensional text space, which can then be aligned using \ac{CLIP}~\cite{radford2021learning}.

    In contrast to natural images, radiological reports describe findings across the entire scan, requiring a sufficiently large \ac{FOV} for global context.
    However, token sequence length increases cubically with \acp{FOV} in 3D imaging, making training computationally expensive.
    We use inputs of size \sizeiso{160} with an isotropic spacing of 2 mm, balancing \ac{FOV} and token length for efficiency.
    We train with the default \ac{CLIP} objective, which outperforms a sigmoid loss~\cite{Zhai_2023_ICCV} in our setting.

    The \findings section in radiology reports often resembles a long list, making the text encoder prone to overfitting to sentence order or stylistic cues rather than semantics.
    To counter this, we introduce two complementary text augmentations:
    \begin{inlinelist}
        \item the \sentenceshuffle augmentation randomly permutes sentences within a report, discouraging reliance on order;
        \item the \shortsentence augmentation replaces long-form reports with \ac{LLM}-shortened statements with a given probability during training. This reduces the domain shift generated by training on long reports and running inference using brief zero-shot prompts (\eg, ``\{abnormality\} present'' or ``No \{abnormality\} present'') but does not fully close it.
    \end{inlinelist}
    Additional ablations and detailed hyperparameters are provided in \cref{apx:method:optimising_clip}.

    \subsubsection{Closing the zero-shot domain gap}
    \label{sec:method:opposite_sentence_loss}
    Using the \ac{LLM}-structured, shortened, and categorised reports (\cref{sec:method:report_preprocessing}), we close the training--inference domain shift through our novel \acf{OSL}.
    Let $x_i$ denote the image for case $i$, and let $s_i^+$ denote a \textit{positive findings} sentence sampled from the report $r_i$. We create $s_i^-$, a negation of $s_i^+$, via the template ``\texttt{No \{sentence\}.}''. Since $s_i^+$ originates from the report of case $i$ then the sentence $s^+_i$ correctly describes the image, yielding label $y_i=1$.
    Additionally, for any clinical subsection of the current report that contains no positive findings, we sample positive findings sentences from the same subsection in other reports.
    These sampled positive findings do not describe the current image and therefore serve as negative examples, paired with their negated counterparts to form $(s_i^+, s_i^-)$ pairs with supervision label $y_i=0$.
    Let $e_i^+$ and $e_i^-$ be the text embeddings of the positive and negated statement pair, and $v_i$ the global image embedding.
    We compute a cosine similarity between image and text embeddings, followed by a temperature-scaled softmax over the pair and optimise a cross-entropy loss between probability $p^+_i$ and label $y_i$ to maximise $sim(e^+_i\mid v_i)$ if $s_i^+$ describes the image and minimises $sim(e^+_i\mid v_i)$ if $s^+_i$ does not describe the image (described in \cref{apx:methods:opposite_sentence_loss} and \cref{alg:ost}).
    This trains the model in a short-prompt manner and with statements of abnormality presence and absence, as used during zero-shot inference, explicitly learning to disambiguate presence vs.\ absence of findings.
    We combine our \ac{OSL} objective with the CLIP objective,
    $\mathcal{L}_{\text{align}}
    = 0.5\cdot\mathcal{L}_{\text{CLIP}}
    + 0.5\cdot\mathcal{L}_{\text{OSL}}$,
    yielding our intermediate \colorbox{microsoftred!20}{\colipric} method (\cref{apx:tab:colipri_eval}).

    \subsection{Increasing supervision from limited data by generating reports}
    \label{sec:methods:rrg}
    To enrich the visual representations generated by the encoder, we introduce a \acf{RRG} objective.
    This task aims to ensure that the vision encoder captures the complete semantic content of the associated report, rather than only discriminative cues required for contrastive pairing.
    Following Tschannen \etal~\cite{tschannen2023image}, we evaluated two decoding modes during training: \textit{causal captioning} and \textit{parallel captioning}.
    In causal captioning mode, the decoder operates autoregressively, predicting each token conditioned on all preceding tokens,
    $p_\text{causal}(y_t \mid y_{<t}, \, \textbf{v})$,
    where $\textbf{v}$ denotes the dense visual features and $y_{<t}$ the previously generated tokens.
    This setup enables sequential report generation, akin to standard language modelling.
    In contrast, the parallel (masked) mode predicts all tokens simultaneously from a fully masked input,
    $p_\text{parallel}(y_t \mid \mathbf{M}, \, \textbf{v})$,
    where $\mathbf{M}$ denotes a mask applied to all text tokens.
    This formulation forces the text decoder to use the dense visual representations $\textbf{v}$ rather than previously generated text $y_{<t}$.
    We found that \textit{parallel captioning} yields better results, and hence discarded \textit{causal captioning} in our final \method models (see \cref{tab:cappa_ablation}).

    We implement \ac{RRG} using a lightweight EVA-02~\cite{FANG2024105171} transformer decoder connected to the vision encoder via cross-attention.
    To mitigate order ambiguity inherent to free-form clinical text, we use the \ac{LLM}-structured reports introduced in \cref{sec:method:report_preprocessing}, where each sentence is assigned to one of eight clinical subsections.
    These structured reports reduce positional uncertainty by providing a consistent ordering of findings across studies, ensuring that the decoder learns to model clinical content rather than author-specific stylistic variation.
    During decoding, clinical subsection headers are kept unmasked to guide generation and preserve report semantics.
    The \ac{RRG} objective is optimised jointly with the contrastive and \ac{OSL} objectives: $\mathcal{L}_{\text{VLM}} = 0.5\cdot\mathcal{L}_{\text{CLIP}} + 0.5\cdot\mathcal{L}_{\text{OSL}} + \lambda_{\text{RRG}}\mathcal{L}_{\text{RRG}}$ on paired image--report batches, yielding our intermediate \colorbox{microsoftblue!20}{\colipricr} method (see \cref{apx:tab:colipri_eval}).

    \subsection{Expanding available data through vision-only self-supervision}
    \label{sec:methods:vision_only}

    While multimodal pre-training provides strong representations for global reasoning tasks, such methods rely on scarce \imagereport pairs and often underperform on dense, spatially localised tasks~\cite{oquab2023dinov2}.
    To increase available data to unpaired images and to improve the localised performance of the learnt representations, we include a vision-only \acf{MAE} objective~\cite{he2022masked}.

    Given the lack of advanced dense \ac{SSL} methods such as \ibot~\cite{zhou2021ibot} or \dinovthree~\cite{simeoni2025dinov3} in 3D medical imaging, \ac{MAE} pre-training provides a straightforward and effective choice.
    It reconstructs missing volumetric patches from context, encouraging the model to learn dense feature representations that complement the global semantic alignment achieved by the \ac{CLIP} and \ac{RRG} objectives.
    To provide a strong initialisation, we pre-train a Primus-M vision encoder first, leveraging the \texttt{nnSSL} framework~\cite{wald2025openmind}, before joint multimodal training.

    We implement \ac{MAE} as a lightweight decoder attached to the vision encoder to reconstruct masked patches in voxel space using a mean-squared-error loss.
    To avoid interference of the vision-only masking with the \visionlanguage objectives, we use a training scheme that alternates between vision-only and \visionlanguage objectives, optimising either $\mathcal{L}_{\text{Align}} = 0.5\cdot\mathcal{L}_{\text{CLIP}} + 0.5\cdot \mathcal{L}_{\text{OSL}} $
    or $ \mathcal{L}_{\text{VO}} = \lambda_{\text{MAE}}\mathcal{L}_{\text{MAE}}$.
    This design allows benefitting from paired supervision when available, while still leveraging the often much larger image-only datasets such as NLST~\cite{national2011national}.

    Finally, since the vision-only objective does not require a global \ac{FOV}, we sample high-resolution (1 mm isotropic) and default-resolution (2 mm isotropic) sub-crops during \ac{MAE} pre-training.
    This exposes the encoder to finer spatial detail, reducing the shift between pre-training and high-resolution downstream data.
    The resulting combination of \colipric with the MAE objective yields our \colorbox{microsoftgreen!20}{\colipricm} \ac{VLE}.
    Further, combining all of the introduced objectives (\ac{CLIP}, \ac{OSL}, \ac{RRG}, and \ac{MAE}) yields our final \ac{VLE}, \colorbox{microsoftyellow!20}{\colipricrm} (\cref{apx:tab:colipri_eval}).

    \section{Experiments}
    \label{sec:experiments}

    We pre-train our \method models on \ctrate~\cite{hamamci_developing_2024} (for the vision-language and vision-only objectives) and NLST~\cite{national2011national} (for the vision-only objective).
    Dataset and training details are presented in \cref{apx:development_framework,apx:methods_and_experiments}.
    The resulting encoders are evaluated on multiple unimodal (semantic segmentation, multilabel classification) and multimodal (zero-shot classification, report generation, retrieval) downstream tasks.

    \subsection{Classification and report generation}
    Classification performance is evaluated on a withheld test set of \ctrate and on the publicly available subset of \radchestct~\cite{draelos_rad-chestct_2020}, which comprises 3.6k chest CT volumes with 16 multi-abnormality labels that can be derived from the original \ctrate abnormality classes.
    We evaluate linear classification probes, as well as zero-shot classification performance on both datasets (\cref{apx:sec:classification_linear_probes}).


    We also evaluate the quality of the frozen image encoder embeddings for report generation.
    To do this, we follow the \llava framework \cite{liu2023visual}, with image tokens passed through a two-layer \ac{MLP} to integrate them into the language space of the \qwen 1B base model~\cite{team2024qwen25}.
    We fine-tune the \ac{LLM} to generate the \findings section of each report (see details in \cref{apx:sec:generative_decoding}).
    To evaluate the clinical accuracy of generated reports, we use the text classifier trained by Hamamci \etal~\cite{hamamci_developing_2024}, based on RadBERT~\cite{yan2022radbert}.
    We also use the \radfactctpm{} and \radfactctp{}, variants of RadFact~\cite{bannur2024maira} with CT-specific system prompts and few-shot examples as described in \cref{apx:sec:evaluation:metrics}.

    As baselines, we compared our method against established CT models, namely CT-CLIP~\cite{hamamci_developing_2024}, CT-FM~\cite{pai2025vision} (both trained on chest CT datasets), and Merlin~\cite{blankemeier_merlin_2024} (trained on abdominal CTs, but we found it to be a competitive baseline for chest CT report generation).
    For zero-shot classification, we additionally compare against fVLM~\cite{shui_large-scale_2025} and BIUD~\cite{cao_bootstrapping_2024}.
    We also compare our method against two 2D \acp{VLE} which included CT slices in their pre-training data: \ac{MI2}~\cite{codella_medimageinsight_2024}, a general biomedical imaging model, and Curia~\cite{dancette2025curiamultimodalfoundationmodel}, a radiology-specific model.
    We evaluate linear and zero-shot classification for the two 2D models by averaging the pixel values of a volume along the axial dimension and encoding the resulting average slice (we also tried max, mean, standard deviation and median embedding pooling~\cite{codella_medimageinsight_2024} but found that averaging intensities across slices works best).
    For \ac{RRG}, we instead encode all axial slices independently and perform max pooling of the embeddings.

    \subsection{Semantic segmentation}
    To evaluate the quality of the vision encoder for dense tasks, we measure 3D medical image segmentation performance after fine-tuning the encoder.
    In this setting, we compare against a Primus-M~\cite{wald2025primus} encoder trained from scratch, as well as a \acl{SOTA} \ac{MAE}-pre-trained~\cite{wald2025openmind} Primus-M encoder trained on CT-RATE and NLST (\cref{sec:dataset_ctrate}).
    Additionally, we compare with the default nnU-Net~\cite{isensee2021nnu} and ResEnc-L~\cite{isensee2024nnu} CNN encoders trained from scratch as references.
    All training runs are conducted using the \nnunet framework~\cite{isensee2021nnu}, with all of the encoders being fine-tuned in a short (37.5k steps) and a long (250k steps) schedule, using the learning rate schedule described in Wald \etal~\cite{wald2025openmind}, with the peak learning rate reduced to $10^{-4}$.
    To remain partially in distribution, we chose to focus on segmentation datasets with targets in the chest region or the upper abdomen, a \ac{FOV} that is often visible during pre-training.
    We use five-fold cross-validation for each dataset.
    We evaluate segmentation performance on
    \begin{inlinelist}
        \item LiTS~\cite{simpson2019large} ($N$ = 131), task 3 of the \ac{MSD}, which contains segmentations for liver and liver tumours.
        \item Lung~\cite{simpson2019large} ($N$ = 64), task 6 of the \ac{MSD}, which contains cases of primary lung cancers.
        \item HVS~\cite{simpson2019large} ($N$ = 303), task 8 of the \ac{MSD}, which focuses on segmenting hepatic vessels and adjacent tumours.
        \item KiTS23~\cite{heller2023kits21} ($N$ = 489), a more recent dataset focused on segmenting tumours, cysts, and kidneys.
    \end{inlinelist}
    We use the \ac{DSC} averaged across all foreground classes as our summary metric for segmentation performance.

    \section{Results and discussion}
    \label{sec:results_discussion}

    \subsection{Classification probes}
    \label{sec:results_and_discussion:classification}

    \begin{table}
        \centering
        \caption{
            \textbf{Classification probing results.}
            We compare the embedding quality of our vision encoders against publicly available baselines.
            Our \colipricm and \colipricrm encoders yield the best classification results, exceeding all baselines on both datasets across all metrics.
            The metrics for ``CT-CLIP (reported)'' are from Shui \etal~\cite{shui_large-scale_2025}.
            Differences in performance with the metrics we computed using the released checkpoints may be due to configuration issues.
            Hence, we report both sets of values for fairness and clarity.
            We report macro average and 95\% \acp{CI} based on 100 bootstrap samples.
            \textbf{Bold} indicates best performance for that metric, or overlapping \acp{CI} with best.
            AUPRC: area under the precision-recall curve;
            AUROC: area under the receiver operating characteristic curve.
        }
        \label{tab:test_classification}
        \resizebox{\linewidth}{!}{\input{tables/results/classification}}
    \end{table}

    We evaluate our \method vision encoders and other baseline vision encoders using classification probes on \ctrate and \radchestct (\cref{tab:test_classification}).

    Our probing setup (\cref{apx:sec:classification_linear_probes}) is largely superior to the ones used in the baselines papers, yielding AUROC values of above 80\% for the majority of baseline methods (vs.\ approx.\ 75\% reported in the original works) as well as our own encoders.
    This likely originates from our training setup, which uses multiple probes with different learning rates and token pooling methods, selecting the best probe on the validation split for testing.
    Moreover, not all of our token aggregation schemes are linear as some include a light-weight attention pooling block, which is more flexible than a linear layer.
    We believe this multi-probe scheme to be more suited for comparison due to its robustness to hyperparameter selection, and its independence of the existence or quality of global-pooling layers attached to the vision backbones.
    The only exception for this performance increase is the \ctclip encoder, which curiously performs worse in our experiments.
    This might be due to a potential configuration issue, hence we additionally report the results from Sui \etal~\cite{shui_large-scale_2025} for the metrics in common.

    Our \method models exceed all baselines across all metrics and datasets, with \colipricm and \colipricrm representing the strongest encoders for classification.
    In particular, \colipricrm increases by 6 points AUPRC and 3.5 points AUROC over the best-performing baseline Merlin.
    The inclusions of our \ac{RRG} and \ac{MAE} objectives increases classification performance, likely due to the added objectives serving as regularisation.

    \subsection{Zero-shot classification}
    \label{sec:results_and_discussion:zeroshot}
    \begin{table}
        \centering
        \caption{
            \textbf{Zero-shot classification results.}
            We report macro AUPRC and AUROC excluding `Medical Material' and `Lymphadenopathy' to be consistent with \cite{shui_large-scale_2025}. An extended version of this table with all abnormalities is provided in \cref{tab:apx:classification_zeroshot}.
            Across both metrics and datasets, our \acp{VLE} exceed the state of the art using short prompts, without relying on segmentation masks at inference as fVLM does.
            We report mean and 95\% \acp{CI} based on 100 bootstrap samples.
            \textbf{Bold} indicates best performance for that metric, or overlapping \acp{CI} with best.
        }
        \label{tab:results_and_discussion:zeroshot_classification_main}
        \resizebox{.85\linewidth}{!}{\input{tables/results/zeroshot_classification_main}}
        \caption*{
            \footnotesize$^*$Values are taken from \cite{shui_large-scale_2025}.
        }
    \end{table}
    \begin{wraptable}{r}{0.45\textwidth}
    \centering
    \caption{
        \textbf{The \acf{OSL} enables short-form zero-shot classification.}
        Effect of the \ac{OSL} on the validation set results.
        Without the \ac{OSL}, zero-shot classification with short prompts performs substantially worse than using a long-form, more native prompting style.
    }
    \label{tab:opposite_sentence_ablation}
    \resizebox{\linewidth}{!}{\input{tables/results/zeroshot_classification_opposite_sentence}}
\end{wraptable}
    For all \acp{VLE}, we report  the zero-shot classification performance on \ctrate and \radchestct in \cref{tab:results_and_discussion:zeroshot_classification_main}, using a short-form prompting scheme for our models (\cref{apx:development:eval_tasks_and_metrics}).
    Values for some baselines are taken from the fVLM paper~\cite{shui_large-scale_2025}.
    However, in their work, results are reported for only 16 of the 18 abnormalities in \ctrate, excluding the non-localisable `Lymphadenopathy' and `Medical Material' as they are not associated with an organ%
    \footnote{\url{https://github.com/alibaba-damo-academy/fvlm/issues/12\#issuecomment-3283463870}}%
    , which is a limitation of fVLM.
    We report results including all abnormalities in \cref{tab:apx:classification_zeroshot}.

    Our \colipricm and \colipricrm encoders outperform all baselines on \ctrate, while the \colipric and \colipricr encoders perform worse.
    When evaluated on \radchestct, \colipricrm generalises better than our other encoders or fVLM, with fVLM decreasing 10 points in AUROC on \radchestct compared to \ctrate.

Including the \ac{OSL} during training closes the gap between a `native', longer-form prompting scheme where we average the embeddings of 50 randomly sampled reports with the abnormality and 50 without, versus the short-form prompts where we query with ``\{abnormality\} present'' and ``no \{abnormality\} present'' (\cref{tab:opposite_sentence_ablation}).
The \ac{OSL} improves the utility of the encoder as short queries are substantially easier to create than the `native', longer-form queries.
We hypothesise that the \ac{OSL} can generalise beyond 3D medical imaging to other domains with highly detailed reports or captions.

    \subsection{Report generation}
    \label{sec:result_and_discussion:reportgen}

    \begin{figure*}[t]
        \centering
        \includegraphics[width=\linewidth]{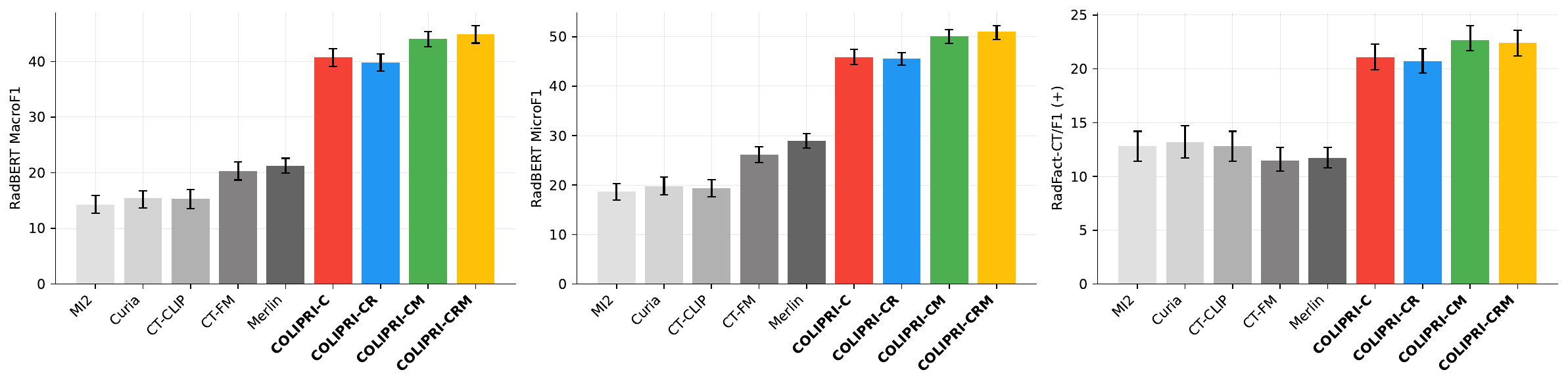}
        \caption{
            \textbf{Report generation results (summary)}.
            Our \method vision encoders enable generating reports substantially better than the state of the art.
            In particular, the \ac{LLM} trained on top of our models generates more accurate statements about abnormalities being present, as measured by \radfactctpf{} and the RadBERT $F_1$.
            The exact values are presented in \cref{tab:apx:results_generation}.
            \radfactctpf{}: summary of logical precision and recall of positive findings (i.e., describing a present abnormality as judged by GPT-4o)~\cite{bannur2024maira}.
            We report median and 95\% \acp{CI} (error bars) based on 500 bootstrap samples.
        }
         \label{fig:report_generation_clinical}
    \end{figure*}

    We evaluate the impact of our pre-training strategy on downstream \ac{RRG} (details in \cref{apx:sec:generative_decoding}) using both lexical and clinical metrics (\cref{fig:report_generation_clinical}, \cref{tab:apx:results_generation}).
    Across all lexical metrics (ROUGE-L, BLEU1, BLEU4, and METEOR), our \colipricrm encoder performs slightly better than the baselines, meaning that they all generate lexically accurate reports.
    However, high lexical overlap does not necessarily imply clinically accurate reports~\cite{li2025reevalmed}.

    When looking at clinical metrics, our \method vision encoders outperform all baselines by a very large margin.
    We use $F_1$ scores from the RadBERT classifier and \radfactctp{}, which measure the correctness of medical entities and factual statements.
    Specifically, \colipricrm improves by 23 points in RadBERT Macro$F_1$ and by 9 points in \radfactctpf{} over the strongest baseline, reflecting that the produced reports contain fewer omissions and more specific diagnostic statements.
    This demonstrates that our pre-training paradigm yields representations that encode more clinically relevant semantics.
    In absolute terms, however, the overall accuracy of generated reports for 3D medical is still low, with Macro $F_1$-Scores of around 45\% for the abnormalities measured by RadBERT and $F_1$-scores of around 22\% for all abnormalities as measured by \radfactctp{}.
    \colipric and \colipricr exhibit slightly lower performance compared to the encoders including \ac{MAE} pre-training.
    Interestingly, the inclusion of the \ac{RRG} objective appears to have a minimal impact on overall \ac{RRG} performance, despite explicitly optimising embeddings to facilitate report generation.
    The inclusion of the \ac{MAE} objective, on the other hand, appears to substantially improve performance.
    As such, we hypothesise that the \ac{MAE} objective, which encourages the vision encoder to learn low-level semantic representations, better supports the \llava framework~\cite{liu2023visual}.
    The \ac{RRG} supervision might lead to more language-aligned features, which may result in the loss of more fine-grained features.
        \begin{table}[t]
        \centering
        \caption{
            \textbf{Report-to-image retrieval results.}
            Retrieval results on our \ctrate test set, after deduplicating identical reports, yielding a total sample size of $N$ = 1493.
            CT-CLIP retrieval values are taken from Hamamci \etal~\cite{hamamci_developing_2024}.
        }
        \resizebox{.8\linewidth}{!}{
        \input{tables/results/test_retrieval_T}
        }
        \label{tab:results:retrieval}
    \end{table}
    Comparing the values of \radfactctp{}, which considers only statements mentioning the presence of abnormalities, and \radfactctpm{}, which considers statements about healthy organs and statements about abnormalities, reveals very different behaviour (see \cref{tab:apx:results_generation}).
    While our encoders reach substantially better metrics in the \mbox{(+)} setting, some of the baselines achieve slightly higher scores than our encoders as measured in the \mbox{($+/{-}$)} setting.

    This is attributable to the vast imbalance of statements about presence vs.\ absence of abnormalities, with the latter dominating the metric.
    Thus, the inclusion of normal (negative) findings is an important but double-edged aspect of medical report generation.
    Since statements about the absence of abnormalities improve apparent completeness and boost \mbox{($+/{-}$)} metrics, they can mask a low diagnostic sensitivity, as highlighted in our \mbox{(+)} results.

    \subsection{Retrieval}
    \label{sec:results_and_discussion:retrieval}
    Report-to-image retrieval results on our \ctrate test set are presented in \cref{tab:results:retrieval}.
    All our \method \acp{VLE} are substantially better in retrieving the associated image given a report.

    In particular, our \colipricrm \ac{VLE} yields a R@5 of 35.1\% compared against a R@5 of 2.9\% of CT-CLIP, highlighting the strong multimodal alignment of our encoders.
    The inclusion of the \ac{RRG} and \ac{MAE} objectives positively benefit retrieval performance, with the \ac{MAE} yielding larger performance improvements.

    \subsection{Segmentation}
    \label{sec:results_and_discussion:semantic_segmentation}
    We evaluate semantic segmentation performance on the LiTS, Lung, HVS and KiTS23 datasets, against a baseline trained from scratch and an MAE pre-trained on NLST and \ctrate at 1-mm and 2-mm isotropic resolutions, using the \texttt{nnSSL} framework~\cite{wald2025openmind} (\cref{tab:segmentation_results}).

        \begin{table}[t]
        \centering
        \caption{
            \textbf{Segmentation fine-tuning results.}
            Mean \ac{DSC} results of five-fold cross-validation trained for 37.5k steps or 250k steps.
            Across all datasets, our \colipricm and \colipricrm vision encoders perform on par or better than the \ac{SOTA} pre-training method \ac{MAE}, when adapted for downstream segmentation.
            We highlight the best short and long fine-tuning results of the same architecture in bold, and second best underlined.
        }
        \label{tab:segmentation_results}
        \resizebox{\linewidth}{!}{\input{tables/results/segmentation_short_T}}
        \end{table}

    In the short training regime of 37.5k steps, our encoders including \ac{MAE} pre-training consistently exceed training from scratch, exceed or match the strong purely \ac{MAE} pre-trained baseline.
    The encoders without the \ac{MAE} objective also exceed training from scratch, but show lower segmentation performance, highlighting the importance of including the \ac{MAE} objective for dense downstream task adaptation.
    This low performance of the \colipric and \colipricr models is not surprising, as it was shown that contrastive baselines struggle to surpass a baseline trained from scratch for segmentation tasks in Wald \etal~\cite{wald2025openmind}.

    When fine-tuning for 250k steps, our final \colipricrm model exceeds the pure \ac{MAE} pre-training and improves for LiTS and KiTS, but decreases for Lung and HVS relative to the short fine-tuning schedule.
    For the Lung dataset, training from scratch yields better results than using pre-training.
    This is not uncommon as fine-tuning for longer may lead to overfitting and decreasing performance~\cite{wald2025openmind}.
    Comparing the nnU-Net default and ResEnc-L references with the results of the Primus-M encoder trained from scratch, it is clear that, on LiTS and KiTS, where Primus-M is close in performance to the CNN references, pre-training allows to exceed the default nnU-Net, while the gap to the ResEnc-L is not closed.
    This suggests that transformer encoders are generally still lagging behind their \ac{CNN} counterparts, indicating the need for transformer architectures with stronger segmentation capabilities.

    \subsection{Qualitative analysis}

    Aside from quantitative results, we provide a \ac{PCA} visualisation of the 3D embeddings of Merlin, CT-FM, CT-CLIP, and our \colipricrm encoder, on a lung cancer case from the \ac{MSD} Lung dataset (\cref{fig:pca}).
    The embedding resolution is very low for Merlin and CT-FM, providing hardly any semantic localisation.
    \ctclip yields embeddings of higher resolution, allowing features to be visually mapped from the input CT to the \ac{PCA} map.
    However, the \ac{PCA} map is inconsistent and noisy, and exhibits a strong bias towards absolute position within the scan, as visible through the anteroposterior green/red shift.
    On the other hand, our \method encoders yield higher-resolution embeddings, which are sharper and more consistent, allowing for clear recognition of the boundaries of the patient, lungs, and the abdominal organs, as well as the lung mass present in the right lung (on the left-hand side of the coronal and axial slice views).

    \subsection{Summary of the results}
    Our encoders learn clinically relevant features, as measured by the linear separability of features to abnormality classes  (\cref{sec:results_and_discussion:classification}) and by our higher clinical accuracy in report generation, yielding \ac{SOTA} results in both aspects.

    They also show great text-to-image alignment, with substantial improvements in report-to-image retrieval, and are sensitive to fine-grained similarity measurements, as evidenced by our \ac{SOTA} zero-shot classification results(\cref{sec:results_and_discussion:zeroshot}).
    While the former is an effect of our additional \ac{MAE} and \ac{RRG} supervision signals, the latter is enabled by our novel \acf{OSL}, which teaches our text encoder to embed short statements in a semantically meaningful way.
    However, a substantial gap remains between classification probes and zero-shot classification, indicating potential for further improvement in alignment.
    Regarding dense tasks, our encoders including the \ac{MAE} objective yield results that are better or on par with the current \ac{SOTA} \ac{SSL} pre-training method \ac{MAE} (\cref{sec:results_and_discussion:semantic_segmentation}).
    Compared to \ac{SOTA} CNNs, however, our transformer architecture still limits the performance of our encoders, yielding worse performance than ResEnc-L.

    Across all our experiments, we find that our \colipricrm encoder provides embeddings that are potent in global tasks and is adaptable to dense downstream tasks, demonstrating the strength and versatility of our method.


    \section{Limitations}
    \label{sec:conclusion}

    We discuss below various opportunities for improvement, from a technical and clinical point of view.

    On the technical side, the inclusion of the \acf{RRG} objective yields only slight improvements,  indicating insufficiencies in the objective formulation.
    This positions the \acf{RRG} objective more as an auxiliary objective improving retrieval. Should one not be interested in retrieval, the \acf{RRG} objective can be considered optional, with the COLIPRI-CM configuration performing similarly to the full COLIPRI-CRM.
    On a more general note, there remains a large gap between the trained classification probe and the zero-shot classification performance, highlighting the need for better alignment.

    On the clinical side, despite the improvements achieved by \colipricrm, the clinical performance metrics remain below the thresholds one would expect in clinical practice.
    We anticipate that training the model on a substantially larger and more diverse dataset would further enhance its performance.
    Additionally, a more comprehensive evaluation, stratified by clinical findings and patient characteristics, would be necessary to fully assess and demonstrate the model's potential clinical utility.


    \section{Conclusion}
    \method achieves \ac{SOTA} performance across all standard tasks, including classification, retrieval, segmentation, and report generation.
    Extensive evaluation on established benchmarks demonstrates consistent improvements over prior work, including stronger generalisation across datasets. These results confirm that \method advances the empirical frontier of the field and establishes a new performance baseline for the commonly studied tasks.

    \section*{Authors contributions}


    Conceptualisation: TW, IEH, YG, FPG.
    Data curation: TW, IEH, OM, MTW, FPG.
    Formal analysis: TW, YG, SBT, HS, MI, CL, AS, FPG.  
    Funding acquisition: KHMH, JAV.
    Investigation: TW, FPG.  
    Methodology: TW, IEH, YG, MTW, FPG.
    Software: TW, IEH, YG, SBT, HS, MI, CL, AS, FPG.
    Supervision: VS, JAV, FPG.
    Validation: TW, YG, SBT, HS, MI, CL, FPG.  
    Visualization: TW, FPG.
    Writing (original draft): TW, YG, FPG.
    Writing (review \& editing): all authors.

    \section*{Acknowledgements}
    The German AI Service Center WestAI provided some of the computational resources used in this work to Tassilo Wald after he left Microsoft.

    We thank Matthew Lungren for his insightful comments regarding report translation and structuring.

    %
    %
    \putbib
    \end{bibunit}

    \clearpage
    \appendix
    \acresetall
    \numberwithin{figure}{section}
    \numberwithin{table}{section}

    \crefalias{section}{appendix}
    \crefalias{subsection}{appendix}
    \crefalias{subsubsection}{appendix}

    \renewcommand{\theHsection}{apx.\Alph{section}}

    \begin{bibunit}

    \section{Development framework}
    \label{apx:development_framework}
    Due to the shortage of public 3D \visionlanguage datasets such as \ctrate~\cite{hamamci_developing_2024}, we believe the domain remains under-researched.
    We revisit key design decisions made by prior work to establish best practices for the 3D medical \visionlanguage domain and iteratively ablate design decisions related to the adaptation of CLIP to 3D. Similarly, we ablate design decisions around the inclusion of radiology report generation and vision-only self-supervised learning.
    This is all conducted on chest CT, due to the availability of a large image-only dataset (\acs{NLST}) and a large paired \imagereport dataset (\ctrate).

    \subsection{Pre-training datasets}
    \label{apx:sec:pretraining_datasets}
    \subsubsection{CT-RATE}
    \label{sec:dataset_ctrate}
    The \ctrate dataset~\cite{hamamci_developing_2024} consists of 25\ 692 non-contrast CT acquisitions with associated reports from the Istanbul Medipol University Mega Hospital.
    Each report contains a \findings section, which describes the contents of the scan, and an \impression section, which represents an interpretation of the findings given the patient's clinical history.
    The dataset is expanded to 50\ 188 unique 3D images by leveraging different reconstruction kernels for each acquisition.
    These kernels, or convolution algorithms, yield volumes with different spacings, with some reconstructions featuring high anisotropy, i.e., high in-plane resolution but low through-plane resolution, and others being less anisotropic.
    As the reconstructions stem from the same image acquisition, their information content is highly redundant.
    Therefore, for each acquisition, we choose the reconstruction with the lowest in-plane size to minimise computational cost.
    This yields a subset of {24\ 108} images, with a median spacing of \sizeaniso{0.7}{0.7}{1.0} mm and image size \sizeaniso{512}{512}{359} voxels (distribution of in-plane sizes: 22\ 417, 1648 and 43 images with size 512, 768, and 1024, respectively).

    \ctrate contains a few head CT scans, which we removed for all experiments and evaluations using the exclusion lists provided on the dataset repository.

    To prevent data leakage between pre-training and downstream datasets, we divide the official \ctrate training split into a train and a validation set.
    This split is identical to the one used in the original \ctclip paper (1k subjects with their associated reports and images)~\cite{hamamci_developing_2024}.
    The official validation split of \ctrate serves as our final test split.

    In total, our training split of \ctrate is composed of \num{18798} subjects with \num{22676} unique 3D volumes, and our validation split (extracted from the official training split) holds \num{992} subjects with \num{1197} unique 3D volumes.

    \subsubsection{NLST}
    \label{sec:dataset_nlst}
    The \ac{NLST} dataset~\cite{national2011national}  contains low-dose chest CT images from 26k patients, acquired at 33 different US centres, with each patient receiving one baseline scan and up to two follow-up scans at one-year intervals, yielding up to three scans per patient.
    Overall, this dataset provides about 72k chest CTs scans without associated reports, with two reconstruction kernels per acquisition.
    As both reconstructions have similar sizes, we randomly pick a reconstruction kernel for each acquisition, yielding our subset of NLST.

    \subsubsection{Other large datasets}
    \label{apx:sec:other_datasets}
    Aside from the aforementioned \ctrate and NLST, there are other large-scale paired \imagereport and large-scale image-only datasets in the domain, which we show in \cref{apx:tab:datasets} for additional context.

    \begin{table}[]
        \centering
        \caption{The scale of paired report-image 3D datasets is lower than the scale of large image-only 3D datasets. The number of acquisitions refers to unique scans, disregarding different CT reconstructions with multiple kernels.}
        \label{apx:tab:datasets}
        \resizebox{\linewidth}{!}{
        \input{tables/dataset_sizes}}
    \end{table}


    \subsection{Image preprocessing}
    \label{apx:sec:image-preprocessing}
    To conduct pre-training and classification experiments, we preprocessed \ctrate, NLST and \radchestct to a unified format using TorchIO~\cite{perez-garcia_torchio_2021}.
    This preprocessing consisted of \begin{enumerate}
        \item Reorientation of all images to RAS+.
        \item Dividing the CTs Hounsfield units by 1000, effectively mapping -1000 to -1 and +1000 to 1, respectively.
        \item Clipping values outside of this range.
        \item Resampling to 0.5-mm isotropic spacing, using an antialiasing Gaussian filter along any downsampled axes and B-Spline interpolation.
    \end{enumerate}
    To perform efficient loading of 3D subvolumes from cloud storage during training, we stored the preprocessed images in the NIfTI-Zarr format, using the \texttt{nifti-zarr-py}~\cite{niizarr} Python package.
    We also used \texttt{nifti-zarr-py} to generate versions of the each volume at 1-mm and 2-mm isotropic spacing, effectively storing a three-level Gaussian pyramid for each volume.
    The 1-mm- and 0.5-mm-spacing images were only used during vision-only pre-training.

    \subsection{Report preprocessing}
    \label{apx:report_preprocessing}
    \ctrate contains image--report pairs, with each report containing \findings and \impression sections, as well as other sections we did not use.
    We processed the reports using GPT-4o (\texttt{gpt-4o} version \texttt{2024-08-06} through an Azure OpenAI endpoint) as explained below.

    \subsubsection{Translating from Turkish}
    The released reports in \ctrate were originally translated from Turkish to English using the Google Translate API.
    We re-translated the reports to leverage the strong machine translation capabilities of modern \acp{LLM}, using the prompt in \cref{apx:sec:prompts:translation}.

    \subsubsection{Structuring into clinical subsections}
    We structured the \findings section by splitting it into eight clinical subsections, using the prompt in \cref{apx:sec:prompts:restructuring}:

    \begin{enumerate}
        \item \textit{Image Quality }
        \item \textit{Lungs and Airways }
        \item \textit{Pleura}
        \item \textit{Mediastinum and Hila}
        \item \textit{Cardiovascular Structures }
        \item \textit{Bones and Soft Tissues}
        \item \textit{Tubes, Lines, and Devices }
        \item \textit{Upper Abdomen}.
    \end{enumerate}

    Each sentence gets assigned in its original, long state to one of these subsections.
    Below is an example of the \textit{Lungs and Airways} subsection:
    \begin{quote}
        \textit{The trachea and both main bronchi are patent, with no obstructive pathology detected. Ventilation of both lungs is normal, and no mass or infiltrative lesion is observed. Additionally, there is a hypodense lesion measuring 15 mm in diameter located in the posterolateral middle segment of the left lung, possibly a cyst.}
    \end{quote}
    These structured reports are used for the \acf{RRG} objective.

    \subsubsection{Shortening individual findings}
    We split each subsection into short sentences representing \textit{positive} and \textit{negative} findings, using the prompt in \cref{apx:sec:prompt:positive_negative}.

    Each sentence is shortened and classified as a \textit{positive} finding (i.e., mentions the presence of an abnormality) or as a \textit{negative} finding (i.e., mentions normality).
    An example of the latter structure is shown in \cref{apx:fig:sentence_shortening} for the \textit{Bones and Soft Tissues} subsection.
    The reports of this style are used for the \shortsentence augmentation and the \acf{OSL}, which substantially improves our zero-shot classification metrics.


    \subsection{Global downstream tasks and datasets}
    \label{apx:development:eval_tasks_and_metrics}
    To measure the quality and guide development of the trained vision and text encoder, we evaluate on global (as opposed to dense) tasks, specifically image-to-report retrieval, classification probes, and zero-shot classification.
    We use \ctrate to evaluate all of these tasks, as it includes \imagereport pairs (necessary for retrieval) as well as multi-abnormality labels for the 18 most common abnormalities in the dataset (necessary for zero-shot classification and the training of probes).

    During the development phase, we quantify retrieval performance through Recall at 1, 5, and 10 (R@1, R@5, R@10), and classification performance through \textit{AUPRC} and \textit{AUROC}, guiding the hyperparameter optimisation process.
    For a detailed explanation of the metrics, we refer to \cref{apx:sec:evaluation:metrics}.

    For linear probing, we train five different sequence aggregation mechanisms with four different learning rates and a batch size of 16 for 15k steps with a cosine annealing learning rate schedule on the training split of \ctrate.
    The best performing probe is selected based on its performance on our validation set of \ctrate.
    This probe is later transferred for testing as-is to the test sets to yield the final predictions.
    More details on this are provided in \cref{apx:sec:classification_linear_probes}.

    For zero-shot classification we differentiate between two zero-shot classification schemes.
    \begin{enumerate}
        \item \textbf{Native (N):} In `native' zero-shot classification, we aggregate 50 reports of a patient with an abnormality and 50 reports of patients without this abnormality.
        Each of the long \findings is passed through the language encoder and the language pooler to yield 50 embeddings for positives and 50 embeddings for negatives.
        Each embeddings group is averaged to yield a representative embedding of the abnormality being present or absent from the reports.
        \item \textbf{Short (S):} For each abnormality, a small template is used to create sentences about whether an abnormality is present or absent.
        We use `\{abnormality\} present' and `no \{abnormality\} present', respectively.
        The resulting embedding represents the presence or absence of this abnormality.
    \end{enumerate}
    This differentiation allows us to disambiguate whether language encoders are able to handle both short-form and long-form text embeddings as inputs.


    \section{Extended methods and ablations}
    \label{apx:methods_and_experiments}
    Translating the well-established \ac{CLIP} method from the 2D imaging domain to the 3D medical domain is difficult due to the large domain gap and has been explored less because of the previously mentioned lack of publicly available data.
    In this section, we start from the basic \ac{CLIP} paradigm, ablating multiple design choices, and iteratively extend the method with additional supervision objectives, yielding our \acf{COLIPRI} encoder family (\cref{fig:colipri_diagram}).
    The final hyperparameter configurations of this process are provided in \cref{sec:apx:hyperparameters}.

    \subsection{Adapting CLIP to 3D radiology}
    \label{apx:method:optimising_clip}

    %
    Due to the large domain differences between natural imaging and 3D medical imaging, crucial training settings can vary substantially, requiring re-tuning of hyperparameters traditionally used for 2D natural images.
    To narrow the overall optimisation search space, we fix a few hyperparameters.
    Namely, we choose to train a Primus-M \ac{ViT} encoder~\cite{wald2025primus} with an AdamW~\cite{loshchilov2017decoupled} optimiser.
    Each model is trained for 250k or 125k steps with a total batch size of 8 or 16, respectively, resulting in 2 million training samples being seen by the model.
    We used 6.25k steps of linear learning rate warm-up, followed by a PolynomialLR schedule.
    We used a learning rate of $3 \times 10^{-4}$ for batch size 8 and scaled linearly it with the batch size, when applicable.
    Aside from these fixed parameters, we chose an initial hyperparameter configuration that we optimise through a star sweep~\cite{alabdulmohsin2023getting}.

    We used a pre-trained \bert model~\cite{boecking2022making} as the default text encoder due to the overlap of abnormalities between chest X-rays and chest CTs.
    By default, we pool dense vision and text tokens through a dedicated multi-head attention-pooling layer with 12 heads, trained from scratch for each encoder.
    We use the raw \findings section for supervision, and an input crop size of \sizeiso{192} at 2-mm isotropic spacing.
    Pre-training experiments are conducted on Azure Machine Learning, using a single node with 4 A100 GPUs (80 GB VRAM), unless specified otherwise.

    \begin{figure}[t]
        \centering
        \includegraphics[width=.9\linewidth]{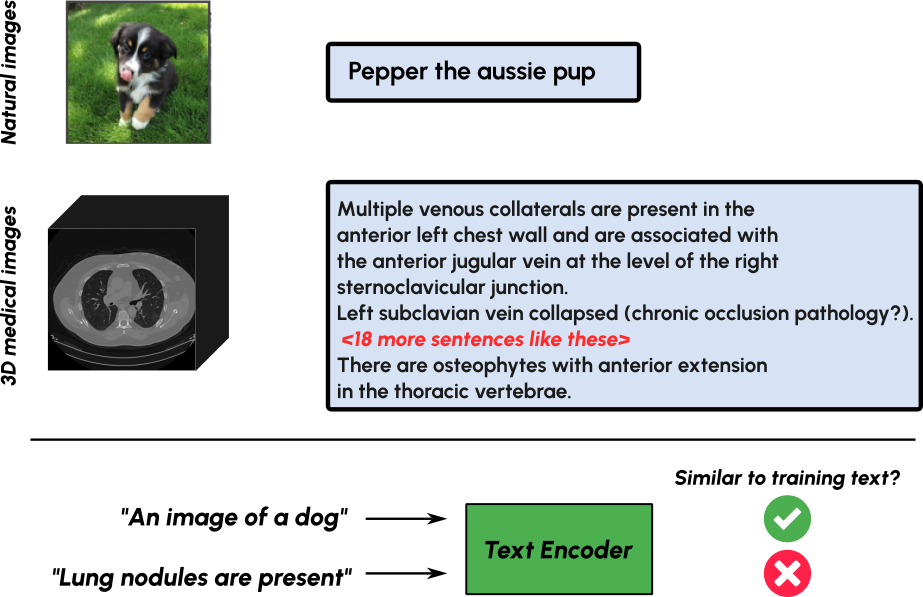}
        \caption{
            \textbf{Reports of 3D medical images are substantially longer than captions of 2D natural images}, leading to a distribution shift between long reports seen during training and short prompts used for zero-shot classification if used naively.
            Additionally, long reports might allow the text encoder to overfit due to their high dimensionality, instead of learning semantics.
            We overcome these challenges with mechanisms such as the \acf{OSL} (\cref{apx:methods:opposite_sentence_loss}) and our pipeline for report preprocessing (\cref{apx:report_preprocessing}).
        }
        \label{fig:long_report_domain_shift}
    \end{figure}

    \subsubsection{Report length}
    \label{apx:sec:report_length}
    Tokenising radiology reports results in token sequences of substantial lengths, with the average \findings section of \ctrate being 243 tokens long when using the tokeniser of \bert%
    \footnote{This is in stark contrast to Zhang \etal~\cite{zhang2022contrastive} or Radford \etal~\cite{radford2021learning}, which use a maximum sequence length of 77 tokens.
    These works either sample a single sentence from the paired text or use datasets containing single-sentence captions.} (\cref{fig:long_report_domain_shift}).  
    Consequently, training with long-form reports would inadvertently lead to a distribution shift when testing zero-shot classification with short-form prompts such as ``\{abnormality\} present''.
    Moreover, medical zero-shot classification is typically performed through negated statements (``No \{abnormality\} present.''), which may be a problem due to such statements being very sparse during training.

    To account for this, we introduce two ways of conducting zero-shot classification: `native' and `short' (\cref{apx:development:eval_tasks_and_metrics}).
    The overall shift between these two zero-shot prompting schemes is presented in \cref{tab:language_ablations}, showing a 10-pp AUROC and AUPRC gap between them for our \colorbox{gray!20}{default} CLIP configuration.
    The difference in performance between evaluation styles reveals that zero-shot classification in medical \acp{VLE} may be highly sensitive to linguistic formulation, with short diagnostic phrases (often not seen during training) yielding weaker alignment than native report-style embeddings.

    \begin{table}[t]
        \centering
        \caption{
            \textbf{
                Report augmentation ablations.
            }
            Long reports might cause the text encoder to overfit, which we mitigate by introducing \sentenceshuffle (here, `Shuffle') and \shortsentence (here, `Shorten') augmentations.
        }
        \label{tab:language_ablations}
        \resizebox{.9\linewidth}{!}{\input{tables/language_table}}
    \end{table}

    \begin{table}[t]
        \centering
        \caption{
            \textbf{Ablations related to the text used for training our \method models}.
            \textit{Impressions}: Training with impressions instead of findings;
            \textit{Findings + Impressions}: Training with findings and impressions concatenated;
            \textit{Re-translated}: Using GPT-4o to re-translate the reports from Turkish to English;
            \textit{Shuffle}: Using the \sentenceshuffle augmentation;
            \textit{DnC}: Did not Converge;
        }
        \label{tab:apx:sentence_shuffle_findings_impressions}
        \resizebox{\linewidth}{!}{\input{tables/appendix/sentence_shuffle_findings_impressions}}
    \end{table}

\begin{figure}[t]
        \centering
        \includegraphics[width=.6\linewidth]{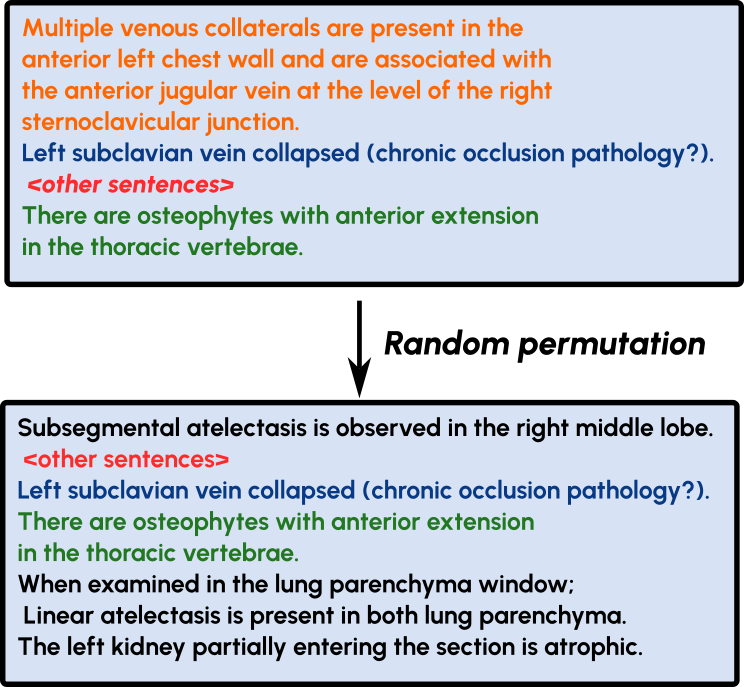}
        \caption{
            \textbf{Sentence shuffling regularises the order of sentences}, removing potential ordering biases of practitioners when writing their reports.
            Sentences are split using periods ``.'' as delimiter.
        }
        \label{apx:fig:sentence_shuffle}
    \end{figure}

\begin{figure}[!ht]
        \centering
        \includegraphics[width=.6\linewidth]{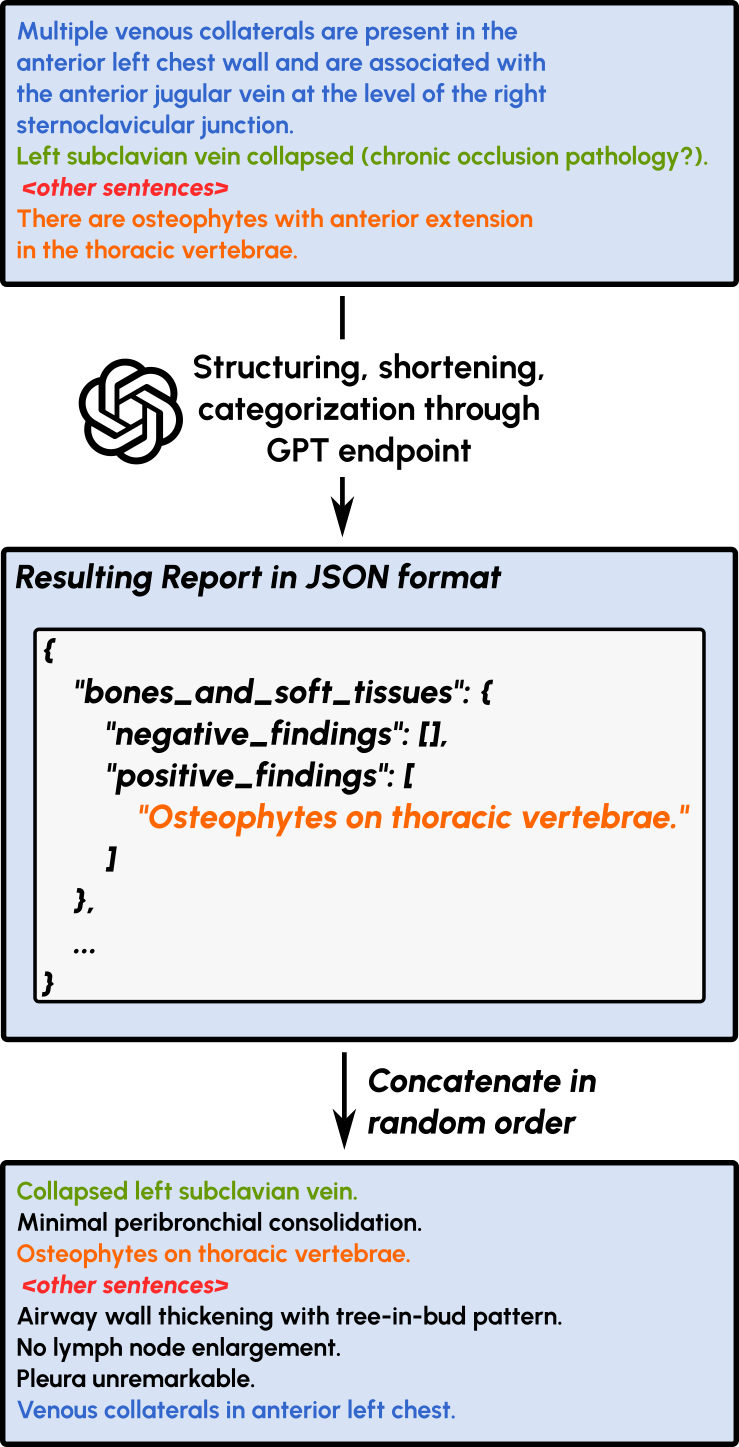}
        \caption{\textbf{Reports structuring and sentence shortening} reduce the domain shift between long reports seen during training time and short texts used for zero-shot classification.}
        \label{apx:fig:sentence_shortening}
    \end{figure}

    To minimise the distribution shift and reduce overfitting to the structure of long text reports, we introduce a \sentenceshuffle transform, which randomly shuffles sentences in the reports, substantially improving both retrieval and classification performance (\cref{apx:fig:sentence_shuffle}).
    Additionally, we introduce a \shortsentence augmentation that replaces long-form reports with a shortened version.
    These shortened reports were created using GPT-4o with instructions to reduce verbosity to a minimum (\cref{apx:report_preprocessing,apx:fig:sentence_shortening}).
    Combining \sentenceshuffle with the \shortsentence augmentations yielded further improvements in retrieval and classification  (\cref{tab:language_ablations}).
    In particular, the addition of the \shortsentence transform increases our short zero-shot classification performance considerably, reducing the gap between our `native' and `short' zero-shot classification settings (\cref{apx:development:eval_tasks_and_metrics}).

    Aside from investigating text augmentations, we evaluate the influence of training with \textit{Impressions} or \textit{Findings + Impressions}, as well as training with a translation of the original Turkish reports to English using modern methods that might reflect clinical terminology more accurately.
    Results are visualised in \cref{tab:apx:sentence_shuffle_findings_impressions}, highlighting that \textit{Findings + Impressions} is better than only \textit{Impressions}; however, using only the \findings is superior to both.
    The results also show that the newer translation does not positively affect performance, hence we do not use the original translation for the report-to-image alignment objective%
    \footnote{Our translations were still used to create the clinical subsections.}.

    Our default text encoder is \bert, a transformer pre-trained on chest X-ray reports that we fine-tune.
    Since this model is relatively small, holding about 110M parameters, and was trained on substantially shorter X-ray reports, we also evaluated the effect of using a larger text encoder.
    We tested the 196M-parameter BiomedCLIP model~\cite{Zhang2025}, which was pre-trained on the entire PubMed collection (15 million biomedical \imagetext pairs), as well as training our \bert architecture from scratch.
    Results are presented in \cref{tab:apx:sentence_shuffle_findings_impressions} at the bottom, showing the superiority of \bert over the larger BiomedCLIP model.

    \subsubsection{Field of view and number of patches}
    \label{apx:sec:field-of-view-problem}
    Radiological reports typically describe findings across the entire \ac{FOV}, rather than a restricted subregion.
    For example, chest CT reports primarily focus on pulmonary disease but often also include findings about visible abdominal organs and other incidental findings.

    In chest CT classification, the relevant abnormalities may be localised in specific lobes or distributed across the lungs.
    Therefore, the model must have access to the entire lungs \ac{FOV} to avoid missing critical context.
    Training only on subvolumes may act as a form of regularisation, but, at inference time, such models may not classify images reliably without a global \ac{FOV} since important abnormalities might lie outside the cropped region.
    Practically, this requires very large input volumes for training.
    At a resolution of 2 mm isotropic spacing, an input size of \sizeiso{192} voxels corresponds to a cube with edge 38.4 cm.
    This \ac{FOV} should suffice to cover the lungs, which are typically under 30 cm across all anatomical axes~\cite{kramer2012linear}.
    However, when using a patch size of \sizeiso{8} voxels in the \ac{ViT} (the default in Primus, as larger patches often degrade performance on high-resolution dense downstream tasks~\cite{wald2025primus}), the number of patches in a sequence reaches 14k tokens.
    This is orders of magnitude longer than typical \visionlanguage settings, where natural images tokenised at standard patch sizes yield only $(224/14)^2 = 256$ tokens~\cite{tschannen_siglip_2025}.
    This raises two fundamental questions:
    \begin{inlinelist}
        \item Are large \acp{FOV} required for good performance?
        \item How does one best aggregate this long token sequence into a global representation?
    \end{inlinelist}
    To address these issues, we evaluate the effects of varying input size, token patch size, and token aggregation strategy, as well as removing \acp{APE} to allow varying the input size at inference time (\cref{tab:field_of_view_ablations}).

    Our results indicate that smaller input sizes are beneficial, while larger inputs reduce overall performance (\cref{tab:field_of_view_ablations}, \textit{Input Size}).
    The only exception to this is the `short' zero-shot classification, where an input size of \sizeiso{128} performs worst.
    We hypothesise that smaller input sizes improve performance by forcing the vision encoder to learn more robust and semantically meaningful representations.
    Larger \ac{FOV}s expose all abnormalities simultaneously, allowing the encoder to rely on only a subset of correlated features.
    In contrast, smaller crops limit the visible context, incentivising the model to capture multiple discriminative cues.
    While this suggests that small input sizes as small as 128 may be better, an excessively small input size will discard a large amount of relevant information, limiting the applicability of the model.
    Additionally, we observe that reducing the sequence length through the use of a larger patch size has detrimental effects on overall performance (\cref{tab:field_of_view_ablations} - \textit{Patch size}), forcing us to keep the fine-grained tokens and rather long resulting sequences at the cost of higher computational resources.
    Aggregating this sequence through max-pooling proved to be the best mechanism for retrieval and linear probing.
    However, our default multi-head attention pooling proved superior for zero-shot classification and yields competitive results across all metrics (\cref{tab:field_of_view_ablations}, \textit{Token aggregation method}), while allowing one to use the MaskCLIP trick \cite{zhou_extract_2022} to generate language-aligned dense embeddings for segmentation.
    Lastly, we find that removing the \ac{APE} only negatively affects \textit{short} zero-shot classification, which we found to be the noisiest metric during our development.
    Hence, we chose to accept these minor penalties as a trade-off to allow dynamically adapting the input size (\cref{tab:field_of_view_ablations}, \textit{No \acl{APE}}).

    \begin{table}[t]
        \centering
        \caption{
            \textbf{Field of view and patch size ablations.}
            Smaller patch sizes are better, while using a large \acl{FOV} is less relevant.
            To yield global representations, max-pooling performs very well for retrieval, while multi-head attention pooling performs well across the board.
        }
        \label{tab:field_of_view_ablations}
        \resizebox{\linewidth}{!}{\input{tables/field_of_view_table}}
    \end{table}

    %
    %

    \subsubsection{Other ablations}
    \begin{table}
        \centering
        \caption{
            \textbf{Other ablations.}
            An optimal trade-off is achieved by balancing batch size with reduced training iterations.
            The sigmoid loss formulation does not improve performance.
            Lastly, minor spatial augmentations and low levels of intensity augmentation or not using intensity augmentations at all are best.
        }
        \label{tab:miscellaneous}
        \resizebox{\linewidth}{!}{\input{tables/miscellaneous}}
    \end{table}

    Contrastive learning in natural imaging benefits from large batch sizes, with e.g.\ 32k in Tschannen \etal~\cite{tschannen_siglip_2025}.
    This is far out of reach for 3D medical imaging, where batch sizes are, e.g., two when training segmentation models with \nnunet~\cite{isensee2021nnu}, due to high VRAM consumption.
    Subsequently, we ablate the trade-off between larger batch sizes versus fewer iterations, while keeping the amount of seen samples identical.
    Lastly, we evaluate whether a sigmoid loss~\cite{zhai_sigmoid_2023} improves performance compared to the default softmax loss~\cite{radford2021learning}, and determine the extent to which strong spatial and intensity augmentations are necessary.
    Results are provided in \cref{tab:miscellaneous}.
    We find mid-sized batches with fewer iterations to be superior and a decrease in performance when using the sigmoid loss.
    We also find low levels of intensity and spatial augmentations optimal, with stronger augmentations degrading performance.

    \subsubsection{Merging all changes}
    Based on our previous ablations, we introduce changes to our \colorbox{gray!20}{default} configuration.
    We \begin{inlinelist}
        \item increase the batch size to 16, in conjunction with doubling our learning rate to $6 \times 10^{-4}$ and halving our iterations to 125k,
        \item reduce the input size to 160$\times$160$\times$160,
        \item add the \sentenceshuffle and \shortsentence text augmentation using \acp{LLM},
        \item add image intensity augmentations,
        \item remove the \ac{APE} as it is not necessary, which also allows varying the input size of the model at inference time.
    \end{inlinelist}
    The final hyperparameters chosen from the prior ablations are presented in \cref{tab:apx:colipric_hyperparameters}, which, together with the \ac{OSL}, yield our \colorbox{microsoftred!20}{\colipric} encoder.

    \subsection{Including text generation}
    \label{apx:methods:generation}
    The goal of \ac{CLIP} is to align \imagereport pairs.
    This objective can be a limiting factor in the medical domain, since there may exist multiple features that differentiate two \imagereport pairs, but a single one can suffice to distinguish them.
    This key insight spurred recent works to introduce the objective of predicting the image caption from the embedding of the image encoder~\cite{tschannen2023image,wan2024locca,tschannen_siglip_2025}.
    To solve this task, the vision embeddings need to contain information about everything mentioned in the text report, as opposed to only about what differentiates two \imagereport pairs.
    Additionally, this objective is independent of the batch size, which is particularly important for a batch-size-constrained domain like 3D medical imaging.
    In this work, we combine the \ac{CLIP} objective with a \acf{RRG} objective based on CapPa~\cite{tschannen2023image}, which conducts either \textit{causal captioning} or \textit{parallel captioning} in an interleaved fashion, i.e., alternating at each training iteration, to generate a report during training.
    \textit{Causal captioning} refers to predicting the report in a next-word-prediction fashion using causal masking, while \textit{parallel captioning} predicts the entire report simultaneously from a fully masked input.

    \subsubsection{Radiology report generation for vision pre-training}
\begin{figure}[!ht]
        \centering\includegraphics[width=.7\linewidth]{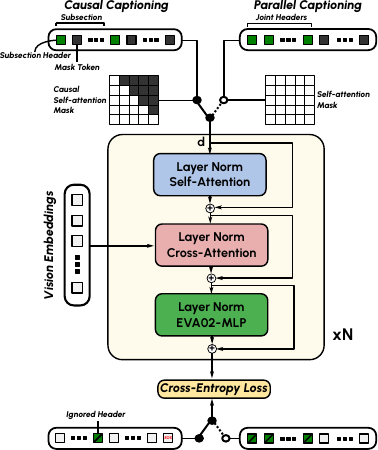}
        \caption{
            \textbf{Text generation pre-training.}
            To yield more semantic image features, we feed them through cross-attention into our text decoder EVA-02 transformer architecture, tasked with \textit{causal captioning} or \textit{parallel captioning}.
            This is optimised simultaneously with the \ac{CLIP} objective.
        }
        \label{fig:text_generator}
    \end{figure}

    As mentioned in \cref{apx:sec:report_length}, medical reports and natural image captions differ significantly, with medical reports being substantially longer and presenting a structure more akin to a list.
    These aspects can pose hurdles in \ac{RRG}, as the ordering of a listing is unpredictable without learning the preferences of the clinician who wrote the report or without memorising the entire report, both of which are undesirable.
    To address this, we structure the reports by assigning each sentence to one of eight semantic categories (\cref{apx:report_preprocessing}).
    Given these structured reports, we train our text decoder to generate the reports in a causal and a parallel fashion, but leave the clinical subsection headers unmasked to guide the generation.
    This suffices for the \textit{captioning}; however, for \textit{parallel captioning}, we expect the amount of masked tokens between two subsection headers to leak information as no causal attention mask is used.
    This is due to subsections being longer when pathological findings are present, which would allow the generative decoder to infer whether abnormalities are present given the number of masked tokens in the subsection.
    To remove this bias, we group the headers at the start of the report, followed by mask tokens.
    This informs the decoder of the desired ordering of the subsections, without leaking the length of each subsection.
    For both of our generative tasks, we shuffle the order of the subsections during training to regularise the decoder.
    Given this task formulation, we followed Tschannen \etal~\cite{tschannen2023image} and used a cross-attention-based approach to integrate vision tokens with a small transformer decoder.
    The causal and parallel text decoding setting is visualised in \cref{fig:text_generator}.

    \subsubsection{Report generation optimisations}
    \label{sec:methods:generation:optimisation}
    \begin{table}
        \centering
        \caption{
            \textbf{\ac{RRG} objective optimisations.}
            Shorter generator depth, lower \ac{RRG} loss weight, and only using parallel captioning improve performance.
            Relative to the \colipric model, the inclusion of the \ac{RRG} objective increases retrieval performance while reducing the classification performance.
        }
        \label{tab:cappa_ablation}
        \resizebox{\linewidth}{!}{\input{tables/clipcap/clipcap_joint}}
    \end{table}
    When combining the optimised \ac{CLIP} configuration with the \ac{RRG} objective, there are a few design decisions that warrant optimisation.
    By default, we choose a generator depth of 12 layers, a 50/50\% probability of causal versus parallel captioning, and a loss weight of $\lambda_{\text{RRG}} = 1$, with  $\mathcal{L}_{\text{total}}=\mathcal{L}_{\text{CLIP}} + \lambda_{\text{RRG}}\cdot \mathcal{L}_{\text{RRG}}$.
    For these parameters, we conduct another star sweep~\cite{alabdulmohsin2023getting} from our \colorbox{gray!20}{default} configuration (\cref{tab:cappa_ablation}).
    Our results indicate that always using the \textit{parallel captioning} loss is better than including also the \textit{causal captioning} objective.
    This is likely due to the next-word (\textit{causal captioning}) objective learning to memorise long reports, resulting in the vision tokens not being required for the decoding.
    Additionally, we observed that lower decoder depths are more beneficial than deeper decoders, forcing vision embeddings to be quickly adoptable for report generation.
    Lastly, we observe an intermediate loss weight of 0.3 to be optimal and found it necessary to decrease the learning rate to 3 $\times$ 10$^{-5}$ due to training instability.
    Combining these changes together with the \acf{OSL} yields the \colorbox{microsoftblue!20}{\colipricr} encoder.

    Interestingly, we observe that the additional generative decoding objective increases retrieval performance.
    However, it improves classification probing and zero-shot classification just slightly.
    Interestingly, this objective did not improve our \ac{RRG} results.
    This indicates that the current generative decoding objective may still be subject to confounders that prevent the vision encoder from learning semantically meaningful representations, which would further increase linear separation of abnormalities.

    \subsection{Including vision-only self-supervision}
    \label{apx:methods:vision_only}
    \begin{table}
        \centering
        \caption{
            \textbf{MAE optimisations.}
            Increasing the \ac{MAE} depth to 6, using block masking, and including the MAE loss only for the last 25\% of training improves performance.
            Other changes were deemed not relevant.
        }
        \label{tab:mae_ablations}
        \resizebox{\linewidth}{!}{\input{tables/clipmae/clipmae_joint}}
    \end{table}
    While \visionlanguage pre-trained encoders tend to learn useful representations for global reasoning tasks, they require paired data and their learnt representations are often less powerful for dense tasks~\cite{oquab2023dinov2}, which represent the majority of the challenges in the medical imaging community~\cite{wald2025primus}.
    To improve the quality of the learnt embeddings of our vision encoder for dense tasks, we pair our \visionlanguage pre-training with an additional \ac{MIM}, vision-only objective, yielding \colorbox{microsoftgreen!20}{\colipricm}.

    Given the recent OpenMind benchmark~\cite{wald2025openmind}, we choose \ac{MAE} as our vision-only objective, mostly due to the lack of more advanced dense pre-training methods such as \ibot, \dinovtwo, and \dinovthree in 3D medical imaging.
    To simplify integration of the vision-only objective and to save time by ablating \colipricr and \colipricm in parallel, we also integrate the \ac{MAE} pre-training with our \colipric encoder and then combine the optimised configurations of \colipricr and \colipricm to yield \colorbox{microsoftyellow!20}{\colipricrm}.

    \Ac{MAE} objectives are prevalent in the domain and often subject to high masking ratios of 60 to 90\%~\cite{munk2024amaes,wald2025revisiting}.
    These high masking ratios occlude most of the image, which we expect to hinder the CLIP task.
    To minimise this interference, we alternate between vision-only and \visionlanguage objectives for each training batch, following Tschannen \etal~\cite{tschannen_siglip_2025}.

    As the vision-only objective is not subject to the large \ac{FOV} problem (\cref{apx:sec:field-of-view-problem}), we can expose our models to sub-crops at higher resolutions and incorporate the image-only NLST dataset into our training data.

    \subsubsection{Vision-only optimisations}
    \label{sec:methods:ssl:optimisation}
    Analogous to \ac{RRG}, the inclusion of the \ac{MAE} objective introduces several factors of variation deserving ablation.
    In particular, we ablate the masking ratio, masking style (random, block and inverse block masks), mask decoder depth, vision-only loss weight $\lambda_{\text{MAE}}$, as well as ablating whether later inclusion of the vision-only objective is beneficial~\cite{tschannen_siglip_2025}.
    We also assess the inclusion of higher-resolution images, evaluating the effect of including images resampled to 1-mm and 0.5-mm isotropic spacings.
    Results are visualised in \cref{tab:mae_ablations}.

    Despite slight performance decreases in retrieval and classification, we chose to always include the vision-only objective in the training of the \visionlanguage model to improve the segmentation performance.
    Initialising the vision-encoder with \ac{MAE} pre-trained weights showed to reduce the negative effect of the \ac{MAE} objective on global tasks and improved segmentation performance substantially (\cref{apx:tab:mae_init_effect}).

    We find the masking ratio of 75\% to be optimal in combination with random masking, as well as using a vision-only loss weight of 1.
    We encountered training stability issues when training with the same learning rate, so it was reduced by a factor of 2 to 3 $\times$ 10$^{-5}$.
    Lastly, we chose to only include images of 2-mm isotropic and 1-mm isotropic spacing, finding no benefit in including images with 0.5-mm isotropic spacing.
    The final hyperparameters are presented in \cref{tab:apx:mae_hyperparameters}.


    \subsection{The Opposite Sentence Loss}
    \label{apx:methods:opposite_sentence_loss}
    The \acf{OSL} is designed to address a core limitation of \visionlanguage pre-training in the medical domain: the linguistic distribution shift between the verbose reports seen during training and the concise diagnostic prompts used at inference time.
    While radiology reports describe multiple organs and findings, often using long-form sentences with redundant context, clinical inference and zero-shot evaluation are typically conducted through short, polarity-sensitive queries such as ``\texttt{Lung nodules}'' or ``\texttt{No lung nodules}''.
    This mismatch can degrade performance because the text encoder learns to align volumetric image features with long-form report embeddings rather than compact, semantically precise statements.

    \subsubsection{Motivation and intuition}
    The \ac{OSL} explicitly trains the model to differentiate between semantically opposing statements that differ only in their diagnostic polarity.
    It encourages the joint embedding space to encode the distinction between the presence and absence of findings, improving the zero-shot classification task.
    By introducing pairs of positive and negated sentences, the \ac{OSL} exposes the text encoder to the same short, diagnostic phrases used at inference in an open-set fashion, narrowing the training--inference gap.

    \subsubsection{Related works}
    In Merlin~\cite{blankemeier_merlin_2024}, every other step of model training is performed with sentences that mention only one organ.
    Sentences not about this particular organ are removed, reducing sentence length.
    In fVLM~\cite{shui_large-scale_2025}, sentences are also assigned to particular organs, aligning the text embeddings only with the image tokens corresponding to the particular organ.
    Both of these approaches reduce the shift between training and zero-shot classification.

    Contrary to the aforementioned methods, our proposed \ac{OSL} does not require to pair sentences with a localised structure (e.g., the liver).
    Our \ac{OSL} removes the need for any explicit spatial priors.
    Consequently, it can learn semantically meaningful embeddings for short sentences even when they are not spatially grounded to a particular anatomical structure.
    Instead, it just requires the practitioner to think of any categorisation of sentences \textit{without requiring those categories to be localisable}.

    In CT-CLIP~\cite{hamamci_developing_2024}, the authors propose a fine-tuning scheme termed \textit{VocabFine}, which fine-tunes the pre-trained encoder on an `{abnormality}' `No {abnormality}' pair that is also fed through the encoder.

    While \textit{VocabFine} can appear similar to our method on first glance, there is a key distinction between the two:
    CT-CLIP's loss is \textit{closed-set}, meaning the resulting classifier is limited to the hand-selected set of abnormalities they trained on.
    This has the consequence that different abnormalities like ``Osteophytes'' which are present in the reports but not in the abnormalities are not learnt. This issue also expands to all fine-grained details that may be stated in the report, such as localisation information like ``Lung nodule \textit{in the left upper lobe}.''.
    On the other hand, our \ac{OSL} uses positive finding \textit{sentences} drawn from the report, keeping all the sentences including finer details about the pathologies a radiologist found noteworthy as training signal.

    \subsubsection{Pair construction}
\begin{figure}[!htbp]
        \centering
        \includegraphics[width=0.4\linewidth]{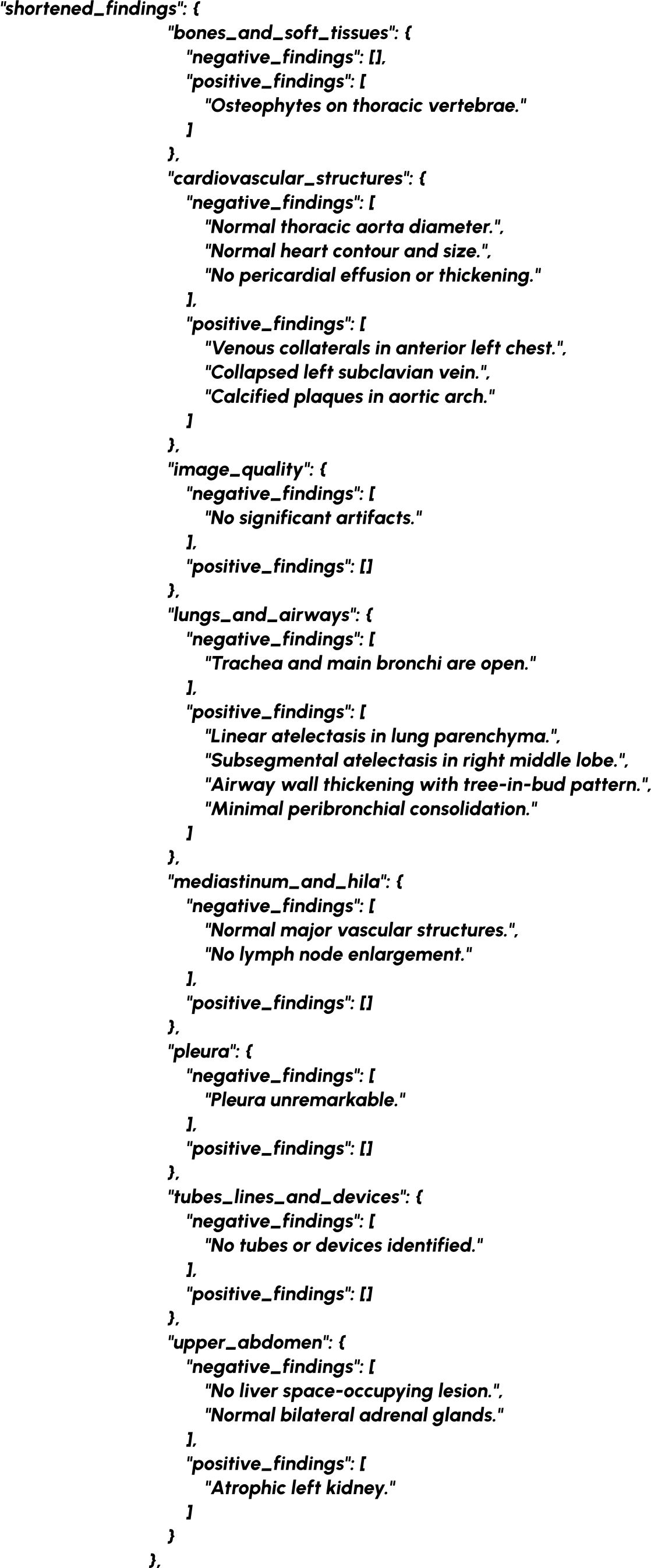}
        \caption{
            \textbf{Example of a structured, shortened and categorised report.} By having the sentences shortened, we can enable our \textit{short sentence} augmentation.
            By structuring the report into eight different semantic groups and into \textit{positive findings} and \textit{negative findings}, we enable the \acf{OSL}.
            This processing allows us to
            a) create simple negations of \textit{positive findings} creating pairs of sentences where the \textit{positive findings} should be more similar to the cases image embedding, and
            b) identify whether a case has \textbf{no} \textit{positive findings} for a certain semantic group (here \textit{image quality}, \textit{pleura}, \textit{mediastinum and hila}, \textit{tubes lines and devices}), which enables the use of \textit{positive findings} and their negation of other reports to create sentence pairs where the negation should be more similar to the image embedding.
        }
        \label{fig:apx:structured_prompt}
    \end{figure}

    Let $\mathcal{R}$ be all \acs{LLM}-preprocessed reports described in \cref{apx:report_preprocessing} (see \cref{fig:apx:structured_prompt} for an example), and let $(x_i, r_i)$ be a random image-report pair with $r_i \in \mathcal{R}$.
    Each report $r_i$ is composed of eight clinical subsections $\mathcal{C}$, with each clinical subsection $c_i = c_i^+ \cup c_i^-$ being composed of \textit{positive findings} and \textit{negative findings}, with $c_i^+ = \{s^+_1, ..., s^+_j\}$ holding $j\geq0$ \textit{positive finding} sentences $s^+$.
    Given all positives sentences in $r_i$
    \begin{equation}
        S_i^+ = \bigcup_{c \in \mathcal{C}} c_i^+
    \end{equation}
    we uniformly sample a positive sentence $s_i^+ \sim S_i^+$.
    We create a negation of $s_i^+$ via a simple negation template:
    \begin{equation}
        s_i^- = \text{``\texttt{No \{sentence\}}''}
    \end{equation}
    This yields a semantically opposing pair $(s^+_i, s^-_i)$ that differ only in diagnostic polarity.

    To ensure that opposing sentence pairs are not generated solely from subsections containing true findings in $r_i$, we additionally identify all clinical subsections $C_{zero}$ such that all $c_i^+ \in C_{zero}$ have no positive findings: $|c_i^+|= 0$.
    Given $c\in C_{zero}$ we sample random positive findings from these subsections from all reports \begin{equation}
        s^+_i \sim \bigcup_{k\in\mathcal{R}}\bigcup_{c\in C_{zero}} c_k^+.
    \end{equation}
    As before, we create a negation $s^-_i$ with the template, yielding another pair ($s_i^+, s_i^-$).
    Since this $s^+_i$ is a \textit{positive finding} that does \textit{not} originate from $r_i$, it does not provide a correct description of the image content, making its negation $s^-$ a correct statement.

    Thus, we define the binary supervision label $y_i\in\{0,1\}$ according to whether the positive statement $s_i^+$ is supported by the current image:
    \begin{equation}
    y_i =
    \begin{cases}
    1, &\text{if } s^{+}_i \in r_i, \\
    0, &\text{otherwise}.
    \end{cases}
    \end{equation}

    As such, if a positive finding originates directly from the current report, $s_i^+$ corresponds to the true observation in the image, and the correct (ground-truth) label is therefore $y_i=1$, otherwise $y_i=0$.
    This ensures that both true and counterfactual examples $(s_i^+, s_i^-)$ are presented to the model, encouraging robust discrimination from clinically valid versus invalid findings.
    Pseudocode for this \textit{Opposed Sentence Transform (OST)} algorithm is provided in \cref{alg:ost}.

    \begin{algorithm}[t]
    \caption{Opposed Sentence Transform (OST)}
    \label{alg:ost}
    \DontPrintSemicolon

    \KwIn{Structured report $r_i$ for image $x_i$; database of positive findings $S^+_{\mathrm{db}}$ grouped by subsection; number of sentence pairs $K$; negation function $\mathrm{Negate}(\cdot)$}
    \KwOut{Sentence list $T_i$ of length $2K$ and label list $Y_i$ of length $K$}

    $T_i \gets [\;]$; \quad $Y_i \gets [\;]$\;
    $S_i^+ \gets$ all positive findings in $r_i$\;
    $C_{\mathrm{zero}} \gets$ subsections of $r_i$ with no positive findings\;

    $P_{\mathrm{true}} \gets$ sample $\min(K, |S_i^+|)$ sentences from $S_i^+$\;

    $P_{\mathrm{fake\_pool}} \gets \emptyset$\ \tcp*{Set to deduplicate}
    \
    \For{each subsection $c \in C_{\mathrm{zero}}$}{
        $P_{\mathrm{fake\_pool}} \gets P_{\mathrm{fake\_pool}} \cup$ positive findings from $S^+_{\mathrm{db}}[c]$\;
    }
    $P_{\mathrm{false}} \gets$ sample $\min(K, |P_{\mathrm{fake\_pool}}|)$ sentences from $P_{\mathrm{fake\_pool}}$\;

    $n_{\mathrm{true}} \gets \min(\lceil K/2 \rceil, |P_{\mathrm{true}}|)$\;
    $n_{\mathrm{false}} \gets \min(\lfloor K/2 \rfloor, |P_{\mathrm{false}}|)$\;
    $P_{\mathrm{true}} \gets$ sample $n_{\mathrm{true}}$ from $P_{\mathrm{true}}$\;
    $P_{\mathrm{false}} \gets$ sample $n_{\mathrm{false}}$ from $P_{\mathrm{false}}$\;

    \For{each $s^+ \in P_{\mathrm{true}}$}{
        $T_i.\mathrm{append}(s^+)$\;
        $T_i.\mathrm{append}(\mathrm{Negate}(s^+))$\;
        $Y_i.\mathrm{append}(1)$\;
    }

    \For{each $s^+ \in P_{\mathrm{false}}$}{
        $T_i.\mathrm{append}(s^+)$\;
        $T_i.\mathrm{append}(\mathrm{Negate}(s^+))$\;
        $Y_i.\mathrm{append}(0)$\;
    }
    \tcp{Padding to fixed length}
    \While{$|Y_i| < K$}{
        $T_i.\mathrm{append}(\text{`` ''})$;
        $T_i.\mathrm{append}(\text{`` ''})$\;
        $Y_i.\mathrm{append}(-1)$ \tcp*{ignored in loss}
    }

    \Return{$T_i, Y_i$}\;

    \end{algorithm}

    \subsubsection{Loss formulation}
    Let $v_i$ denote the global image embedding for case $i$, and let $s_i^+$ and $s_i^-$ represent the positive and negated plain-text sentences.
    If $s_i^+$ originates from the report of case $i$, the sentence correctly describes the image, yielding label $y_i=1$.
    Should $s_i^+$ originate from another report, the statement would not apply to the image, as no positive findings have been reported for this image, hence the label of the sentence pair would be $y_i=0$.
    Using our text encoder, we create embeddings for both sentences, yielding embedding $e_i^+$ and $e_i^-$ for the corresponding positive and negated sentence embeddings, respectively.
    For each pair $(e_i^+, e_i^-)$, we compute cosine similarity scores between the image and both sentence embeddings.
    After applying a temperature scaling factor $\tau$, we convert the two similarities into probabilities using a softmax function:
    \begin{equation}
        p_i^+ =
        \frac{\exp(\text{sim}(v_i, e_i^+) / \tau)}
             {\exp(\text{sim}(v_i, e_i^+)) / \tau) + \exp(\text{sim}(v_i, e_i^-) / \tau)}
    \end{equation}
    The \ac{OSL} objective is then a binary cross-entropy loss encouraging the true embeddings to be more similar to the image embedding than the false embedding:
    \begin{equation}
    \mathcal{L}_{\text{OSL}} = - \frac{1}{N} \sum_{i=1}^{N}
    \Big[
        y_i \log p_i^+
        +
        (1 - y_i)\log (1 - p_i^+)
    \Big]
    \end{equation}
    where $N$ is the batch size.

    Due to the larger sequence length of our image tokens, doing an image forward pass is expensive.
    As the text sequence is two orders of magnitude shorter, we create eight positive and negative sentence pairs (16 sentences total) for each image, for an efficient forward pass.
    The formalisation above does not include the index for the multiple sentences per \imagereport pair for clarity in this paragraph.

    \subsubsection{Integration with CLIP}
    During multimodal training, the \ac{OSL} operates alongside the standard \ac{CLIP} contrastive loss.
    Both losses act on the same \imagetext pairs but at different levels of semantic granularity:
    CLIP aligns the overall image representation with the full report embedding, whereas \ac{OSL} enforces the similarity of a true concise diagnostic statement to be higher than its faulty opposite.
    The two are balanced equally in the alignment objective:
    \begin{equation}
    \mathcal{L}_{\text{align}} = 0.5\,\mathcal{L}_{\text{CLIP}} + 0.5\,\mathcal{L}_{\text{OSL}}.
    \end{equation}
    This joint formulation ensures that embeddings preserve both global contextual alignment and fine-grained diagnostic contrast.


    \subsubsection{\ac{CLIP}, \ac{RRG}, \ac{MAE} and \ac{OSL}}

    \begin{table}[t]
        \centering
        \caption{
            \textbf{\method development results.}
            \colipricrm and \colipricm exceed \colipricr and \colipric. In retrieval \colipricrm further exceeds \colipricm yielding a well-rounded encoder with high global task performance.
         }
        \label{apx:tab:colipri_eval}
        \resizebox{\linewidth}{!}{\input{tables/colipri_table}}
    \end{table}
    Having integrated the MAE independently with \ac{CLIP} and \ac{RRG} with \ac{CLIP}, we merge the optimal configurations (\cref{sec:methods:generation:optimisation,sec:methods:ssl:optimisation}) of \ac{CLIP} + \ac{RRG}+ \ac{MAE} without further ablations together with our \ac{OSL}, yielding our final \colorbox{microsoftyellow!20}{\colipricrm} \ac{VLE}.
    Validation results of all our final configurations are provided in \cref{apx:tab:colipri_eval}.

    \subsubsection{OSL compared to other template approaches}
    The \ac{OSL} leverages LLMs to identify, shorten, and categorise sentences about abnormalities and uses this and a simple negation template to create positive and negative pairs to train on.
    However, the loss design allows training with any positive and negative sentence pair, as long as one knows which of the two sentences applies to the current image. To evaluate the benefits and limitations of our positive--negative sentence pair generation procedure, we compare it against two alternative template-based approaches:
    \begin{enumerate}
        \item \textbf{`Abnormality'} Instead of using LLMs to identify, shorten, and categorise abnormality sentences, one can use them to extract abnormality labels directly. Based on the abnormality labels, one can use short positive and short negative templates to create positive and negative sentences to train with. Here, we use the \textit{`\{A\}'} and \textit{`No \{A\}'} template where \textit{\{A\}} represents the abnormality class, such as \textit{`Cardiomegaly'}.
        \item \textbf{`Sentences'} Reports are split using periods (\textit{`.'}) into sentences \textit{S} and paired \textit{`\{S\}'} and \textit{`No \{S\}'}. This naive strategy is naive, therefore some sentences might be linguistically incorrect. Avoiding this would require using LLMs or more advanced rule-based templates.
    \end{enumerate}
    In the abnormality-label-based template experiments, we train using either all 18 abnormality classes or only half of the classes.
    This allows us to measure the performance of this approach when all classes are in distribution (training on all 18 classes) and when one is out of distribution and wants to classify new abnormalities (training on the first nine abnormalities and evaluating on the last nine abnormalities).
    The experiment is conducted by training the final COLIPRI-CRM configuration, with just the sentence pairs of the OSL being replaced by the aforementioned alternatives.
    All configurations train with the same number ($n=8$) of positive/negative sentence pairs per image--report pair.
    For the closed-set abnormality setting, abnormalities are sampled randomly without duplicates.
    For the case of training just on the first nine, almost every batch trains with all available classes.\\

    \noindent Results are presented in \cref{tab:osl_baselines}. The naive \textbf{Sentences} configuration does not help short prompting performance.
    Its `No' prefix is exploited as a shortcut since its associated label is always 0. This shortcut is hard to avoid when one uses free-text sentences without any categorisation and without differentiating them into positive and negative sentences. The \textbf{Abnormality} label-based template using \textit{`\{A\}'}/\textit{`No \{A\}'}, improves performance substantially and exceeds our zero-shot default \ac{OSL} performance.
    Notably, its AUPRC of 61.36 and AUROC of 86.63 even exceeds the linear-probing accuracy of COLIRPRI-CRM with AUPRC 61.12 and AUROC of 86.27.
    However, upon closer inspection, this closed-set formulation does not generalise well to unseen classes.
    Specifically, supervising the first nine abnormalities and evaluating the performance on the last nine abnormalities (the not-trained-on abnormalities), the performance of this approach degrades to AUPRC of 41.26 and AUROC of 73.17, while our original \ac{OSL} formulation reaches AUPRC of 47.28 and AUROC of 78.64.
    This highlights that this closed-set formulation is limited in its performance to the included abnormalities, while performance on not-included abnormalities cannot be expected to benefit a lot from this scheme.

    \begin{table}[t]
    \centering
    \caption{\textbf{OSL training sentence variation.} Replacing the open-set positive/negative sentences with either naive sentences or label-based closed-set positive/negative sentence templates is evaluated. `Very Short' means prompting with the `\{A\}'/`No \{A\}' template. \colorbox{red!10}{Red rows} indicate evaluation on the same classes as explicitly supervised on.}\label{tab:osl_baselines}
    \resizebox{\linewidth}{!}{\begin{tabular}{lc|cc|cc|cc}
        \toprule
         & Prompt Style & \multicolumn{2}{c|}{Short} & \multicolumn{2}{c|}{Very Short} & \multicolumn{2}{c}{Native} \\
        OSL-type & Abnormality Classes & AUPRC & AUROC & AUPRC & AUROC & AUPRC & AUROC \\
        \midrule
        \cellcolor{microsoftyellow!20} & All        & 50.57 & 81.53 & 49.64 & 80.00 & 51.19 & 81.64 \\
          \cellcolor{microsoftyellow!20}       Default (COLIPRI-CRM)                     & First Nine & 53.85 & 84.41 & 52.81 & 83.05 & 52.33 & 83.83 \\
           \cellcolor{microsoftyellow!20}                            & Last Nine  & 47.28 & 78.64 & 46.47 & 76.95 & 50.05 & 79.45 \\
        \midrule
         Sentences                 & All        & 35.57 & 68.47 & 33.44 & 67.78 & 50.49 & 81.22 \\
        \midrule
        \rowcolor{red!10}                    & All        & 61.36 & 86.63 & 60.19 & 85.98 & 53.82 & 82.99 \\
        \rowcolor{red!10} Abnormality (Trained on all)  & First Nine & 61.64 & 88.00 & 60.24 & 87.50 & 53.20 & 84.30 \\
        \rowcolor{red!10}                    & Last Nine  & 61.07 & 85.25 & 60.14 & 84.47 & 54.43 & 81.68 \\
        \midrule
        \rowcolor{red!10}                    & All        & 52.25 & 80.68 & 50.36 & 78.97 & 51.99 & 82.04 \\
        \rowcolor{red!10} Abnormality (Trained on First Nine) & First Nine & 63.23 & 88.19 & 62.76 & 87.97 & 53.37 & 84.10 \\
                                             & Last Nine  & 41.26 & 73.17 & 37.95 & 69.98 & 50.61 & 79.98 \\
        \bottomrule
    \end{tabular}}
    \end{table}

    \subsubsection{OSL sensitivity to different prompt templates}
    The negated sentences used in the \ac{OSL} use a \textit{`No \{S\}'} structure. Our zero-shot prompt evaluation uses \textit{`\{A\} present'} and \textit{`No \{A\} present'}, which is similar.
    While a downstream user using these models for zero-shot classification could adhere to this provided zero-shot prompt template to optimise the model's classification performance, the question about the robustness of the model to varying zero-shot prompts remains.
    To gauge the limitations of our models to different zero-shot classification prompt templates six additional prompt templates are evaluated. Three of them follow the preceding \textit{`No ...'} structure for the negative sentence, while three more avoid this pattern in their negation.
    Specifically, the zero-shot classification prompts are:
    \begin{enumerate}
        \item \textbf{No 1:} \textit{`appearance of \{A\}'}/\textit{`No appearance of \{A\}'},
        \item \textbf{No 2:} \textit{`\{A\} is visible'}/\textit{`No \{A\} is visible'},
        \item \textbf{No 3:} \textit{`presence of \{A\}'}/\textit{`No presence of \{A\}'},
        \item \textbf{Alt 1:} \textit{`\{A\} is visible'}/\textit{`\{A\} is not visible'},
        \item \textbf{Alt 2:} \textit{`\{A\} is observed'}/\textit{`\{A\} is absent'},
        \item \textbf{Alt 3:} \textit{`evidence of \{A\}'}/\textit{`absence of \{A\}'}.
    \end{enumerate}

    Results are provided in \cref{tab:zeroshot_prompt_variation}. Across all three \textit{`No ...'} prompt-template variations, performance remains very stable, remaining within two points AUPRC and AUROC of our default short prompt style. Interestingly, the performance of prompt-template \textit{No 2} \textit{`\{A\} is visible'}/\textit{`No \{A\} is visible'}, even exceeds the performance of the short prompt template, reaching AUPRC of 51.86 and AUROC of 81.96. When straying away from the preceding \textit{`No ...'} prompt template, the performance degrades more strongly. Alt 1 and Alt 3 decrease by approximately 5 points in AUROC and AUPRC, while Alt 2 decreases by 15 points in AUROC and AUPRC.
    Overall, it can be concluded that adhering broadly to the negation template used during training is advisable and is sufficient to maintain high zero-shot classification performance. Further, the performance increases of No 2 and the much higher linear-probing results indicate that optimising the zero-shot classification prompts may lead to further increases in zero-shot classification performance. Lastly, it is advisable not to diverge too much from the sentence negation, as performance will degrade more heavily. This performance degradation indicates that the small language model we used is incapable of embedding these semantically very similar sentences into a similar text embedding. Hence, we hypothesise that more powerful, recent language embedding models, which may better capture the high semantic similarity between these different phrases, should yield greater robustness to prompt variations and improve overall zero-shot classification performance.

    \begin{table}[t]
        \centering
        \caption{\textbf{Sensitivity of COLIPRI-CRM to other zero-shot classification prompt templates.} Performance changes less when keeping a preceding \textit{`No ...'} sentence structure, while diverging from this pattern incurs higher zero-shot classification performance penalties.}\label{tab:zeroshot_prompt_variation}
        \resizebox{.8\linewidth}{!}{\begin{tabular}{lcc|ccc|ccc}
            \toprule
                   &       &        & \multicolumn{3}{c|}{Negation with `No ...'} & \multicolumn{3}{c}{Alternatives} \\
            Metric & Short & Native & No 1 & No 2 & No 3 & Alt 1  & Alt 2  & Alt 3  \\
            \midrule
            AUROC  & 81.53 & 81.64 & 80.65 & \textbf{81.96} & 79.74 & 76.73 & 65.32 & 76.88 \\
            AUPRC  & 50.56 & 51.19 & 49.16 & \textbf{51.86} & 48.83 & 45.52 & 35.37 & 44.70 \\
            \bottomrule
        \end{tabular}}
    \end{table}

    \subsubsection{LLM restructuring robustness}
    To train with the \ac{OSL}, LLMs are used to identify positive and negative findings and to map them into semantic categories. If the LLM fails to correctly identify and classify positive sentences, this introduces noise during training. While a recent study by \cite{stock2026evaluating} showed that LLMs are highly reliable at extracting classification labels from reports, we conduct an additional qualitative study to verify this in our setting.
    We randomly selected 20 findings and manually checked whether each of the 50 corresponding extracted short positive sentences could be mapped to a sentence in the original report with the same semantics. We did not evaluate whether the LLM missed any positive findings during extraction. All 50 extracted sentences (100\%) could be mapped back to the original report, indicating that the LLM extraction is highly precise.
    Two qualitative examples are provided below:\\

    \noindent \textit{``Trachea, lumen of both main bronchi are open. No obstructive pathology was detected in the lumen of the trachea and both main bronchi. Mediastinal structures were evaluated as suboptimal since the examination was unenhanced. As far as can be seen; mediastinal main vascular structures, heart contour, size are normal. Thoracic aorta diameter is normal. Pericardial effusion - no thickening was detected. Thoracic esophagus calibration was normal and no significant pathological wall thickness increase was detected in the examination limits. There are lymph nodes in the mediastinal upper-lower paratracheal, subcarinal region, with a short axis smaller than 1 cm, with a fatty hilus visible. When both lung parenchyma windows are evaluated; In the right lung upper lobe posterior, soft tissue density and \textcolor{CornflowerBlue}{\textbf{paracastricial mild bronchiectatic changes}}, \textcolor{orange}{\textbf{which are primarily evaluated in favor of parenchymal fibrosis}}, showing structural distortion and \textcolor{PineGreen}{\textbf{calcification in places}} causing volume loss are observed. \textcolor{purple}{\textbf{Nodular ground glass density increases were observed in the peripheral subpleural and peribronchovascular areas in both lungs}}. \textcolor{Periwinkle}{\textbf{The outlook was evaluated as consistent with typical-probable findings of Covid-19 pneumonia.}} Clinical and laboratory correlation is recommended. Upper abdominal organs included in the sections are normal. No space-occupying lesion was detected in the liver that entered the cross-sectional area. Bilateral adrenal glands were normal and no space-occupying lesion was detected. No lytic-destructive lesion was detected in bone structures.''}.\\
    Short sentences extracted were \textcolor{orange}{\textbf{``Parenchymal fibrosis.''}}, \textcolor{PineGreen}{\textbf{``Calcification in right upper lobe.''}}, \textcolor{CornflowerBlue}{\textbf{``Bronchiectatic changes.''}},\textcolor{purple}{\textbf{``Ground-glass opacities.''}} and \textcolor{Periwinkle}{\textbf{``COVID-19 pneumonia typical features.''}}\\

    \noindent \textit{``\textcolor{PineGreen}{\textbf{At the level of the left nipple, there is a nodular, hypodense lesion with a diameter of approximately 6 mm in the lateral part}}. Trachea, both main bronchi are open. No occlusive pathology was observed in the lumen. The ascending aorta is 49mm in diameter and the descending aorta is 35mm in diameter, and it has an aneurysmatic appearance. Wall calcifications were observed in the aorta. Heart size increased. Pericardial effusion-thickening was not observed. Thoracic esophagus calibration was normal and no significant tumoral wall thickening was detected. \textcolor{orange}{\textbf{When the lung parenchyma window was examined, pleuroparenchymal sequelae changes were observed in the bilateral upper lobe apicoposterior segments of the lung}}. \textcolor{Periwinkle}{\textbf{1 subpleural calcified nodule was observed in the middle lobe of the right lung.}} \textcolor{purple}{\textbf{A few nodules smaller than 5 mm in diameter were observed in both lungs.}} \textcolor{red}{\textbf{Subsegmental atelectatic changes were observed in the right lung middle lobe medial segment, left lung upper lobe lingula, and bilateral lower lobe basal segment of the lung.}} Liver, both adrenal glands, spleen and pancreas are normal, as can be seen on non-contrast images. Degenerative changes were observed in the bones within the sections.''}\\
    Short sentences extracted were \textcolor{orange}{\textbf{``Pleuroparenchymal sequelae in upper lobes.''}}, \textcolor{Periwinkle}{\textbf{``Calcified nodule in right lung.''}}, \textcolor{purple}{\textbf{``Multiple small nodules in both lungs.''}}, `\textcolor{red}{\textbf{``Subsegmental atelectasis in middle lobe, lingula, and lower lobes.''}} and \textcolor{PineGreen}{\textbf{``Nodular lesion near left nipple.''}}.

    \noindent While these short qualitative examples showed that extracted positive findings were highly precise, it is apparent that the recall of abnormalities is not perfect. For example ``Wall calcifications were observed in the aorta.'', which would correspond to the ``Arterial wall calcification'' label of CT-RATE, was missed in the second example. Moreover, statements that are ambiguous in their interpretation were not classified as positive sentences. For example ``Heart size increased.'' may not be cardiomegaly, yet one might interpret it as an abnormality. Similarly, ``Degenerative changes were observed in the bones within the sections'' may or may not be problematic depending on the age of the subject. This highlights the ambiguous nature of findings, and the difficulty in drawing the line when an abnormality becomes pathological and \textit{clinically significant}. We believe this ambiguity issue can likely be resolved by adapting the instruction prompt and introducing clear rules on how ambiguity should be taken into account. Regarding the sensitivity problem of missed, clear pathological findings, it is worth noting that newer LLMs have most likely improved a lot, which might solve the occasional misses.

    \subsection{Hyperparameter configurations}
    \label{sec:apx:hyperparameters}
    The final hyperparameter configurations for \colipric, \colipricr and \colipricm are provided in \cref{tab:apx:colipric_hyperparameters,tab:apx:reportgen_hyperparameters,tab:apx:mae_hyperparameters}.
    Not explicitly noted hyperparameters are `inherited' from \colipric.
    Hyperparameters of \colipricrm are the combination of \colipric, \colipricr and \colipricm with a combination of the $\mathcal{L}_{VLM}$ loss of the \colipricr and the $\mathcal{L}_{VO}$ of the \colipricm model.

    \begin{table}
        \centering
        \caption{\textbf{Hyperparameters of the \colorbox{microsoftred!20}{\colipric} encoder.}
         Num.\ warm-up steps: Steps for learning rate warm-up (deducted from total steps).}
        \resizebox{.7\linewidth}{!}{\input{tables/hyperparams_clip}}
        \label{tab:apx:colipric_hyperparameters}
    \end{table}

    \begin{table}
        \centering
        \caption{\textbf{Hyperparameters of the \colorbox{microsoftblue!20}{\colipricr} encoder.} The majority of hyperparameters is kept identical to \colipric, only changes and newly included hyperparameters are reported.
        RRG: Radiology Report Generation; OSL: Opposite Sentence Loss}
        \resizebox{.7\linewidth}{!}{\input{tables/hyperparameters_rrg}}
        \label{tab:apx:reportgen_hyperparameters}
    \end{table}

    \begin{table}
        \centering
        \caption{\textbf{Hyperparameters of the \colorbox{microsoftgreen!20}{\colipricm} encoder.} The majority of hyperparameters are kept identical to \colipric, only changes and newly included hyperparameters are reported. Loss is alternating every batch between $\mathcal{L}_{Align}$ and $\mathcal{L}_{VO}$.
        MAE: Masked Autoencoder; OSL: Opposite Sentence Loss}
        \resizebox{.7\linewidth}{!}{\input{tables/hyperparameters_mae}}
        \label{tab:apx:mae_hyperparameters}
    \end{table}

    \newpage

    \section{Evaluation details}
    \label{apx:sec:evaluation}
    \subsection{Metrics}
    \label{apx:sec:evaluation:metrics}
    \subsubsection{Retrieval metrics}
    \paragraph{Recall}
    Given image report-pairs, we evaluate the image-to-report retrieval through using Recall @ 1/5/10.
    This is calculated by embedding the entire validation set or test set of image-report pairs, yielding $N$ validation images and reports.

    This yields $N$ global image embeddings and $N$ report embeddings, for which we calculate the similarities between all pairs, identically as during the CLIP training.
    Following this, we measure whether the image embedding of the actual image is
    \begin{inlinelist}
        \item the most similar,
        \item within the five most similar or
        \item within the 10 most similar
    \end{inlinelist}
    images.
    From these results, we compute Recall@1, Recall@5, and Recall@10, which quantify the proportion of test samples for which the correct image appears among the top-1, top-5, or top-10 retrieved results, respectively.
    A higher recall value indicates that the learnt embedding space more effectively aligns visual and textual representations, allowing relevant image–report pairs to be retrieved more reliably.

    The results for report-to-image retrieval are computed analogously.

    \subsubsection{Classification metrics}
    \paragraph{Area under the receiver operating characteristic curve (AUROC)}
    The AUROC metric evaluates a model's ability to distinguish between positive and negative cases across all possible decision thresholds and is computed as the area under the curve defined by the true positive rate (sensitivity) plotted against the False Positive Rate (1 – specificity).
    An AUROC of 0.5 indicates random performance, whereas a value of 1.0 represents perfect discrimination.

    \paragraph{Area under the precision-recall curve (AUPRC)}
    The AUPRC metric measures a model's ability to identify positive cases across varying decision thresholds, emphasising performance on imbalanced datasets.
    It is computed as the area under the curve defined by precision (positive predictive value) plotted against recall (sensitivity).
    Unlike AUROC, which considers both positive and negative classes equally, AUPRC focuses on the model's effectiveness in detecting the positive class, making it particularly informative when positive cases are rare, as is the case for many abnormalities.
    A higher AUPRC indicates that the model maintains strong precision even at high recall levels, reflecting its capacity to identify true positives while minimising false detections.

    Because of this characteristics, we select the classification probe used for testing as the one that yields the highest AUPRC on the validation set.



    \newpage

    \subsubsection{Report generation metrics}
    \paragraph{RadBERT}
    RadBERT~\cite{hamamci_developing_2024} is a text classification BERT model trained on \ctrate, which allows to predict class probabilities for the 18 different multi-abnormality classes of the \ctrate dataset.
    We use Micro and Macro RadBERT $F_1$-scores to evaluate the report generation performance on top of the encoders.

    \paragraph{\mbox{RadFact (+/-)} and \mbox{RadFact (+)}}
    RadFact~\cite{bannur2024maira} assesses the factuality of a generated report by evaluating whether each sentence in the generated report is supported by a sentence in the reference (ground-truth) report and vice versa.
    This is achieved by leveraging the reasoning capabilities of GPT-4o.

    Because our data differ from the X-ray reports used in the original work, we adapt RadFact's system prompt and introduce two distinct RadFact variants: RadFact-CT (+/-) and RadFact-CT (+).
    RadFact-CT (+/-) evaluates both positive and negative radiological statements, while RadFact-CT (+) focuses exclusively on positive findings, excluding statements about the absences of abnormality, unremarkable observations, or normal anatomy.

    In this study, we employ RadFact's \textit{Logical Precision} and \textit{Logical Recall} to compute a \textit{Logical $F_1$} score.
    The grounding and spatial reasoning capabilities of RadFact are not considered in our evaluation.

    \subsection{Classification linear probing}
    \label{apx:sec:classification_linear_probes}
    Given a pre-trained encoder, we conduct linear probing to measure the quality of our vision encoders embeddings for classifying the abnormalities labelled in \ctrate and \radchestct.
    To do so, we discard the original token aggregation scheme of the vision encoder, which was aimed at aligning image and report, and instead train a new one for classification.
    As the best token aggregation scheme is unknown, we conduct a grid search over five different schemes and four different learning rates.
    The token aggregation schemes are:
    \begin{enumerate}
        \item \textbf{Average pooling}: a simple averaging across the sequence dimension.
        \item \textbf{Max pooling}: a simple max-pooling across the sequence dimension.
        \item \textbf{Learnt attention pooling}: an attention pooling head with a learnt query token, steering how the tokens are recombined to yield the final global representation.
        \item \textbf{Average attention pooling}: same as above but with the learnt query replaced by a token created through average pooling.
        \item \textbf{Multi-learnt attention pooling}:
        As learnt attention pooling but with four learnt query tokens instead of one. As we get one representation for each query, the four outputs are averaged to yield the global representation.
    \end{enumerate}
    All of these token pooling schemes yield a global embedding which we project down through a linear layer to the 18 abnormalities annotated in \ctrate.

    The four learning rates over which we sweep are $lr \in \{3\times 10^{-3}, 1\times10^{-2}, 3\times10^{-2}, 1\times10^{-1}, \}$.
    As we keep the encoder frozen, we can allocate the majority of VRAM to the probes, allowing us to train all of them jointly~\cite{oquab2023dinov2}.

    The probes are trained using a batch size of 16, for 12.5k steps, using an SGD optimiser with momentum 0.95 and no weight decay, following a cosine annealing learning rate schedule.
    As the input volumes are larger than the input size, we conduct centre-cropping of the volume, extracting a central $s\times s \times s$ crop (the same as the vision encoder's input size $s$).
    Once training concluded, the best probe is selected based on the \ac{AUPRC} values on the validation set. 
    When using the probes for testing, the best probe on the validation set is used for evaluation on the test set, yielding the final metrics.

    \subsection{Zero-shot classification}
    Opposed to the trained classification probes of \cref{apx:sec:classification_linear_probes} the originally trained multi-head attention pooling as well as the language-encoder and language-pooling is reused to evaluate zero-shot classification performance.
    In this paper we differentiate between two zero-shot classification schemes: `native' and `short' (\cref{apx:development:eval_tasks_and_metrics}).

    Given the language embeddings representing the presence or absence of an abnormality, a global vision embedding is extracted from a centre-crop of each image.
    For each of these global vision embedding the cosine similarity between the vision and the two language embeddings is calculated, the similarities are temperature-scaled (divided by 0.07), and the resulting logits are fed through a softmax to yield probabilities associated with the presence and absence of the abnormality.
    The probability associated with the positive embedding is used to calculate the metrics as in \cref{apx:sec:classification_linear_probes}.

    \subsection{Report generation}
    \label{apx:sec:generative_decoding}
    We tested the potential of our patch embeddings on the \visionlanguage task of report generation.
    For the language component, we used the Qwen2.5-1B base model \cite{team2024qwen2}, which was not instruction-tuned to ensure fair evaluation of intrinsic alignment.

    Our training recipe adheres to the LLaVA-style framework~\cite{liu2023visual}, where a canonical frozen vision encoder and trainable decoder paradigm is used for multimodal \visionlanguage generation.
    The 3D vision backbone remains frozen throughout training to preserve pre-learnt visual representations.
    On top of this encoder, we train both a cross-modal alignment module and the language decoder.
    We employ a causal language modelling loss with teacher forcing, applied to tokenised radiology reports.
    The optimisation objective is thus purely autoregressive, and no auxiliary objectives are introduced.

    Following prior work~\cite{liu2023visual,perez-garcia_exploring_2025}, we integrated vision tokens into the language space through a two-layer \ac{MLP} projection head.
    We constrain supervision to the \findings section of the CT report.
    The \findings provide high-density, structured clinical interpretation of the CT volume, covering organ-level abnormalities and radiographic evidence.
    In contrast, the \impression section, although often used in clinical practice, introduces redundancy without providing additional information that could be extracted from the input image.
    We therefore omit the \impression section in all experiments.

    Each \ac{LLM} is fine-tuned on the \ctrate training set with a batch size of 32 for 10 epochs, with no weight decay.
    The maximum learning rate is $5\times10^{-5}$ and a cosine learning rate schedule is used with a linear warm-up for 3\% of the training steps.
    These hyperparameters were selected to maximise the RadBERT Micro-$F_1$ scores on the validation set.

    \subsection{Retrieval}
    Multimodal retrieval evaluate how well a model aligns visual and text representations in a shared embedding space.
    By retrieving the correct clinical report given an image (image-to-report retrieval), we directly measure whether the model captures clinically meaningful visual semantics and associates them with corresponding textual descriptions.
    In a clinical context, this task could be leveraged for \ac{RAG} of reports.
    Analogously, retrieving an image given a report assesses similar model capabilities and could be used to visualise multiple manifestations of the reported abnormalities.
    Although we do not deal with unimodal retrieval in this paper (i.e., image-to-image or report-to-report), this task might also be of interest for certain applications.

    Retrieval thus serves as a strong proxy for multimodal understanding and \visionlanguage alignment.
    As the pre-training objectives include \imagereport alignment, no additional adaptation step is required for this task.
    Hence, the vision and text encoders with their respective pooling mechanisms are used as-is to evaluate this task.

    \subsection{Segmentation fine-tuning}
    To evaluate segmentation performance, we leverage the pre-training adaptation framework proposed in \texttt{nnSSL} \cite{wald2025openmind}, which introduces fine-tuning of pre-trained vision encoders into the well-established nnU-Net framework\cite{isensee2021nnu}.
    In particular, a longer training schedule of 1000 nnU-Net pseudoepochs (250k iterations) and a shorter training schedule of 150 nnU-Net pseudoepochs (37.5k iterations) were proposed in this paper.
    We leverage both to evaluate the embedding quality of our vision encoders, with details on the explicit settings available in Wald \etal~\cite{wald2025openmind} and the \texttt{nnSSL} repository%
    \footnote{\url{https://github.com/MIC-DKFZ/nnssl}}.

    \paragraph{Segmentation dataset preprocessing}
    During pre-training we trained our vision encoder on CT data that was mapped from $[-1000,1000]$ to $[-1,1]$ and clipped to $[-1,1]$.
    However, in initial tests, we found this to yield sub-par results for semantic segmentation.
    Consequently, we stick to the official nnU-Net normalisation, referred to as \texttt{CTNormalisation}, which clips values outside the 0.5th percentile and the 99.5th percentile before standardising to zero mean and unit variance (the standardisation statistics are computed on the whole dataset).
    Moreover, despite the majority of encoders trained on 2-mm isotropic spacing (with the exception of some MAEs), we chose to resample the segmentation datasets to 1-mm isotropic spacing, as this resolution is substantially closer to the median spacings the downstream datasets come with.
    We note that, while this shift in normalisation and spacing is not optimal and may negatively influence segmentation results, this setting still allows us to compare the performance of different models.
    While the spacing issue is not easily avoidable, the normalisation choice could be adapted easily in future work by preprocessing the pre-training data identically.


    \section{Extended results}
\begin{figure*}
        \centering
        \label{fig:apx:reportgen-abnormalitywise}
        \resizebox{\linewidth}{!}{
        \includegraphics{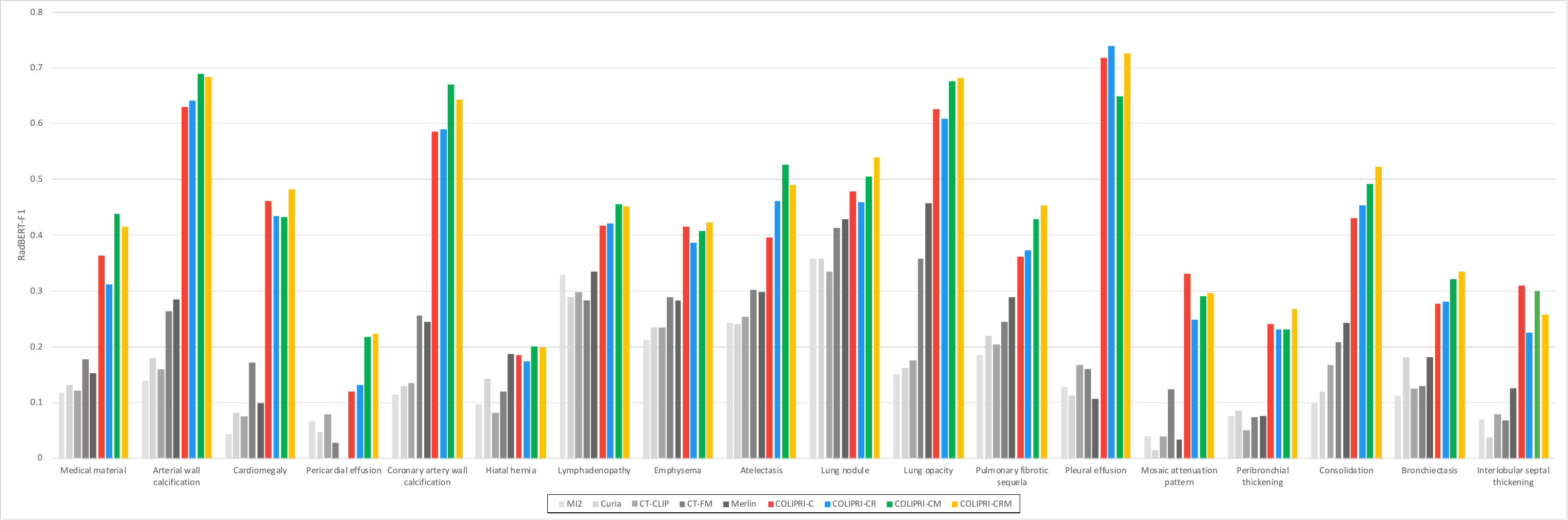}
        }
        \caption{
            \textbf{Abnormality-wise report generation results.}
            Additional to the macro and micro $F_1$-Scores across abnormalities (\cref{tab:apx:results_generation}), we report abnormality-wise RadBERT-$F_1$ scores.
        }
    \end{figure*}

    In addition to the results presented in the main manuscript, we provide extended result tables displaying:
    \begin{enumerate}
        \item Zero-shot classification results in \cref{tab:apx:classification_zeroshot}.
        \item Report generation results in \cref{tab:apx:results_generation}.
        \item Report generation results for each abnormality in \cref{fig:apx:reportgen-abnormalitywise}.
        \item Segmentation results of all folds in \cref{tab:apx:segmentation_results}.
        \item Segmentation results on OOD datasets (ACDC~\cite{bernard2018deep}, MAMA-Mia~\cite{Garrucho2024MAMAMIA}, HNTS-MRG~\cite{HNTSMRG}) in \cref{tab:apx:segmentation_generalization_results}. All of these datasets use an MRI modality and have a different field of view than a chest CT.
        \item Image-to-report and report-to-image retrieval including mean and median ranks in \cref{tab:apx:retrieval_test_extended}.
        \item Segmentation and classification results highlighting the importance of initialising with MAE pre-trained weights \cref{apx:tab:mae_init_effect}.
t
    \end{enumerate}


    \begin{table*}
        \centering
        \caption{
            \textbf{Additional zero-shot classification results} complementing \cref{tab:results_and_discussion:zeroshot_classification_main}.
            When conducting zero-shot classification with long-form ``native'' prompts, performance is slightly higher, yet within confidence intervals for our encoders.
            Overall, \colipricrm and \colipricm reach highest performance.
            \textbf{Bold} indicates best performance for that metric, or overlapping \acp{CI} with best.
            Prompt: prompt style;
            AUPRC: area under precision recall curve;
            AUROC: area under receiver operating characteristic curve.
            $^*$Values taken from Shui \etal~\cite{shui_large-scale_2025}.
        }
    \label{tab:apx:classification_zeroshot}
    \resizebox{\linewidth
    }{!}{\input{tables/results/zeroshot_classification_appendix}}
    \end{table*}

    \begin{table*}
        \centering
        \caption{
            \textbf{Complete report generation results}.
            We compare the embedding quality of our pre-trained vision encoders against available baselines when used for \acf{RRG}.
            Across most metrics, our \method models exceed the baselines.
            In particular, we exceed the baselines by $>$17 points as measured by RadBERT Macro$F_1$ and by $>$7 points when focusing on sentences about pathological findings, as measured by \radfactctp{}.
            Note that RadFact-CT/Logical $F_1$ (+/-) relies aggressively on negative findings, which are very frequent and less important than positive findings (\cref{sec:result_and_discussion:reportgen}).
            We report median and 95\% \acp{CI} based on 500 bootstrap samples.
            \textbf{Bold} indicates best performance for that metric, or overlapping \acp{CI} with best.
        }
        \label{tab:apx:results_generation}
        \resizebox{\linewidth}{!}{\input{tables/results/report_gen}}
    \end{table*}

    \begin{table*}
    \centering
        \caption{
            \textbf{Extended segmentation results}.
            We report the \acf{DSC} of our pre-trained encoders and baselines for each fold of a five-fold cross-validation on all datasets.
        }
        \label{tab:apx:segmentation_results}
        \resizebox{\linewidth}{!}{\input{tables/results/segmentation}}
    \end{table*}

\begin{table*}[t]
  \centering
  \caption{Five-fold cross-validation results on three OOD segmentation datasets. All values are reported in percent; the best result in each column is in \textbf{bold}. \textbf{ACDC}: Cardiac Cine-MRI \cite{bernard2018deep}; \textbf{MAMA-Mia}: Breast cancer segmentation on DCE-MRI \cite{Garrucho2024MAMAMIA}; \textbf{HNTS-MRG}: Head and Neck tumor segmentation on MRI \cite{HNTSMRG}.}
  \label{tab:apx:segmentation_generalization_results}
  \resizebox{\textwidth}{!}{%
  \begin{tabular}{l ccccc|c| ccccc|c |ccccc|c}
    \toprule
    & \multicolumn{6}{c}{\textbf{ACDC}}
    & \multicolumn{6}{c}{\textbf{MAMA-Mia}}
    & \multicolumn{6}{c}{\textbf{HNTS-MRG}} \\
    \cmidrule(lr){2-7} \cmidrule(lr){8-13} \cmidrule(lr){14-19}
    \textbf{References} (250k steps)
    & Fold 0 & Fold 1 & Fold 2 & Fold 3 & Fold 4 & Avg.
    & Fold 0 & Fold 1 & Fold 2 & Fold 3 & Fold 4 & Avg.
    & Fold 0 & Fold 1 & Fold 2 & Fold 3 & Fold 4 & Avg. \\
    \midrule
    nnU-Net def.
    & 92.43 & 90.24 & 91.95 & 91.44 & 91.72 & 91.56
    & 78.41 & \textbf{77.05} & 79.16 & \textbf{79.53} & 77.47 & 78.32
    & 66.54 & 63.73 & 68.28 & 61.27 & 64.96 & 64.96 \\
    nnU-Net ResEnc-L
    & \textbf{93.05} & 92.11 & \textbf{92.60} & 92.76 & \textbf{92.68} & 92.64
    & \textbf{79.68} & 76.78 & 80.85 & 78.82 & \textbf{78.88} & \textbf{79.00}
    & \textbf{69.22} & 62.43 & \textbf{69.02} & 59.98 & \textbf{65.48} & \textbf{65.22} \\
    \midrule
    \multicolumn{19}{l}{\textbf{Primus-M} (37.5k steps)} \\
    \quad Scratch
    & 92.11 & 92.09 & 91.94 & 92.55 & 91.88 & 92.11
    & 71.59 & 72.51 & 73.58 & 73.56 & 71.92 & 72.63
    & 57.91 & 55.41 & 58.44 & 60.28 & 58.74 & 58.16 \\
    \rowcolor{microsoftyellow!20} \quad COLIPRI-CRM
    & 93.02 & \textbf{93.00} & 92.58 & \textbf{93.00} & 92.50 & \textbf{92.82}
    & 72.98 & 73.17 & 75.68 & 75.66 & 71.95 & 73.89
    & 65.78 & 63.72 & 67.38 & \textbf{63.38} & \textbf{65.48} & 65.15 \\
    \midrule
    \multicolumn{19}{l}{\textbf{Primus-M} (250k steps)} \\
    \quad Scratch
    & 92.50 & 92.01 & 91.86 & 92.36 & 91.43 & 92.03
    & 77.49 & 76.45 & 78.95 & 78.11 & 76.39 & 77.48
    & 62.39 & 56.84 & 60.83 & 59.66 & 58.76 & 59.70 \\
    \rowcolor{microsoftyellow!20} \quad COLIPRI-CRM
    & 92.57 & 92.28 & 92.30 & 92.44 & 91.65 & 92.25
    & 79.34 & 76.90 & \textbf{81.13} & 79.05 & 78.18 & 78.92
    & 66.37 & \textbf{63.90} & 65.60 & 61.93 & 63.36 & 64.23 \\
    \bottomrule
  \end{tabular}
  }
\end{table*}

    \begin{table*}
        \centering
        \caption{
            \textbf{Extended retrieval results.}
            Additional retrieval results on the CT-RATE test set.
            We present the report-to-image retrieval results after deduplication of duplicate reports and removal of head CT cases ($N$ = 1493), and image-to-report retrieval after removal of head CTs ($N$ = 1551).
        }
        \label{tab:apx:retrieval_test_extended}
        \resizebox{.6\linewidth}{!}{\input{tables/results/test_retrieval_extended}}
    \end{table*}

    \begin{table*}[]
        \centering
        \caption{
            \textbf{Effect of MAE initialisation.}
            MAE initialisation and continued MAE training drives segmentation performance.
            When not initialising the \colipricm vision encoder with MAE pre-trained weights, \colipricm is unable to achieve reach segmentation performance.
            Instead it yields similar performance as the \colipric encoder, which trains without the MAE objective.
        }
        \label{apx:tab:mae_init_effect}
        \resizebox{.6\linewidth}{!}{\input{tables/appendix/segmentation_mae_init}}
    \end{table*}

    \onecolumn
    \section{Prompts and exemplary reports}
    \subsection{Prompt to translate reports from Turkish to English}
    \label{apx:sec:prompts:translation}
    \begin{adjustbox}{max width=.6\textwidth, max totalheight=.7\textheight, valign=T}
    \begin{minipage}{\linewidth}
    \begin{verbatim}
    """You are a board-certified radiologist-translator.
    Translate the Turkish radiology report contained inside a single <report> … </report> element into fluent, precise English.

    ###############################################################################
    ##  OUTPUT — COPY THIS SHAPE EXACTLY
    ###############################################################################
    **1. Clinical Information**
    English text here.

    **2. Technique**
    English text here.

    **3. Findings**
    English text here.

    **4. Impression**
    English text here.

    • **The four numbered headings must stay exactly as above and remain in bold.**
    • If any section is empty, whitespace, or literally “nan”, write Not provided. (plain text, **not** bold) under that heading.
    • Do **NOT** output anything outside these four labelled sections.
    • No bullet characters (•, –, *, etc.) or markdown lists inside the body text.

    ###############################################################################
    ##  INPUT
    ###############################################################################
    You will receive one well-formed XML block:
    <report>
      <clinical_information>…</clinical_information>
      <technique>…</technique>
      <findings>…</findings>
      <impression>…</impression>
    </report>

    ###############################################################################
    ##  STYLE RULES
    ###############################################################################
    • Literal, complete translation — no omissions, additions, or summaries.
    • Concise, objective radiology tone (passive voice preferred).
    • Use RSNA / ACR terminology; convert decimal commas to periods (7,5 mm → 7.5 mm).
    • Expand abbreviations on first mention: “CT pulmonary angiography (CTPA)”.
    • Preserve original sentence order and punctuation.

    ###############################################################################
    ##  REQUIRED GLOSSARY — replace the Turkish term with the English term verbatim
    ###############################################################################
    buzlu cam görüntüsü             → ground-glass opacity
    plevral efüzyon                 → pleural effusion
    septal kalınlaşma               → interlobular septal thickening
    konsolidasyon                   → consolidation
    akciğer nodülü                  → pulmonary nodule
    retiküler opasiteler            → reticular opacities
    bronşiektazi                    → bronchiectasis
    hiler lenfadenopati             → hilar lymphadenopathy
    mediastinal şift                → mediastinal shift
    trakea orta hatta               → trachea is midline
    perikardiyal efüzyon            → pericardial effusion
    şüpheli kitle                   → suspicious mass
    subplevral bant                 → subpleural band
    havayolu duvar kalınlaşması     → airway wall thickening
    lenf bezi büyümesi              → lymph-node enlargement
    ateşli infiltrasyon             → inflammatory infiltration
    atelektazi                      → atelectasis
    bal peteği görünümü             → honeycombing pattern
    fibrotik değişiklikler          → fibrotic changes
    amfizem                         → emphysema
    tomurcuklanmış ağaç             → tree-in-bud pattern
    kontrastsiz                     → non-contrast enhanced
    kontrast verilmeden             → non-contrast enhanced
    """
    \end{verbatim}
    \end{minipage}
    \end{adjustbox}
    \newpage

    \subsection{Prompt to structure \findings sections into different subsections}
    \label{apx:sec:prompts:restructuring}

    \begin{adjustbox}{max width=.6\linewidth, valign=t}
    \begin{minipage}{1.1\linewidth}
    \begin{verbatim}
    """You are a radiology report editor.
    Restructure a non-contrast chest CT report (supplied in four free-text blocks)
    into the fixed template below **without altering a single medical fact**.
    ───────────────────────────────  INPUT  ────────────────────────────────
    The incoming text always uses these bold labels:
    **Clinical Information:** …
    **Technique:** …
    **Findings:** …
    **Impression:** …
    ──────────────────────────────  OUTPUT  ────────────────────────────────
    Copy this skeleton exactly. Section and subsection titles must be **bold** and end
    with a colon. After each colon insert one space, then the content or the fallback line.
    **1. Clinical Information:**
    …
    **2. Technique:**
    …
    **3. Comparison:**
    …  ← If prior imaging referenced; else: No prior imaging available for comparison.
    **4. Findings:**
    **4.1 Image Quality:**
    …  ← If no limitations: Diagnostic image quality. No significant artifacts noted.
    **4.2 Lungs and Airways:**
    …  ← If no pulmonary findings: No pulmonary abnormalities detected.
    **4.3 Pleura:**
    …  ← If no pleural findings: Pleura unremarkable.
    **4.4 Mediastinum and Hila:**
    …  ← If no findings: Mediastinal and hilar structures unremarkable.
    **4.5 Cardiovascular Structures:**
    …  ← If no findings: Cardiovascular structures unremarkable.
    **4.6 Bones and Soft Tissues:**
    …  ← If no findings: No osseous or soft-tissue abnormalities detected.
    **4.7 Tubes, Lines, and Devices:**
    …  ← If none present: No tubes or devices identified.
    **4.8 Upper Abdomen:**
    …  ← If unremarkable or not imaged: No upper-abdominal abnormalities detected.
    **5. Impression:**
    …  ← If missing: No impression provided.
    ───────────────────────────  EDITING RULES  ────────────────────────────
    • Zero-omission: every medical statement from the original “Findings” and “Impression”
      MUST reappear once (and only once) in an appropriate subsection.
    • Do not add, delete, combine, or reinterpret abnormalities.
    • Re-phrase into concise, passive radiology English (RSNA/ACR style).
    • If a section/subsection is entirely absent, insert the exact fallback line.
    • No lists, bullets, metadata, or commentary—return only the final formatted report.
    After drafting, mentally cross-check that every clinical phrase from the original is present.
    Begin when you receive the four-block input.
    """
    \end{verbatim}
    \end{minipage}
    \end{adjustbox}
    \newpage

    \subsection{Prompt to extract positive and negative findings}

    \label{apx:sec:prompt:positive_negative}
    \begin{adjustbox}{max width=.6\textwidth}
    \begin{minipage}{1.1\textwidth}
    \begin{verbatim}
    You are an AI assistant that makes radiology reports more succinct. These reports are being used
    to train a 3D CLIP-style deep learning model. You will be given the full findings section.
    You will extract, for each of the 8 sections in the findings text, a list with negative findings
    and a list with positive findings.

    In the first list, you must collect a summarized sentence for each negative finding mentioned.
    For example, a sentence like "Esophagus is within normal limits.  In the sections passing through
    the upper part of the abdomen, the bilateral adrenal glands appear natural. No significant
    pathology was detected in the abdominal sections." must be mapped to a list like  ["Normal
    esophagus.", "Natural bilateral adrenal glands.", "No abdominal pathologies."]. The exact sentences
    must be short but maintain their core message. Positive findings are not allowed in this list
    and have to be ignored.

    In the second list you must summarize only the positive findings that are denoted.
    In this version sentences like 'The heart and mediastinal vascular structures have
    a natural appearance', 'Esophagus is within normal limits.', 'No occlusive pathology
    was detected in the trachea and both main bronchi.' or 'Trachea and main bronchi are open.'
    have to be left out. When positive (pathological) findings are mentioned, summarize them
    very briefly. E.g. a sentence like 'atypical infiltration areas of septal thickenings are
    observed in places' can be summarized as 'Septal thickenings.'. Similarly as before create
    a list of short sentences about positive abnormalities ["Septal thickenings.", "Multiple lung
    nodules.", ...]. Make sure the sentences you create are a statement and less of a description,
    like how someone would search for the case as opposed to how one would describe it
    in a findings report.

    Ignore all information that cannot possibly be predicted from the corresponding single image
    or provided clinical information section. Any comparison or reference to prior imaging must
    be ignored from the output. Do not output findings about how the image was acquired.

    Output this in JSON format with one key for each of the eight sections. Each section is a mapping
    from section name (e.g. "image quality" or "cardiovascular structures") to the "negative findings"
    and "positive findings" lists. This is the structure:

    {
      "image_quality": {
        "negative_findings": [
          ...
        ],
        "positive_findings": [
          ...
        ],
      "lungs_and_airways": {
        ...
        },
      ...
      }
    }
    \end{verbatim}
    \end{minipage}
    \end{adjustbox}
    \newpage
    \subsection{Exemplary structured, shortened, categorised report}
    \label{sec:apx:short_report}

    \newpage

    \putbib
    \end{bibunit}

\end{document}

%% file: setup/packages.tex
\usepackage{amsmath,amsfonts,bm}
\usepackage[nolist,nohyperlinks]{acronym}
\usepackage{url}
\usepackage{graphicx}
\usepackage{subcaption}
\usepackage{booktabs}
\usepackage[table]{xcolor}
\definecolor{lightgray}{gray}{0.9}
\usepackage{subcaption}    
\usepackage{multirow}
\usepackage{enumitem}
\usepackage{lineno}  
\usepackage{xspace}
\usepackage{siunitx}
\usepackage{adjustbox}
\usepackage[utf8]{inputenc} 
\DeclareUnicodeCharacter{2500}{-} 
\usepackage{duckuments}  
\usepackage[ruled,vlined]{algorithm2e}
\usepackage{microtype}  


%% file: setup/commands.tex
\newcommand{\bert}{CXR-BERT\xspace}

\newcommand{\ctclip}{CT-CLIP\xspace}
\newcommand{\ctrate}{CT-RATE\xspace}
\newcommand{\dinovtwo}{DINOv2\xspace}
\newcommand{\dinovthree}{DINOv3\xspace}
\newcommand{\ibot}{iBOT\xspace}
\newcommand{\llava}{LLaVA\xspace}
\newcommand{\medimageinsight}{MedImageInsight\xspace}
\newcommand{\nnunet}{nnU-Net\xspace}

\newcommand{\qwen}{Qwen2.5\xspace}
\newcommand{\radchestct}{RAD-ChestCT\xspace}
\newcommand{\sigliptwo}{SigLIP~2\xspace}
\newcommand{\radfactctpm}{RadFact-CT ($+/{-}$)}
\newcommand{\radfactctp}{RadFact-CT (+)}

\newcommand{\radfactctpf}{RadFact-CT (+)/$F_1$}

\newcommand{\modelurl}{\url{https://huggingface.co/microsoft/colipri}}

\newcommand{\visionlanguage}{vision\nobreakdash--language\xspace}

\newcommand{\imagereport}{image\nobreakdash--report\xspace}
\newcommand{\imagetext}{image\nobreakdash--text\xspace}

\newcommand{\findings}{\textit{Findings}\xspace}
\newcommand{\impression}{\textit{Impression}\xspace}

\newcommand{\shortsentence}{\textit{Short Sentence}\xspace}
\newcommand{\sentenceshuffle}{\textit{Sentence Shuffle}\xspace}

\newcommand{\method}{\ac{COLIPRI}\xspace}
\newcommand{\methodshort}{\acs{COLIPRI}\xspace}

\newcommand{\sizeiso}[1]{#1 $\times$ #1 $\times$ #1}
\newcommand{\sizeaniso}[3]{#1 $\times$ #2 $\times$ #3}

\newcommand{\colipric}{COLIPRI-C\xspace}
\newcommand{\colipricm}{COLIPRI-CM\xspace}
\newcommand{\colipricr}{COLIPRI-CR\xspace}
\newcommand{\colipricrm}{COLIPRI-CRM\xspace}

\definecolor{microsoftblue}{HTML}{2196F3} 
\definecolor{microsoftyellow}{HTML}{FFC107}
\definecolor{microsoftred}{HTML}{F44336}
\definecolor{microsoftgreen}{HTML}{4CAF50}

\newcommand{\ci}[1]{\textcolor{gray!60}{\scriptsize #1}}

\newlist{inlinelist}{enumerate*}{1}
\setlist*[inlinelist,1]{label=\arabic*)}

\def\eqref#1{equation~\ref{#1}}

\def\1{\bm{1}}

\DeclareMathAlphabet{\mathsfit}{\encodingdefault}{\sfdefault}{m}{sl}
\SetMathAlphabet{\mathsfit}{bold}{\encodingdefault}{\sfdefault}{bx}{n}



%% file: setup/acronyms.tex
\begin{acronym}
    \acro{APE}{absolute positional encoding}
    \acro{AUROC}{area under the receiver operating characteristic curve}
    \acro{AUPRC}{area under the precision-recall curve}
    \acro{CI}{confidence interval}
    \acro{CLIP}{contrastive language--image pre-training}
    \acro{CLM}{causal language modelling}
    \acro{CNN}{convolutional neural network}
    \acro{COLIPRI}{Comprehensive Language--Image Pre-training}
    \acro{CXR}{chest X-ray}
    \acro{DSC}{Dice similarity coefficient}
    \acro{EHR}{electronic health record}
    \acro{FOV}{field of view}
    \acro{LLM}{large language model}
    \acro{MAE}{masked autoencoder}
    \acro{MI2}{\medimageinsight}
    \acro{MIM}{masked image modelling}
    \acro{MLP}{multilayer perceptron}
    \acro{MSD}{Medical Segmentation Decathlon}
    \acro{NLST}{National Lung Screening Trial}
    \acro{NSD}{normalised surface distance}
    \acro{OSL}{Opposite Sentence Loss}
    \acro{PCA}{principal component analysis}
    \acro{PHI}{protected health information}
    \acro{RAG}{retrieval-augmented generation}
    \acro{RRG}{radiology report generation}
    \acro{SSL}{self-supervised learning}
    \acro{SOTA}{state-of-the-art}
    \acro{ViT}{vision transformer} 
    \acro{VLE}{vision--language encoder}
    \acro{VLM}{vision--language model}
    \acro{VQA}{visual question answering}
\end{acronym}

%% file: figures/pca_small.tex
\newcommand{\pcafigwidth}{1}
\newlength{\figw}\setlength{\figw}{\pcafigwidth\textwidth}

\begin{figure*}[t]
  \centering

    \newcommand{\pcawidth}{0.19} 

    \newcommand{\imgct}[1]{figures/slices_pca_figure/ct_slice_#1.png}
    \newcommand{\imgmerlin}[1]{figures/slices_pca_figure/pca_merlin_ras_slice_#1.png}
    \newcommand{\imgctfm}[1]{figures/slices_pca_figure/pca_ct-fm_ras_slice_#1.png}
    \newcommand{\imgctclip}[1]{figures/slices_pca_figure/pca_ct-clip_ras_slice_#1.png}

    \newcommand{\imgcolipriC}[1]{%
      figures/slices_pca_figure/pca_OptimizedClipTrainerOpposedSentences__nnsslPlans__noresample_window_slice_#1.png}
    \newcommand{\imgcolipriCM}[1]{%
      figures/slices_pca_figure/pca_OptimizedCLIPMAETrainer_160_vo1_OS_random_MAEinit__nnsslPlans__noresample_window_slice_#1.png}
    \newcommand{\imgcolipriCR}[1]{%
      figures/slices_pca_figure/pca_OptimizedCLIPCapTrainer_160_OS__nnsslPlans__noresample_window_slice_#1.png}
    \newcommand{\imgcolipriCRM}[1]{%
      figures/slices_pca_figure/pca_CLIPMAECAPTrainer_160_OS_vo1_MAEinit__nnsslPlans__noresample_window_slice_#1.png}

    \newlength{\colw}\setlength{\colw}{\pcawidth\linewidth}
    \newsavebox{\refsag}
    \sbox{\refsag}{\includegraphics[width=\colw]{\imgctclip{sagittal}}}
    \newlength{\Hsag}\setlength{\Hsag}{\ht\refsag}
    \newsavebox{\refcor}
    \sbox{\refcor}{\includegraphics[width=\colw]{\imgctclip{coronal}}}
    \newlength{\Hcor}\setlength{\Hcor}{\ht\refcor}

    \begin{subfigure}[t]{\pcawidth\linewidth}
      \centering
      \includegraphics[width=\linewidth]{\imgct{sagittal}}
    \end{subfigure}\hfill
    \begin{subfigure}[t]{\pcawidth\linewidth}
      \centering
      \includegraphics[width=\linewidth,height=\Hsag]{\imgmerlin{sagittal}}
    \end{subfigure}\hfill
    \begin{subfigure}[t]{\pcawidth\linewidth}
      \centering
      \includegraphics[width=\linewidth]{\imgctfm{sagittal}}
    \end{subfigure}\hfill
    \begin{subfigure}[t]{\pcawidth\linewidth}
      \centering
      \usebox{\refsag}
    \end{subfigure}\hfill
    \begin{subfigure}[t]{\pcawidth\linewidth}
      \centering
      \includegraphics[width=\linewidth]{\imgcolipriCRM{sagittal}}
    \end{subfigure}

    \vspace{0.6ex}

    \begin{subfigure}[t]{\pcawidth\linewidth}
      \centering
      \includegraphics[width=\linewidth]{\imgct{coronal}}
    \end{subfigure}\hfill
    \begin{subfigure}[t]{\pcawidth\linewidth}
      \centering
      \includegraphics[width=\linewidth,height=\Hcor]{\imgmerlin{coronal}}
    \end{subfigure}\hfill
    \begin{subfigure}[t]{\pcawidth\linewidth}
      \centering
      \includegraphics[width=\linewidth]{\imgctfm{coronal}}
    \end{subfigure}\hfill
    \begin{subfigure}[t]{\pcawidth\linewidth}
      \centering
      \usebox{\refcor}
    \end{subfigure}\hfill
    \begin{subfigure}[t]{\pcawidth\linewidth}
      \centering
      \includegraphics[width=\linewidth]{\imgcolipriCRM{coronal}}
    \end{subfigure}

    \vspace{0.6ex}

    \captionsetup[subfigure]{font=tiny, justification=centering, singlelinecheck=false}

    \begin{subfigure}[t]{\pcawidth\linewidth}
      \centering
      \includegraphics[width=\linewidth]{\imgct{axial}}
      \caption{Input}\label{fig:pca:input}
    \end{subfigure}\hfill
    \begin{subfigure}[t]{\pcawidth\linewidth}
      \centering
      \includegraphics[width=\linewidth]{\imgmerlin{axial}}
      \caption{Merlin}\label{fig:pca:merlin}
    \end{subfigure}\hfill
    \begin{subfigure}[t]{\pcawidth\linewidth}
      \centering
      \includegraphics[width=\linewidth]{\imgctfm{axial}}
      \caption{CT-FM}\label{fig:pca:ctfm}
    \end{subfigure}\hfill
    \begin{subfigure}[t]{\pcawidth\linewidth}
      \centering
      \includegraphics[width=\linewidth]{\imgctclip{axial}}
      \caption{CT-CLIP}\label{fig:pca:ctclip}
    \end{subfigure}\hfill
    \begin{subfigure}[t]{\pcawidth\linewidth}
      \centering
      \includegraphics[width=\linewidth]{\imgcolipriCRM{axial}}
      \caption{COLIPRI-CRM (ours)}\label{fig:pca:colipri-crm}
    \end{subfigure}

  \caption{
    \textbf{
        \Ac{PCA} maps of dense 3D features
    }
    obtained from (\subref{fig:pca:input}) a CT scan with a lung mass using different encoders.
    Compared to the baseline methods
    (\subref{fig:pca:merlin},
    \subref{fig:pca:ctfm},
    \subref{fig:pca:ctclip}),
    our \colipricrm encoder
    (\subref{fig:pca:colipri-crm})
    generates sharper and more coherent features.
    The dense embeddings have sizes
    \sizeaniso{7}{7}{10} (\subref{fig:pca:merlin}),
    \sizeaniso{24}{24}{8} (\subref{fig:pca:ctfm}),
    \sizeiso{24} (\subref{fig:pca:ctclip})
    and \sizeiso{48}
    (\subref{fig:pca:colipri-crm}),
    and are interpolated here using bicubic interpolation for visualisation purposes.
    The sagittal (top), coronal (middle) and axial (bottom) slices of the 3D scan are centred on the lung mass.
    The \methodshort (\subref{fig:pca:colipri-crm}) map was computed from the input resampled to 1 mm isotropic, cropped and padded to \sizeiso{384}, using a sliding window of size \sizeiso{192}.
  }
  \label{fig:pca}
\end{figure*}

%% file: figures/diagram.tex

\newcommand{\diagramwidth}{0.44}  

\newcommand{\diagrampanel}[1]{\includegraphics[trim=15 7 15 0, clip, width=\linewidth]{#1}}

\begin{figure}[!t]
    \centering
    \captionsetup[subfigure]{justification=centering}
    \begin{subfigure}[t]{\diagramwidth\linewidth}
      \centering
      \diagrampanel{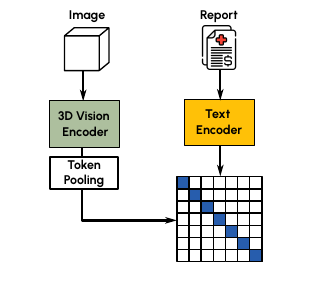}
      \caption{Contrastive image--report alignment}
      \label{fig:colipri_diagram:alignment}
    \end{subfigure}\hfill
    \begin{subfigure}[t]{\diagramwidth\linewidth}
      \centering
      \diagrampanel{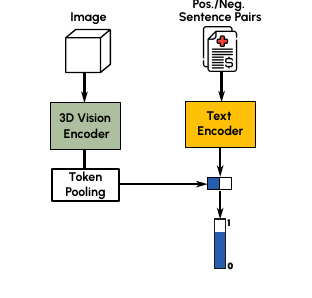}
      \caption{Opposite sentence loss}
      \label{fig:colipri_diagram:opposite_sentence_loss}
    \end{subfigure}%
    \\[\floatsep]
    \begin{subfigure}[t]{\diagramwidth\linewidth}
      \centering
      \diagrampanel{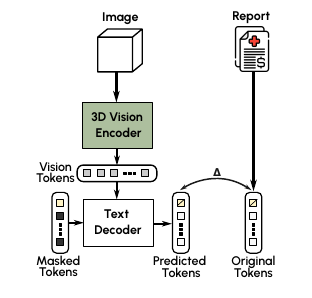}
      \caption{Radiology report generation}
      \label{fig:colipri_diagram:rrg}
    \end{subfigure}\hfill
    \begin{subfigure}[t]{\diagramwidth\linewidth}
      \centering
      \diagrampanel{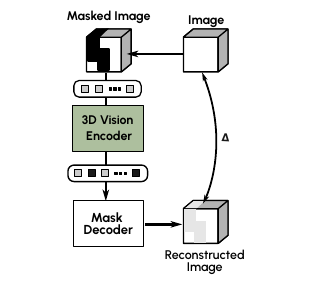}
      \caption{Masked autoencoder}
      \label{fig:colipri_diagram:mae}
    \end{subfigure}%
    \caption{
        Combining
        (\subref{fig:colipri_diagram:alignment}) contrastive image--report pre-training,
        (\subref{fig:colipri_diagram:opposite_sentence_loss}) the novel opposite sentence loss,
        (\subref{fig:colipri_diagram:rrg}) radiology report generation,
        and (\subref{fig:colipri_diagram:mae}) \acp{MAE}
        yields various versions of our \acf{COLIPRI} \acfp{VLE}.
    }
    \label{fig:colipri_diagram}
\end{figure}

%% file: tables/results/classification.tex
\begin{tabular}{l@{\hskip 18pt}l@{\hskip 12pt}l@{\hskip 18pt}l@{\hskip 12pt}l}
\toprule
 & \multicolumn{2}{c}{CT-RATE} & \multicolumn{2}{c}{RAD-ChestCT} \\
\cmidrule(lr){2-3}\cmidrule(lr){4-5}
Model & AUPRC & AUROC & AUPRC & AUROC\\
\midrule
MI2 (2D) & 19.70 & 50.52  & 27.94 & 52.25 \\
Curia (2D) & 45.93 & 78.01 & 39.23 & 65.73 \\
CT-CLIP & 25.96 \ci{[24.74, 27.48]} & 61.21 \ci{[60.02, 62.45]} & 28.77 \ci{[28.23, 29.49]} & 54.05 \ci{[53.12, 55.01]} \\
CT-CLIP (reported) & - & 75.1 & - & 64.7 \\
CT-FM & 53.54 \ci{[51.93, 55.02]} & 82.14 \ci{[81.44, 82.82]} & 42.41 \ci{[41.66, 43.21]} & 68.49 \ci{[67.97, 69.15]} \\
MAE & 53.79 \ci{[52.55, 55.39]} & 82.70 \ci{[82.02, 83.31]} & 45.24 \ci{[44.71, 46.07]} & 71.20 \ci{[70.60, 71.71]} \\
Merlin & 54.81 \ci{[53.19, 56.81]} & 82.62 \ci{[81.84, 83.30]} & 45.30 \ci{[44.45, 46.20]} & 70.91 \ci{[70.30, 71.48]} \\\midrule
\cellcolor{microsoftred!20}\colipric & 55.13 \ci{[53.59, 56.80]} & 82.63 \ci{[81.87, 83.34]} & 46.33 \ci{[45.61, 47.37]} & 71.20 \ci{[70.65, 71.78]} \\
\cellcolor{microsoftblue!20}\colipricr & 57.44 \ci{[56.00, 59.02]} & 83.50 \ci{[82.90, 84.04]} & 48.28 \ci{[47.44, 49.12]} & 72.78 \ci{[72.16, 73.42]} \\
\cellcolor{microsoftgreen!20}\colipricm & \textbf{61.52 \ci{[60.14, 63.03]}} & \textbf{86.15 \ci{[85.66, 86.78]}} & \textbf{51.79 \ci{[50.93, 52.76]}} & \textbf{76.16 \ci{[75.65, 76.61]}} \\
\rowcolor{gray!20}  \cellcolor{microsoftyellow!20}{\colipricrm}  & \textbf{61.28 \ci{[59.59, 63.10]}} & \textbf{86.12 \ci{[85.52, 86.85]}} & \textbf{52.55 \ci{[51.72, 53.65]}} & \textbf{76.57 \ci{[76.08, 77.18]}} \\
\bottomrule
\end{tabular}

%% file: tables/results/zeroshot_classification_main.tex
\begin{tabular}{l|ll|ll}
\toprule
 & \multicolumn{2}{c|}{CT-RATE} & \multicolumn{2}{c}{RAD-ChestCT} \\
 Model & AUPRC & AUROC & AUPRC & AUROC \\
\midrule
\ac{MI2} (2D) & 18.80 & 49.36 & 26.32 & 48.82 \\
Curia (2D) & 19.18 & 50.17 & 26.84 & 50.13 \\
CT-CLIP & - & 70.4$^*$ & - &  63.2$^*$\\
BIUD & - & 71.3$^*$ & - & 62.9$^*$ \\
Merlin &  - & 72.8$^*$ & - & 64.4$^*$ \\
fVLM & - & 77.8$^*$ & - &  68.0$^*$ \\\midrule
\cellcolor{microsoftred!20}{\colipric} & 43.81 \ci{[42.9, 45.2]} & 76.25 \ci{[75.5, 77.2]} & 40.98 \ci{[40.3, 41.8]} & 69.12 \ci{[68.6, 69.7]} \\ 
\cellcolor{microsoftblue!20}{\colipricr} & 44.70 \ci{[43.7, 46.5]} & 76.41 \ci{[75.7, 77.2]} & 39.38 \ci{[38.7, 40.1]} & 67.30 \ci{[66.6, 68.1]} \\
\cellcolor{microsoftgreen!20}{\colipricm} & \textbf{50.32 \ci{[48.7, 51.8]}} & \textbf{80.52 \ci{[79.7, 81.4]}} & 43.26 \ci{[42.5, 44.3]} & 72.05 \ci{[71.4, 72.7]} \\
\rowcolor{gray!15}\cellcolor{microsoftyellow!20}{\colipricrm } & \textbf{49.39 \ci{[48.3, 51.3]}} & \textbf{79.81 \ci{[79.2, 80.7]}} & \textbf{44.48 \ci{[43.7, 45.4]}} & \textbf{73.23 \ci{[72.6, 73.8]}} \\
\bottomrule
\end{tabular}

%% file: tables/results/zeroshot_classification_opposite_sentence.tex
\begin{tabular}{l|rr|rr}
\toprule
Prompt Style & \multicolumn{2}{c}{Native} & \multicolumn{2}{c}{Short} \\
 & AUPRC & AUROC & AUPRC & AUROC \\
\midrule
\rowcolor{gray!15}\cellcolor{microsoftyellow!20}\colipricrm & 51.17 & 81.91 & 50.70 & 81.66 \\
without OSL & 47.48 & 80.48 & 39.08 & 74.09 \\

\bottomrule
\end{tabular}

%% file: tables/results/test_retrieval_T.tex
\begin{tabular}{lrrrrr}
\toprule
 & CT-CLIP & \cellcolor{microsoftred!20}\colipric & \cellcolor{microsoftblue!20}\colipricr & \cellcolor{microsoftgreen!20}\colipricm & \cellcolor{microsoftyellow!20}\colipricrm \\
\midrule
Recall @1  & -     & 4.49  & 7.23  & 14.53 & \textbf{15.27} \\
Recall @5  & 2.90  & 13.46 & 19.36 & 34.16 & \textbf{35.10} \\
Recall @10 & 5.00  & 19.69 & 26.99 & 45.34 & \textbf{46.01} \\
\bottomrule
\end{tabular}

%% file: tables/results/segmentation_short_T.tex
\begin{tabular}{l|rr|rrrrrr|rrr}
\toprule
 & \multicolumn{2}{c}{\textbf{References} (250k steps)} 
 & \multicolumn{6}{c}{\textbf{Primus-M} (37.5k steps)}
 & \multicolumn{3}{c}{\textbf{Primus-M} (250k steps)} \\
\cmidrule(lr){2-3}\cmidrule(lr){4-9}\cmidrule(lr){10-12}
\textbf{Dataset}
& \textbf{nnU-Net} & \textbf{ResEnc-L}
& \textbf{Scratch} & \textbf{MAE}
& \cellcolor{microsoftred!20}\textbf{C}
& \cellcolor{microsoftblue!20}\textbf{CR}
& \cellcolor{microsoftgreen!20}\textbf{CM}
& \cellcolor{microsoftyellow!20}\textbf{CRM}
& \textbf{Scratch} & \textbf{MAE} & \cellcolor{microsoftyellow!20}\textbf{CRM} \\
\midrule
\textbf{LiTS}
& 80.09 & 81.60
& 74.39 & \underline{80.27}
& 76.77 & 77.09 & 79.97 & \textbf{80.46}
& 79.75 & \underline{80.32} & \textbf{81.11} \\
\textbf{Lung}
& 70.32 & 70.34
& 62.74 & 67.12
& 65.89 & 65.32 & \underline{68.74} & \textbf{68.98}
& \textbf{69.11} & 67.46 & \underline{67.56} \\
\textbf{HVS}
& 68.38 & 67.73
& 64.67 & \textbf{67.10}
& 64.64 & 65.08 & 66.58 & \underline{67.05}
& 65.34 & \underline{65.73} & \textbf{66.42} \\
\textbf{KiTS23}
& 86.04 & 88.17
& 78.90 & 85.34
& 81.02 & 80.71 & \textbf{86.03} & \underline{85.79}
& 86.25 & \underline{87.46} & \textbf{87.68} \\
\bottomrule
\end{tabular}

%% file: tables/dataset_sizes.tex
\begin{tabular}{llr}
\toprule
\textbf{Dataset} & \textbf{Body part and technique} & \textbf{Num. acquisitions} \\\midrule
\multicolumn{3}{c}{\textbf{Paired images and reports}} \\\midrule
BIMCV-R~\cite{vaya2020bimcv}& Chest CT        & 8k \\
INSPECT~\cite{huang2023inspect}        & Chest CT pulmonary angiography (only \textit{Impressions})    & 23k \\
CT-RATE~\cite{hamamci_developing_2024}        & Chest CT         & 26k \\
Merlin~\cite{blankemeier_merlin_2024}         & Abdomen and pelvis CT  & 25k \\
\midrule
\multicolumn{3}{c}{\textbf{Images only}} \\\midrule
NLST~\cite{national2011national}           & Chest CT         & 72k \\
OpenMind~\cite{wald2025openmind}       & Brain MRI        & 114k \\
Osteoarthritis Initiative (OAI)         & Knee / thigh MRI & 140k \\
UK Biobank~\cite{littlejohns2020uk}     & Whole-body MRI & $>100$k \\\bottomrule
\end{tabular}

%% file: tables/language_table.tex
\begin{tabular}{l|rcc|cc|cc|cc}
\toprule
 & \multicolumn{3}{c}{Retrieval} & \multicolumn{2}{c}{Probing} & \multicolumn{2}{c}{Zero-shot (N)} & \multicolumn{2}{c}{Zero-shot (S)} \\
 & R@1 & R@5 & R@10 & AUPRC & AUROC & AUPRC & AUROC & AUPRC & AUROC \\\midrule
\rowcolor{gray!20}Default & 8.27 & 22.64 & 31.66 & 55.41 & 83.11 & 43.48 & 76.48 & 34.77 & 66.91 \\\midrule
Shuffle & 11.11 & 28.57 & 37.93 & 56.66 & 83.94 & 44.05 & 76.55 & 35.13 & 69.21 \\\midrule
Shorten & & & & & & & & & \\
$p$ = 0.10 & 12.71 & 29.01 & 39.30 & 56.91 & 83.76 & 45.81 & 77.93 & 39.55 & 71.15 \\
$p$ = 0.25 & 11.54 & 27.68 & 38.04 & 56.45 & 83.87 & 47.12 & 78.67 & 35.13 & 68.10 \\
$p$ = 0.50 & 11.20 & 28.01 & 38.21 & 56.32 & 83.91 & 46.07 & 78.63 & 37.09 & 70.24 \\
$p$ = 0.75 & 9.78 & 25.92 & 34.36 & 56.97 & 84.01 & 46.41 & 78.74 & 34.70 & 68.19 \\
\bottomrule
\end{tabular}

%% file: tables/appendix/sentence_shuffle_findings_impressions.tex
\begin{tabular}{l|ccc|cc|cc|cc}
\toprule
& \multicolumn{3}{c}{Retrieval} & \multicolumn{2}{c}{Probing} & \multicolumn{2}{c}{Zero-shot (N)} & \multicolumn{2}{c}{Zero-shot (S)} \\
& R1 & R5 & R10 & AUPRC & AUROC & AUPRC & AUROC & AUPRC & AUROC \\
\midrule
\rowcolor{gray!20} Default & 8.27 & 22.64 & 31.66 & 55.41 & 83.11 & 43.48 & 76.48 & 34.77 & 66.91 \\
Sentence Shuffle & 11.11 & 28.57 & 37.93 & 56.66 & 83.94 & 44.05 & 76.55 & 35.13 & 69.21 \\\midrule
\rowcolor{gray!20} Findings & 8.27 & 22.64 & 31.66 & 55.41 & 83.11 & 43.48 & 76.48 & 34.77 & 66.91 \\
Impressions & 7.77 & 21.80 & 30.58 & 54.89 & 83.29 & 43.74 & 76.52 & 29.11 & 62.40 \\
Findings + Impressions & 8.69 & 22.22 & 31.83 & 55.31 & 83.27 & 43.64 & 76.74 & 31.39 & 67.64 \\\midrule
\rowcolor{gray!20} Original translation & 8.27 & 22.64 & 31.66 & 55.41 & 83.11 & 43.48 & 76.48 & 34.77 & 66.91 \\
Re-translate & 7.53 & 20.90 & 29.35 & 55.24 & 83.00 & 42.69 & 76.29 & 28.29 & 60.11 \\
Re-translate + Shuffle & 9.95 & 24.58 & 32.27 & 56.21 & 83.64 & 44.22 & 76.73 & 32.05 & 67.19 \\\midrule
BiomedCLIP & 3.43 & 13.28 & 19.47 &  52.79 &  81.89 &  37.43& 74.21 & 28.16 & 59.30 \\
CXR-BERT (scratch) & DnC & DnC & DnC & DnC & DnC & DnC & DnC & DnC & DnC \\
\rowcolor{gray!20} CXR-BERT (pre-trained)  & 8.27 &  22.64 &  31.66 &  55.41 & 83.11 & 43.48 & 76.48 & 34.77  & 66.91\\
\bottomrule
\end{tabular}

%% file: tables/field_of_view_table.tex
\begin{tabular}{l|rcc|cc|cc|cc}
\toprule
& \multicolumn{3}{c}{Retrieval} & \multicolumn{2}{c}{Probing} & \multicolumn{2}{c}{Zero-shot (N)} & \multicolumn{2}{c}{Zero-shot (S)} \\
& R@1 & R@5 & R@10 & AUPRC & AUROC & AUPRC & AUROC & AUPRC & AUROC \\\midrule
\multicolumn{10}{l}{\textbf{Patch size}}\\
16$\times$16$\times$16 & 5.76 & 14.62 & 22.14 & 49.88 & 81.05 & 40.71 & 75.12 & 29.58 & 61.80 \\
\rowcolor{gray!20}8$\times$8$\times$8 & 8.27 & 22.64 & 31.66 & 55.41 & 83.11 & 43.48 & 76.48 & 34.77 & 66.91 \\\midrule
\multicolumn{10}{l}{\textbf{Input size}}\\
128$\times$128$\times$128 & 8.35 & 22.06 & 29.66 & 56.19 & 83.78 & 44.60 & 77.17 & 32.00 & 63.05 \\
160$\times$160$\times$160 & 7.94 & 23.64 & 32.75 & 55.91 & 83.51 & 43.22 & 76.24 & 34.77 & 66.70 \\
\rowcolor{gray!20}192$\times$192$\times$192 & 8.27 & 22.64 & 31.66 & 55.41 & 83.11 & 43.48 & 76.48 & 34.77 & 66.91 \\
224$\times$224$\times$224 & - & - & - & 54.44 & 83.06 & 42.69 & 76.13 & 29.99 & 64.78 \\\midrule
No \acl{APE} & 9.27 & 23.48 & 33.33 & 55.38 & 83.20 & 43.30 & 76.18 & 25.07 & 58.56 \\\midrule
\multicolumn{10}{l}{\textbf{Token aggregation method}}\\
Avg.\ pooling & 5.93 & 18.88 & 27.90 & 54.34 & 83.03 & 40.20 & 75.48 & 35.65 & 67.46 \\
Max pooling & 11.45 & 27.57 & 38.01 & 56.18 & 83.56 & 42.80 & 75.32 & 25.86 & 56.57 \\
Single-head attention pooling & 5.43 & 19.13 & 27.57 & 54.05 & 82.82 & 40.84 & 75.83 & 33.27 & 62.19 \\
\rowcolor{gray!20}Multi-head attention pooling & 8.27 & 22.64 & 31.66 & 55.41 & 83.11 & 43.48 & 76.48 & 34.77 & 66.91 \\
\bottomrule
\end{tabular}

%% file: tables/miscellaneous.tex
\begin{tabular}{l|ccc|cc|cc|cc}
\toprule
& \multicolumn{3}{c}{Retrieval} & \multicolumn{2}{c}{Probing} & \multicolumn{2}{c}{Zero-shot (N)} & \multicolumn{2}{c}{Zero-shot (S)} \\
& R@1 & R@5 & R@10 & AUPRC & AUROC & AUPRC & AUROC & AUPRC & AUROC \\\midrule
\multicolumn{10}{l}{\textbf{Batch size - training Steps}}\\
\rowcolor{gray!20} 8 - 250k & 8.27 & 22.64 & 31.66 & 55.41 & 83.11 & 43.48 & 76.48 & 34.77 & 66.91  \\
16 - 125k & 9.77 & 24.98 & 32.75 & 55.66 & 83.44 & 41.68 & 75.48 & 31.82 & 63.40\\
24 - 62.5k & 8.10 & 23.89 & 31.75 & 55.95 & 83.51 & 43.13 & 76.05 & 30.87 & 62.27\\
32 - 31.7k & 8.27 & 20.05 & 29.32 & 55.19 & 83.23 & 40.55 & 74.77 & 23.69 & 53.89 \\\midrule
\multicolumn{10}{l}{\textbf{Loss type}}\\
\rowcolor{gray!20} Softmax loss & 8.27 & 22.64 & 31.66 & 55.41 & 83.11 & 43.48 & 76.48 & 34.77 & 66.91  \\
Sigmoid loss & 5.60 & 16.21 & 24.06 & 53.88 & 82.81 & 39.45 & 74.66 & 30.90 & 63.44\\\midrule
\multicolumn{10}{l}{\textbf{Image augmentation (spatial -- intensity}}\\
\rowcolor{gray!20} low -- off  & 8.27 & 22.64 & 31.66 & 55.41 & 83.11 & 43.48 & 76.48 & 34.77 & 66.91 \\
low -- low & 9.02 & 22.47 & 30.49 & 55.73 & 83.65 & 43.43 & 76.05 & 31.90 & 63.28 \\
high -- high & 4.59 & 15.96 & 23.73 & 54.91 & 83.11 & 41.44 & 76.16 & 30.71 & 59.63 \\
\bottomrule
\end{tabular}

%% file: tables/clipcap/clipcap_joint.tex
\begin{tabular}{l|rrr|cc|cc|cc}
\toprule
 & \multicolumn{3}{c}{Retrieval} & \multicolumn{2}{c}{Probing} & \multicolumn{2}{c}{Zero-shot (N)} & \multicolumn{2}{c}{Zero-shot (S)} \\
 & R@1 & R@5 & R@10 & AUPRC & AUROC & AUPRC & AUROC & AUPRC & AUROC \\
\midrule
\multicolumn{4}{l}{\textbf{Depth of text decoder}}\\
 4 & 9.52 & 25.31 & 33.25   & 55.85 & 83.40 & 46.29 & 76.53 & 37.69 & 69.41 \\
 6 & 11.70 & 26.57 & 35.42 & 55.72 & 83.28 & 44.73 & 75.88 & 34.56 & 69.39 \\
 8 &  10.44 & 24.56 & 32.41 & 54.14 & 82.59 & 45.20 & 75.80 & 31.23 & 60.90 \\
 \rowcolor{gray!20} 12 & 10.03 & 24.06 & 32.33 &55.09 & 82.89 & 43.87 & 75.35 & 32.94 & 66.49 \\\midrule
\multicolumn{4}{l}{\textbf{CapPa--Cap Probability (\%)}}\\

0--100 & 8.10 & 22.72 & 30.33  & 54.53 & 82.80 & 44.66 & 74.97 & 35.11 & 63.22 \\
25--75 & 9.44 & 21.81 & 29.24  & 54.27 & 73.52 & 43.48 & 75.59 & 29.28 & 61.28 \\
\rowcolor{gray!20} 50--50 &   10.03 & 24.06 & 32.33 & 55.09 & 82.89 & 43.87 & 75.35 & 32.94 & 66.49 \\
75--25 &  10.36 & 25.31 & 34.34 & 55.32 & 83.28 & 45.80 & 77.03 & 35.82 & 68.63 \\
100--0 &  11.03 & 24.65 & 34.25  & 56.36 & 83.42 & 44.89 & 75.64 & 31.13 & 61.41 \\\midrule
\multicolumn{4}{l}{\textbf{$\lambda_{\text{RRG}}$}} \\
 0.1 & 11.03 & 26.15 & 35.09 & 55.79 & 83.14 & 45.56 & 76.43 & 33.65 & 64.59 \\
 0.3 & 9.27 & 24.48 & 33.00  & 55.29 & 83.17 & 45.38 & 76.70 & 36.46 & 68.09 \\
\rowcolor{gray!20} 1 & 10.03 & 24.06 & 32.33  & 55.09 & 82.89 & 43.87 & 75.35 & 32.94 & 66.49 \\
\bottomrule
\end{tabular}

%% file: tables/clipmae/clipmae_joint.tex
\begin{tabular}{l|ccc|cc|cc|cc}
\toprule
 & \multicolumn{3}{c}{Retrieval} & \multicolumn{2}{c}{Probing} & \multicolumn{2}{c}{Zero-shot (N)} & \multicolumn{2}{c}{Zero-shot (S)} \\
 & R@1 & R@5 & R@10 & AUPRC & AUROC & AUPRC & AUROC & AUPRC & AUROC \\
\midrule
\multicolumn{10}{l}{\textbf{MAE decoder depth}} \\
 2 & 14.04 & 30.91 & 39.77  & 55.15 & 82.99 & 44.61 & 76.31 & 38.15 & 70.84 \\
 \rowcolor{gray!20} 4 & 13.28 & 30.41 & 38.60 & 54.81 & 83.13 & 44.22 & 76.76 & 40.55 & 73.70 \\
 6 & 14.45 & 33.42 & 41.52 & 55.71 & 83.44 & 45.22 & 77.04 & 38.14 & 71.50 \\
 8 & 12.78 & 29.99 & 39.68 & 55.04 & 82.95 & 44.73 & 76.81 & 37.30 & 70.70 \\\midrule
 \multicolumn{10}{l}{\textbf{Masking ratio}} \\
  60\% & 13.78 & 30.83 & 38.85 & 55.08 & 83.17 & 44.71 & 76.14 & 38.76 & 72.57 \\
 \rowcolor{gray!20} 75\%& 13.28 & 30.41 & 38.60 & 54.81 & 83.13 & 44.22 & 76.76 & 40.55 & 73.70 \\
 90\% & 14.20 & 32.08 & 41.85 & 55.41 & 83.02 & 44.32 & 76.32 & 37.92 & 71.94 \\\midrule
  \multicolumn{10}{l}{\textbf{Mask style}} \\
  \rowcolor{gray!20} Random & 13.28 & 30.41 & 38.60 & 54.81 & 83.13 & 44.22 & 76.76 & 40.55 & 73.70 \\
 Block & 14.87 & 29.74 & 38.68 & 55.46 & 83.12 & 44.63 & 76.84 & 39.33 & 72.45 \\
  Inverse block & 13.37 & 30.41 & 40.02 & 55.44 & 83.19 & 45.01 & 76.83 & 39.22 & 72.37 \\\midrule
  \multicolumn{10}{l}{\textbf{Included at last X\% of training}} \\
  25\% & 13.95 & 29.07 & 36.51 & 56.41 & 83.83 & 45.42 & 76.57 & 41.99 & 74.95 \\
  50\% & 12.95 & 29.91 & 38.68& 56.40 & 83.39 & 45.26 & 76.63 & 40.65 & 72.83 \\
  75\% & 14.54 & 30.08 & 38.35 & 55.62 & 83.28 & 45.64 & 77.47 & 36.92 & 70.81 \\
  \rowcolor{gray!20}100\% & 13.28 & 30.41 & 38.60 & 54.81 & 83.13 & 44.22 & 76.76 & 40.55 & 73.70 \\\midrule
\multicolumn{10}{l}{\textbf{$\lambda_{\text{MAE}}$}} \\
 0.1 & 14.62 & 29.16 & 37.59 & 54.99 & 83.07 & 44.82 & 76.57 & 32.75 & 66.17 \\
0.5 &13.62 & 29.91 & 38.68 & 55.34 & 83.06 & 44.77 & 76.49 & 40.10 & 72.25 \\
 \rowcolor{gray!20} 1.0 & 13.28 & 30.41 & 38.60 & 54.81 & 83.13 & 44.22 & 76.76 & 40.55 & 73.70 \\
  2.0 & 14.62 & 31.75 & 40.52 & 55.08 & 83.07 & 45.05 & 76.96 & 37.31 & 69.74 \\\midrule
\multicolumn{10}{l}{\textbf{Smallest isotropic spacing included}} \\
 \rowcolor{gray!20} 2 mm    & 13.28 & 30.41 & 38.60 & 54.81 & 83.13 & 44.22 & 76.76 & 40.55 & 73.70 \\
 1 mm                       & 15.46 & 31.66 & 39.77 & 56.19 & 83.44 & 43.16 & 76.48 & 41.35 & 74.90 \\
0.5 mm                      & 14.29 & 29.57 & 38.60 & 55.19 & 82.98 & 43.38 & 76.12 & 40.67 & 73.51 \\
\bottomrule
\end{tabular}


%% file: tables/colipri_table.tex
\begin{tabular}{l|rrr|ll|ll|ll}
\toprule
 & \multicolumn{3}{c}{Retrieval} & \multicolumn{2}{c}{Probing} & \multicolumn{2}{c}{Zero-shot (N)} & \multicolumn{2}{c}{Zero-shot (S)} \\
 & R@1 & R@5 & R@10 & AUPRC & AUROC & AUPRC & AUROC & AUPRC & AUROC \\
\midrule
\rowcolor{microsoftred!20} \colipric  & 11.36 & 27.74 & 37.34  & 55.59 & 83.13 & 44.53 & 78.93 & 43.74 & 77.05 \\
\rowcolor{microsoftblue!20} \colipricr & 
16.46 & 33.25 & 41.60 & 57.05 & 83.46 & 48.27 & 79.34 & 45.05 & 77.53 \\
\rowcolor{microsoftgreen!20} \colipricm & \underline{26.40} & \underline{51.13} & \underline{61.57} & \underline{61.11} & \underline{86.17} & \underline{51.17} & \textbf{81.91} & \textbf{50.70} & \textbf{81.66} \\
\rowcolor{microsoftyellow!20} \colipricrm & \textbf{28.74} & \textbf{53.55} & \textbf{62.16} & \textbf{61.12} & \textbf{86.27} & \textbf{51.19} & \underline{81.64} & \underline{50.57} & \underline{81.53}\\
\bottomrule
\end{tabular}

%% file: tables/hyperparams_clip.tex
\begin{tabular}{ll}
     \toprule
     \multicolumn{2}{l}{\textbf{Vision Encoder}}\\
     Architecture & Primus-M \\
     Layer scale & 0.1 \\
     Drop path rate & 0.2 \\
     Post-attention normalisation & True \\
     \Acl{APE} & False \\
     3D RoPE & True \\
     Patch Size & 8$\times$8$\times$8\\
     Pooling & Multi-head attention pooling \\
     Num.\ att. pooling heads & 12 \\
     Attention pooling query & Mean pooling \\
     Vision encoder init & Random \\\midrule
     
     \multicolumn{2}{l}{\textbf{Text encoder}}\\
     Architecture & CXR-BERT\footnote{\url{https://huggingface.co/microsoft/BiomedVLP-CXR-BERT-specialized}}\\
     Tokeniser & Tokeniser associated with CXR-BERT \\
     Max.\ token length & 512 \\
     Text pooling & Multi-head attention pooling \\
     Num.\ attention pooling heads & 12 \\
     Att. pooling query method & Mean pooling \\\midrule
     
     \multicolumn{2}{l}{\textbf{Training Parameters}}\\
     Input size & 160$\times$160$\times$160\\
     Batch size & 16\\
     Learning rate & 6 $\times$ 10$^{-5}$ \\
     Weight decay & 5 $\times$ 10$^{-3}$ \\
     Optimiser & AdamW \\
     Betas & (0.9, 0.98)\\
     Num.\ warm-up steps & 6250 \\
     Warm-up schedule & Linear \\
     Total num.\ steps & 125k \\
     Learning rate schedule & PolynomialLR \\
     Image augmentation & Default nnU-Net augmentation \\
     \textit{Sentence Shuffle} aug. & True \\
     \textit{Short Sentence} aug. prob. & 25\% \\
     Text used & \textit{Findings}\\\midrule
     
     \multicolumn{2}{l}{\textbf{Loss:} $\mathcal{L}_{align}$ = $ \lambda_{CLIP}\cdot\mathcal{L}_{CLIP} + \lambda_{OSL}\cdot\mathcal{L}_{OSL}$}\\
     $\mathcal{L}_{CLIP}$ & Symmetric InfoNCE \\
     $\mathcal{L}_{OSL}$ & Cross-entropy \\
     CLIP temperature $\mathcal{T}$ & 0.07 \\
     $\lambda_{CLIP}$ & 0.5 \\
     $\lambda_{OSL}$ & 0.5 \\
     Num. \ac{OSL} sentence pairs &  8 \\\bottomrule
\end{tabular}

%% file: tables/hyperparameters_rrg.tex
\begin{tabular}{ll}
     \toprule
     \multicolumn{2}{l}{\textbf{Report Generator}}\\
     Architecture & EVA-02 \\
     Depth & 4 \\\midrule
     \multicolumn{2}{l}{\textbf{Training Parameters}}\\
     Learning rate & 3 $\times$ 10$^{-5}$ \\
     CapPa -- Cap Probability & 100\% -- 0\%\\\midrule
     \multicolumn{2}{l}{\textbf{Loss:} $\mathcal{L}_{VLM}$ = $ \lambda_{\text{CLIP}}\cdot\mathcal{L}_{\text{CLIP}} + \lambda_{\text{OSL}}\cdot\mathcal{L}_{\text{OSL}} +\lambda_{\text{RRG}}\cdot\mathcal{L}_{\text{RRG}}$} \\
     $\mathcal{L}_{RRG}$ & Cross-entropy \\
     $\lambda_{RRG}$ & 0.3 \\\bottomrule
\end{tabular}

%% file: tables/hyperparameters_mae.tex
\begin{tabular}{ll}
     \toprule
     \multicolumn{2}{l}{\textbf{Mask Generator}}\\
     Architecture & EVA-02 \\
     Depth & 6 \\\midrule
     \multicolumn{2}{l}{\textbf{Training Parameters}}\\
     Learning rate & 3 $\times$ 10$^{-5}$ \\
     Mask ratio & 75\% \\
     Mask style & Random masking \\ 
     Inclusion at last X\% of training & 100\% (i.e., since the beginning) \\
     Smallest spacing & 1 mm \\
     Vision encoder init. & \Acf{MAE} \\\midrule
     \multicolumn{2}{l}{\textbf{Loss:} $\mathcal{L}_{Align}$ or $ \mathcal{L}_{VO} = \lambda_{\text{MAE}}\cdot\mathcal{L}_{\text{MAE}}$ }\\
     $\lambda_{MAE}$ & 1 \\
     $\mathcal{L}_{MAE}$ & Mean squared error (where masked)\\\bottomrule
\end{tabular}

%% file: tables/results/zeroshot_classification_appendix.tex
\begin{tabular}{ll|ll|ll|ll|ll}
\toprule
Model & Prompt  & \multicolumn{4}{c|}{CT-RATE} & \multicolumn{4}{c}{RAD-ChestCT} \\
Included Classes &  & \multicolumn{2}{c|}{All abnormalities} & \multicolumn{2}{c|}{like fVLM} & \multicolumn{2}{c|}{All abnormalities} & \multicolumn{2}{c}{like fVLM} \\
 &  & AUPRC & AUROC & AUPRC & AUROC& AUPRC & AUROC & AUPRC & AUROC  \\
\midrule
MI2 & short & 19.03 & 49.47 & 18.80 & 49.36 & 26.57 & 48.78 & 26.32 & 48.82 \\
Curia & short & 19.33 & 49.93 & 19.18 & 50.17 & 27.14 & 50.21 & 26.84 & 50.13 \\
CT-CLIP & (own) & - & - &  - & 70.4$^*$ & - & 63.2$^*$ \\
BIUD & (own) & - & - &  - & 71.3$^*$ & - & 62.9$^*$ \\
Merlin & (own) & - & - &  - & 72.8$^*$ & - & 62.9$^*$  \\
fVLM & (own) & - & - & - & 77.8$^*$ & - & 68.0$^*$ \\
\cellcolor{microsoftred!20}{\colipric} & short &  42.04 \ci{[41.1, 43.3]} & 75.02 \ci{[74.3, 75.9]} & 43.81 \ci{[42.9, 45.2]} & 76.25 \ci{[75.5, 77.2]} & 41.05 \ci{[40.6, 41.9]} & 68.71 \ci{[68.3, 69.2]} & 40.98 \ci{[40.3, 41.8]} & 69.12 \ci{[68.6, 69.7]} \\
\cellcolor{microsoftblue!20}{\colipricr} & short & 43.36 \ci{[42.4, 45.0]} & 75.23 \ci{[74.5, 76.1]} & 44.70 \ci{[43.7, 46.5]} & 76.41 \ci{[75.7, 77.2]} & 39.93 \ci{[39.3, 40.6]} & 67.07 \ci{[66.5, 67.8]} & 39.38 \ci{[38.7, 40.1]} & 67.30 \ci{[66.6, 68.1]} \\
\cellcolor{microsoftgreen!20}{\colipricm} & short & \textbf{48.62 \ci{[47.2, 50.1]}} & \textbf{79.78 \ci{[79.0, 80.6]}} & \textbf{50.32 \ci{[48.7, 51.8]}} & \textbf{80.52 \ci{[79.7, 81.4]}} & \textbf{44.45 \ci{[43.7, 45.4]}} & 72.20 \ci{[71.7, 72.8]} & 43.26 \ci{[42.5, 44.3]} & 72.05 \ci{[71.4, 72.7]} \\
\rowcolor{gray!20} \cellcolor{microsoftyellow!20}{\colipricrm}  & short & \textbf{47.53 \ci{[46.4, 49.5]}} & 78.53 \ci{[77.8, 79.3]} & \textbf{49.39 \ci{[48.3, 51.3]}} & \textbf{79.81 \ci{[79.2, 80.7]}} & \textbf{45.15 \ci{[44.4, 45.9]}} & \textbf{72.98 \ci{[72.4, 73.5]}} & \textbf{44.48 \ci{[43.7, 45.4]}} & \textbf{73.23 \ci{[72.6, 73.8]}} \\\midrule
MI2 & native & 19.33 & 49.99 & 19.51 & 49.97 & 26.85 & 49.93 & 27.12 & 49.85 \\
Curia & native & 19.54 & 50.27 & 19.80 & 50.46 & 26.94 & 50.29 & 27.23 & 50.24 \\
\cellcolor{microsoftred!20}{\colipric} & native & 41.49 \ci{[39.7, 42.9]} & 75.48 \ci{[74.8, 76.2]} & 42.16 \ci{[40.5, 43.6]} & 75.74 \ci{[74.9, 76.5]} & 39.11 \ci{[38.4, 39.9]} & 66.40 \ci{[65.8, 67.0]} & 38.20 \ci{[37.5, 39.1]} & 66.34 \ci{[65.7, 66.9]} \\
\cellcolor{microsoftblue!20}{\colipricr} & native & 43.57 \ci{[42.0, 45.0]} & 75.90 \ci{[74.9, 76.7]} & 43.56 \ci{[42.1, 44.9]} & 75.77 \ci{[74.8, 76.5]} & 40.85 \ci{[40.2, 41.6]} & 67.72 \ci{[67.1, 68.3]} & 40.30 \ci{[39.7, 41.2]} & 67.86 \ci{[67.3, 68.4]} \\
\cellcolor{microsoftgreen!20}{\colipricm} & native & \textbf{49.53 \ci{[47.7, 50.9]}} & \textbf{80.27 \ci{[79.4, 81.0]}} & \textbf{49.94 \ci{[48.2, 51.3]}} & \textbf{80.44 \ci{[79.5, 81.2]}} & \textbf{43.88 \ci{[43.2, 44.8]}} & \textbf{70.89 \ci{[70.4, 71.4]}} & \textbf{42.21 \ci{[41.6, 43.1]}} & \textbf{70.42 \ci{[69.8, 71.0]}} \\
 \rowcolor{gray!20} \cellcolor{microsoftyellow!20}{\colipricrm} & native & \textbf{48.78 \ci{[47.0, 50.5]}} & \textbf{79.53 \ci{[78.5, 80.3]}} & \textbf{49.34 \ci{[47.8, 51.0]} }& \textbf{79.70 \ci{[78.7, 80.4]}} & \textbf{43.84 \ci{[43.3, 44.6]}} & \textbf{71.19 \ci{[70.7, 71.7]}} & \textbf{42.36 \ci{[41.8, 43.2]}} & \textbf{70.86 \ci{[70.4, 71.4]}} \\
\bottomrule
\end{tabular}

%% file: tables/results/report_gen.tex
\begin{tabular}{l|lllll|llll}
\toprule
Metric & MI2 & Curia & CT-CLIP & CT-FM & Merlin & \cellcolor{microsoftred!20}{\colipric} & \cellcolor{microsoftblue!20}{\colipricr} &  \cellcolor{microsoftgreen!20}{\colipricm}& \cellcolor{microsoftyellow!20}{\colipricrm} \\
\midrule
\multicolumn{8}{l}{\textbf{Lexical  metrics}}\\
ROUGE-L & \textbf{54.2 \ci{[53.1, 55.4]}} & \textbf{54.4 \ci{[53.1, 55.5]}} & \textbf{54.2 \ci{[53.0, 55.3]}} & 52.4 \ci{[51.4, 53.4]} & 53.2 \ci{[52.1, 54.3]} & \textbf{55.0 \ci{[53.8, 56.1]}} & 54.0 \ci{[53.0, 55.2]} & \textbf{55.0 \ci{[53.7, 56.2]}} & \textbf{55.2 \ci{[54.1, 56.3]}} \\
BLEU-1 & 55.5 \ci{[54.2, 57.0]} & 55.1 \ci{[53.8, 56.5]} & 55.6 \ci{[54.3, 57.1]} & 57.8 \ci{[56.6, 59.1]} & 58.7 \ci{[57.5, 59.9]} & \textbf{62.8 \ci{[61.7, 63.8]}} & 61.9 \ci{[60.8, 63.0]} & \textbf{63.3 \ci{[62.1, 64.5]}} & \textbf{63.4 \ci{[62.2, 64.5]}} \\
BLEU-4 & 41.1 \ci{[39.8, 42.4]} & 40.9 \ci{[39.6, 42.3]} & 41.0 \ci{[39.7, 42.4]} & 41.2 \ci{[40.0, 42.6]} & 42.2 \ci{[40.9, 43.5]} & \textbf{44.7 \ci{[43.5, 46.0]}} & \textbf{43.8 \ci{[42.5, 45.1]}} & \textbf{45.0 \ci{[43.7, 46.4]}} & \textbf{45.2 \ci{[43.8, 46.5]}} \\
METEOR & 53.7 \ci{[52.6, 54.8]} & 53.7 \ci{[52.6, 54.9]} & 53.7 \ci{[52.5, 54.8]} & 53.5 \ci{[52.5, 54.6]} & 54.6 \ci{[53.6, 55.7]} & \textbf{57.3 \ci{[56.2, 58.4]}} & 56.3 \ci{[55.3, 57.5]} & \textbf{57.6 \ci{[56.4, 58.8]}} & \textbf{57.8 \ci{[56.6, 58.8]}} \\\midrule
\multicolumn{8}{l}{\textbf{Clinical metrics}}\\
RadBERT macro $F_1$ & 14.3 \ci{[12.7, 15.9]} & 15.4 \ci{[13.7, 16.8]} & 15.3 \ci{[13.6, 17.0]} & 20.3 \ci{[18.7, 21.9]} & 21.2 \ci{[19.9, 22.6]} & 40.8 \ci{[39.1, 42.3]} & 39.8 \ci{[38.3, 41.3]} & \textbf{44.1 \ci{[42.6, 45.4]}} &\textbf{44.9 \ci{[43.3, 46.4]}} \\
RadBERT micro $F_1$ & 18.7 \ci{[17.0, 20.3]} & 19.8 \ci{[18.1, 21.6]} & 19.4 \ci{[17.7, 21.1]} & 26.2 \ci{[24.6, 27.8]} & 28.9 \ci{[27.5, 30.4]} & 45.9 \ci{[44.4, 47.4]} & 45.6 \ci{[44.3, 46.8]} & \textbf{50.1 \ci{[48.7, 51.4]}} & \textbf{51.0 \ci{[49.5, 52.2]}} \\
RadFact-CT/Logical $F_1$ (+/-) & 66.5 \ci{[65.4, 67.6]} & 66.3 \ci{[65.3, 67.3]} & \textbf{68.7 \ci{[67.7, 69.6]}} & 67.4 \ci{[66.4, 68.4]} & \textbf{68.2 \ci{[67.3, 69.2]}} & 66.7 \ci{[65.6, 67.9]} & 66.3 \ci{[65.2, 67.3]} & 67.0 \ci{[66.0, 68.1]} & 66.9 \ci{[65.6, 68.0]} \\
RadFact-CT/Logical $F_1$ (+) & 12.8 \ci{[11.4, 14.2]} & 13.2 \ci{[11.7, 14.7]} & 12.8 \ci{[11.4, 14.2]} & 11.5 \ci{[10.5, 12.7]} & 11.7 \ci{[10.8, 12.7]} & 21.1 \ci{[19.9, 22.3]} & 20.7 \ci{[19.6, 21.9]} & \textbf{22.7 \ci{[21.7, 24.0]}} & \textbf{22.4 \ci{[21.2, 23.6]}} \\
RadFact-CT/Logical Precision (+) & 15.8 \ci{[13.7, 17.9]} & 15.2 \ci{[13.3, 17.4]} & 14.8 \ci{[13.0, 16.8]} & 13.9 \ci{[12.6, 15.4]} & 14.1 \ci{[12.9, 15.4]} & 24.4 \ci{[22.9, 26.0]} & 24.6 \ci{[23.2, 26.2]} & \textbf{26.9 \ci{[25.4, 28.5]}} & \textbf{27.0 \ci{[25.4, 28.8]}} \\
RadFact-CT/Logical Recall (+) & 10.8 \ci{[9.4, 12.3]} & 11.6 \ci{[10.2, 13.2]} & 11.3 \ci{[9.7, 12.7]} & 9.9 \ci{[8.9, 11.0]} & 10.1 \ci{[9.1, 11.1]} & \textbf{18.5 \ci{[17.1, 19.8]}} & 17.9 \ci{[16.7, 19.1]} & \textbf{19.7 \ci{[18.5, 20.9]}} & \textbf{19.1 \ci{[17.9, 20.3]}} \\

\bottomrule
\end{tabular}

%% file: tables/results/segmentation.tex
\begin{tabular}{l|rrrrr|r|rrrrr|r|rrrrr|r|rrrrr|r}
\toprule
 & \multicolumn{6}{c|}{LiTS} & \multicolumn{6}{c|}{Lung} & \multicolumn{6}{c|}{HVS} & \multicolumn{6}{c}{KiTS23} \\
 & \multicolumn{5}{c|}{Folds} & & \multicolumn{5}{c|}{Folds} &  &\multicolumn{5}{c|}{Folds} & &\multicolumn{5}{c|}{Folds} & \\
 & 0 & 1 & 2 & 3 & 4 & Avg. & 0 & 1 & 2 & 3 & 4 & Avg. & 0 & 1 & 2 & 3 & 4 & Avg. & 0 & 1 & 2 & 3 & 4 & Avg. \\
\midrule
\multicolumn{3}{l}{\textbf{References} (250k steps)} \\
nnU-Net def. & 83.30 & 77.49 & 81.07 & 83.14 & 75.46 & 80.09 & 66.31 & 69.33 & 66.38 & 79.52 & 70.03 & 70.32 & 70.36 & 69.78 & 67.67 & 69.51 & 64.59 & 68.38 & 86.25 & 86.41 & 85.70 & 88.67 & 83.17 & 86.04 \\
nnU-Net ResEnc-L & 83.21 & 79.30 & 83.04 & 83.71 & 78.75 & 81.60 & 67.58 & 70.52 & 66.46 & 77.31 & 69.83 & 70.34 & 69.54 & 67.98 & 67.06 & 69.09 & 64.98 & 67.73 & 88.60 & 88.82 & 87.97 & 89.39 & 86.05 & 88.17 \\\midrule
\multicolumn{2}{l}{\textbf{Primus-M} (37.5k steps)} \\
From scratch & 78.24 & 71.03 & 74.27 & 76.20 & 72.22 & 74.39 & 72.10 & 53.60 & 61.78 & 55.39 & 70.84 & 62.74 & 67.25 & 65.83 & 64.90 & 64.05 & 61.34 & 64.67 & 79.34 & 78.96 & 77.49 & 81.82 & 76.88 & 78.90 \\
MAE & 81.79 & \underline{78.96} & \textbf{81.50} & 82.97 & \textbf{76.15} & \underline{80.27} & \textbf{74.03} & 57.26 & 65.03 & 67.75 & \textbf{71.52} & 67.12 & \underline{68.78} & 67.44 & \underline{67.23} & \textbf{68.02} & \textbf{64.01} & \textbf{67.10} & 85.03 & 84.09 & 84.24 & \textbf{88.55} & 84.81 & 85.34 \\
\cellcolor{microsoftred!20}{COLIPRI-C} & 77.96 & 76.30 & 74.94 & 81.05 & 73.59 & 76.77 & 69.19 & 54.66 & 63.59 & \textbf{70.96} & 71.04 & 65.89 & 67.08 & 65.37 & 65.47 & 63.74 & 61.55 & 64.64 & 82.31 & 81.23 & 79.41 & 82.64 & 79.51 & 81.02 \\
\cellcolor{microsoftblue!20}{COLIPRI-CR} & 80.01 & 76.14 & 76.77 & 78.65 & 73.87 & 77.09 & 73.49 & 49.34 & 62.15 & \underline{70.80} & 70.83 & 65.32 & 66.40 & 66.73 & 65.84 & 64.48 & 61.95 & 65.08 & 81.06 & 81.06 & 78.95 & 82.70 & 79.81 & 80.71 \\
\cellcolor{microsoftgreen!20}{COLIPRI-CM} & \textbf{84.19} & 78.63 & 78.19 & \underline{83.17} & 75.66 & 79.97 & \underline{73.95} & \textbf{64.99} & \underline{67.50} & 66.54 & 70.74 & 68.74 & \underline{67.78} & \underline{67.67} & \textbf{67.79} & 66.28 & \underline{63.38} & 66.58 & \underline{85.69} & \underline{85.59} & \textbf{85.21} & \underline{88.41} & \underline{85.23} & \textbf{86.03} \\
\cellcolor{microsoftyellow!20}{COLIPRI-CRM} & \underline{83.81} & \textbf{79.47} & \underline{80.08} & \textbf{83.19} & \underline{75.77} & \textbf{80.46} & 72.50 & \underline{63.72} & \textbf{71.09} & 66.36 & \underline{71.27} & \textbf{68.98} & \textbf{69.67} & \textbf{68.23} & 67.14 & \underline{67.05} & 63.18 & \underline{67.05} & \textbf{85.77} & \textbf{85.60} & \underline{84.91} & 87.21 & \textbf{85.48} & \underline{85.79} \\\midrule
\multicolumn{2}{l}{\textbf{Primus-M} (250k steps)} \\
From scratch & 80.77 & 78.11 & 81.20 & 82.42 & 76.23 & 79.75 & 72.79 & 62.93 & 63.98 & 73.07 & 72.79 & 69.11 & 67.88 & 65.59 & 65.33 & 65.78 & 62.13 & 65.34 & 86.94 & 87.47 & 85.46 & 88.37 & 83.02 & 86.25 \\
MAE long & 84.72 & 76.63 & 81.92 & 83.39 & 74.96 & 80.32 & 76.18 & 62.38 & 66.83 & 61.24 & 70.70 & 67.46 & 69.02 & 67.38 & 65.12 & 65.51 & 61.59 & 65.73 & 88.10 & 88.20 & 86.38 & 88.92 & 85.67 & 87.46 \\
COLIPRI-CRM long & 84.03 & 79.63 & 82.58 & 83.57 & 75.72 & 81.11 & 75.70 & 59.49 & 65.51 & 66.53 & 70.59 & 67.56 & 68.62 & 67.74 & 67.14 & 66.11 & 62.50 & 66.42 & 88.65 & 88.21 & 86.21 & 89.66 & 85.69 & 87.68 \\
\bottomrule
\end{tabular}

%% file: tables/results/test_retrieval_extended.tex
\begin{tabular}{ll|rrr|rr}
\toprule
& & \multicolumn{3}{c|}{Recall}& \multicolumn{2}{c}{Rank}\\
Retrieval direction & Model & R@1 & R@5 & R@10 & Median & Mean \\
\midrule
\multirow{5}{*}{Report to image} & CT-CLIP & - & 2.90 & 5.00 & - & - \\
 & \cellcolor{microsoftred!20}{\colipric} & 4.49 & 13.46 & 19.69 & 74 & 198.19 \\
 & \cellcolor{microsoftblue!20}{\colipricr} & 7.23 & 19.36 & 26.99 & 47 & 175.53 \\
 & \cellcolor{microsoftgreen!20}{\colipricm} & 14.53 & 34.16 & 45.34 & 13 & 68.49 \\
 & \cellcolor{microsoftyellow!20}{\colipricrm} & 15.27 & 35.10 & 46.01 & 13 & 73.56 \\\midrule

\multirow{4}{*}{Image to report} & \cellcolor{microsoftred!20}{\colipric} & 2.58 & 9.48 & 14.96 & 107 & 218.17 \\
 & \cellcolor{microsoftblue!20}{\colipricr} & 4.71 & 13.93 & 20.76 & 69 & 184.37 \\
 & \cellcolor{microsoftgreen!20}{\colipricm} & 10.57 & 26.50 & 37.52 & 23 & 76.08 \\
 & \cellcolor{microsoftyellow!20}{\colipricrm} & 11.54 & 29.08 & 40.10 & 20 & 79.82 \\
\bottomrule
\end{tabular}

%% file: tables/appendix/segmentation_mae_init.tex
\begin{tabular}{l|rrrr|rr}
\toprule
 & \multicolumn{4}{c|}{Segmentation } & \multicolumn{2}{c}{Zero-shot (S)}\\
Datasets & LiTS & Lung & HSV & KiTS23 & \multicolumn{2}{c}{CT-RATE}\\
Configuration & \multicolumn{4}{c|}{Dice Similarity Coefficient} & AUPRC & AUROC \\
\midrule
From scratch & 74.39 & 62.74 & 64.67 & 78.90 & - & - \\
\cellcolor{microsoftred!20}{\colipric} & 76.77 & 65.89 & 64.64 & 81.02 & 43.74 & 77.05 \\\midrule
\cellcolor{microsoftgreen!20}\cellcolor{microsoftgreen!20}{\colipricm} & 79.97 & 68.74 & 66.58 & 86.03 & 50.70 & 81.66 \\
\textit{w/o MAE init} & 76.87 & 63.11 & 64.53 & 79.82 & 45.53 & 78.18\\
$\Delta$ & -3.10 & -5.63 & -2.05 & -6.21 & -5.17 & -3.48 \\

\bottomrule
\end{tabular}